%% file: main.tex
\pdfoutput=1
\documentclass{article}
\usepackage[T1]{fontenc}
\usepackage[utf8]{inputenc}
\usepackage{textcomp}
\usepackage{microtype}
\usepackage{graphicx}
\usepackage{subcaption}
\usepackage{booktabs}
\usepackage{multirow}
\usepackage{array}
\usepackage{tikz}
\usetikzlibrary{positioning,calc,arrows.meta}
\usepackage{adjustbox}
\usepackage{float}
\usepackage{placeins}
\usepackage{longtable}
\usepackage{hyperref}

\usepackage[preprint]{icml2026}
\usepackage{amsmath}
\usepackage{amssymb}
\usepackage{mathtools}
\usepackage{amsthm}
\usepackage[capitalize,noabbrev]{cleveref}
\hypersetup{pdfsubject={Preprint}}
\newcommand{\holmt}{\textsuperscript{\dag}}
\newcommand{\holma}{\textsuperscript{\ddag}}
\DeclareUnicodeCharacter{2212}{\textminus}
\DeclareUnicodeCharacter{00D7}{\texttimes}
\DeclareUnicodeCharacter{2264}{\ensuremath{\le}}
\DeclareUnicodeCharacter{2265}{\ensuremath{\ge}}
\DeclareUnicodeCharacter{2248}{\ensuremath{\approx}}
\DeclareUnicodeCharacter{2192}{\ensuremath{\rightarrow}}
\DeclareUnicodeCharacter{00B2}{\textsuperscript{2}}
\definecolor{cTilt}{HTML}{2A78D6}
\definecolor{cKeep}{HTML}{EB6834}
\definecolor{cConst}{HTML}{8A8984}
\definecolor{cContrast}{HTML}{52514E}
\newcommand{\clip}{\operatorname{clip}}
\newcommand{\sig}{*}
\newcommand{\dsv}{DeepSeek-V2-Lite}

\icmltitlerunning{Replacement Operators and Routing Semantics in Training-Free MoE Expert Pruning}

\begin{document}

\twocolumn[
\icmltitle{\texorpdfstring{What We Keep of What We Forget: \\[4pt]
  {\large Replacement Operators and Routing Semantics in Training-Free MoE Expert Pruning\endgraf}}%
  {What We Keep of What We Forget: Replacement Operators and Routing Semantics in Training-Free MoE Expert Pruning}}
  \input{authors}
  \icmlkeywords{mixture of experts, expert pruning, model compression, routing, calibration}
\vskip 0.3in
]

\printAffiliationsAndNotice{}

\begin{abstract}
Training-free expert pruning makes two decisions: which experts to remove, and what the layer computes when its router still selects a removed expert. Published methods change both at once. Holding the mask fixed, we compare replacement operators from three families on OLMoE, DeepSeek-V2-Lite and Moonlight, with Granite as a fourth router, using paired, multiplicity-corrected tests. The largest effect is a choice usually left implicit: the survivors' gate scale. On routers trained without top-$k$ renormalization, renormalizing the survivors' gates, a common implementation choice, costs 0.8--8.5 points and reverses whether rerouting beats MoNE's mean-output replacement; on two routers trained with renormalization, the trained scale wins at 50\% pruning. We introduce the \emph{tilted novice}, a closed-form low-rank map of the pruned expert's input whose rank-0 case is MoNE's constant. At 50\% pruning it adds 1.2--1.8\% parameters and gains 3.45 and 1.37 points over the constant on OLMoE and Moonlight, stable across calibration draws and largely unchanged by 4- and 3-bit quantization. No operator dominates: trained-scale rerouting is best on DeepSeek-V2-Lite, and a partial copy of each pruned expert is statistically indistinguishable from the tilted novice on two models. Evaluation can change conclusions: at 25\% pruning MoNE's constant keeps 92--97\% of dense multiple-choice accuracy but only 5--65\% of GSM8K and 0--17\% of code scores, and the calibration domain moves the mask 4--9 times more than resampling. We propose a five-item reporting checklist.
\end{abstract}

\section{Introduction}
\label{sec:intro}

Mixture-of-Experts (MoE) layers let a language model hold many more parameters than it uses for any one token \citep{shazeer2017moe,lepikhin2021gshard,fedus2022switch}. Recent open models use fine-grained designs: dozens to hundreds of small routed experts per layer, a top-$k$ router, and often a few always-on shared experts \citep{dai2024deepseekmoe,deepseek2024v2,muennighoff2025olmoe,liu2025moonlight}. Conventional deployments retain the routed expert weights in accelerator-accessible memory even though each token activates only a subset, making expert storage a substantial deployment cost. Training-free expert pruning removes selected experts without gradient-based retraining and can be combined with quantization \citep{lasby2026reap,tseng2026hope}.

A pruning method makes two decisions. The first is \emph{selection}: which experts to remove. It has received almost all of the attention, from routing frequency \citep{muzio2024seer} to router-weighted activation norms \citep{lasby2026reap}, unified scoring families \citep{liu2026unified}, second-order interaction objectives \citep{tseng2026hope}, reconstruction-based search \citep{lu2024naee,li2026omp} and learned importance \citep{bai2025diep,sun2026mora}. The second is \emph{replacement}: what the layer computes when the router still selects a removed expert. Methods answer it differently, mostly implicitly. REAP and HOPE drop the expert and renormalize the surviving gates. MoNE keeps the routing slot and substitutes the expert's mean output \citep{zhang2026mone}. MoRA substitutes a retained expert passed through a fitted affine map \citep{sun2026mora}, ConMoE hands the slot to a retained expert with the original gate \citep{yao2026conmoe}, and COMPEL adds a share of the pruned experts' weights into the survivors \citep{yoon2026compel}. Because methods differ in both decisions at once, a published comparison of pruning methods is also an uncontrolled comparison of replacement operators. The choice is also entering deployment formats: in October 2026, vLLM's compression tooling added a checkpoint schema for REAP-style pruned and MoNE-style novice experts.\footnote{\url{https://github.com/vllm-project/compressed-tensors/pull/938} and \url{https://github.com/vllm-project/llm-compressor/pull/3343}.} The distinction itself is not new: ConMoE separates building the retained pool from assigning the removed experts' workload and studies assignment on a fixed pool \citep{yao2026conmoe}, OptEP treats selection and post-pruning adaptation as two coupled problems \citep{markovic2026optep}, and REAP shows that discriminative and generative evaluation can reach different conclusions \citep{lasby2026reap}. What is missing is a controlled comparison across replacement families, routing semantics, masks and evaluation protocols; MoNE's and ConMoE's ablations each compare operators within their own method.

We isolate replacement by holding the pruning mask fixed and varying the computation used for a removed expert's routing slot. We study three 64-expert architectures---OLMoE-1B-7B, DeepSeek-V2-Lite and Moonlight-16B-A3B---and use Granite-3.0-3B-A800M for an additional routing-normalization test. At 50\% pruning we also cross four expert-selection criteria with the main operators on OLMoE and DeepSeek-V2-Lite. The comparison includes mean-output constants, a fitted low-rank input map, survivor-based replacements, equal-budget compressed copies, and two forms of rerouting that differ only in the gate scale the surviving experts receive. That last difference turns out to matter as much as the choice of operator. To our knowledge, this is the first fixed-mask comparison of replacement operators across operator families and router conventions.

\begin{table}[t]
\caption{At 50\% pruning, the better gate scale is the one the router was trained with. Rerouting at the trained gate scale minus rerouting at the other scale, nine-task accuracy (points), MoNE masks, calibration draw A; positive values favor the trained scale. Trained scale: the unpruned router's gates (keep) on OLMoE and \dsv{}, renormalization over the selected survivors on Moonlight and Granite (\cref{sec:routing}). \dsv{} and Moonlight both have 64 routed experts, top-6 and two shared experts. \holmt{} / \holma: survives Holm within its table / also over all 111 comparisons; unmarked values have 95\% intervals containing zero. Intervals: \cref{tab:renorm,tab:granite,tab:moonkeep}.}
\label{tab:teaser}
\centering\footnotesize
\setlength{\tabcolsep}{3pt}
\begin{adjustbox}{max width=\columnwidth}
\begin{tabular}{@{}lccccc@{}}
\toprule
 & Trained with & Trained & \multicolumn{3}{c}{Pruned} \\
\cmidrule(l){4-6}
Model & renorm.? & scale & 25\% & 50\% & 75\% \\
\midrule
OLMoE & no & keep & +4.50\holma & +3.39\holma & +3.09\holma \\
\dsv{} & no & keep & +0.79\holmt & +4.67\holma & +8.47\holma \\
Moonlight & yes & renorm & $-$0.17 & +1.68\holma & +1.60\holmt \\
Granite & yes & renorm & +0.28 & +2.08\holma & +0.27 \\
\bottomrule
\end{tabular}
\end{adjustbox}
\end{table}

Our principal findings are:
\begin{itemize}
\item \textbf{Gate scale is part of the pruning algorithm: the trained-scale rule.} After experts are deleted, rerouting can give each selected survivor the gate the unpruned router would give it, or rescale the survivors as if the pruned experts had never existed. At 50\% pruning the better choice, on all four routers, is the one the router was trained with; we call this pattern the \emph{trained-scale rule} (\cref{tab:teaser}). It is easy to violate. On OLMoE and \dsv{}, trained without top-$k$ renormalization, renormalizing the survivors, a common implementation choice, costs 0.8--8.5 points at every ratio, lowers ranking ability on two-choice tasks, and reverses whether rerouting beats MoNE's constant; on \dsv{} it costs more than any operator or competitive selection criterion in our grid. On Granite and Moonlight, trained with renormalization, the direction reverses (at 50\% on Granite, at 50\% and 75\% on Moonlight), and with each router at its trained scale Granite orders five operators at 50\% exactly as OLMoE does. \dsv{} and Moonlight share an expert layout but not a router convention (they also differ in gate function and training). Because keep-scale rerouting needs the pruned experts' router rows, the rule also constrains what a pruned checkpoint must store (\cref{sec:routing,sec:renorm,sec:granite}).
\item \textbf{The tilted novice, a fitted input-dependent replacement.} We introduce the \emph{tilted novice} (\emph{tilt} for short), a rank-$r$ linear map of the pruned expert's input, fitted in closed form from streamed calibration statistics, whose rank-0 case is MoNE's constant (its ``novice''). At 50\% pruning it adds 1.2--1.8\% of the pruned model's parameters and improves on the constant by 3.45 points on OLMoE and 1.37 on Moonlight, both surviving Holm correction over all comparisons; the gain moves by at most 0.20 points across three calibration draws. It also improves on the constant at 75\% on all three models (Holm-significant on two), and on Granite it beats both the constant and trained-scale rerouting at every ratio (outside the Holm families). On OLMoE its rank-8 version beats keeping one more expert per layer with a sixth of the parameters, and at 75\% pruning it lowers WikiText perplexity from 1,434 to 172. Under 4- and 3-bit GPTQ quantization its gain over the constant is not detectably reduced in 11 of 12 settings, whereas rerouting's gain falls in 11 of 12 (significantly in 5) (\cref{sec:operators,sec:copies,sec:quant}).
\item \textbf{A fixed-mask comparison across replacement families.} No operator is best on every model: trained-scale rerouting beats the tilt on \dsv{} at 50\% and 75\%, and the tilt beats rerouting on Moonlight at both ratios and on OLMoE at 50\%; averaged over four masks, the OLMoE and \dsv{} differences survive Holm correction (+2.39 and $-$1.22). A partial copy of each pruned expert, its highest-energy neurons kept exactly at the tilt's parameter budget, is the second budgeted replacement that works: it is statistically indistinguishable from the tilt on OLMoE and \dsv{} (post-hoc bounds rule out differences beyond 0.81 and 0.73 points) and behind it on Moonlight, whereas whitened or plain SVD copies do not detectably beat the constant, even at four times the budget. Survivor-based replacements (cross-expert proxies, slot remapping, Fisher-weighted weight compensation) do not reach trained-scale rerouting on any model in our re-implementations. Across four selection masks on OLMoE and \dsv{}, the highest-scoring of the constant, the tilt and rerouting is the same within each model, but the rest of the order and the size of every gap depend on the mask (\cref{sec:grid,sec:copies,sec:others}).
\item \textbf{Standard evaluation can change the conclusion.} At 25\% pruning MoNE's constant keeps 92--97\% of dense multiple-choice accuracy but only 5--65\% of GSM8K and 0--17\% of code scores (MBPP and HumanEval), and generation can rank operators differently. Changing the calibration domain moves the pruned set 4--9 times as much as resampling general text at a matched token budget, and in a crossed test on OLMoE fitted operators inherit the domain of their fit. Calibrated in-domain, OLMoE pruned by half with trained-scale rerouting keeps 97\% of its dense GSM8K score (44.28 against 45.49) with no fitted parameters. On \dsv{}, 25\% pruning with the tilt and 4-bit weights beats uniformly quantized experts at matched storage on GSM8K by 4.25 points; on general tasks, uniform 3-bit expert quantization remains stronger at the budget we tested (\cref{sec:eval,sec:domain,sec:quant}).
\end{itemize}
The study also contributes a selection analysis (seven published criteria against MoNE's, operator- and interaction-aware selection, Fisher-weighted selection, and a single super expert that explains the frequency criterion's failure on \dsv{}; \cref{app:selection}); re-implementations of three replacement operators whose code is not public, each validated against its paper's equations (\cref{app:mora,app:others}); a reproduction of MoNE within 0.32 points that documents two implementation details its results depend on (\cref{app:G2}); and a statistical protocol of paired item-bootstrap intervals, Holm correction over the 111 contrasts the conclusions rest on, and replication across disjoint calibration draws (\cref{app:holm}). From these results we derive a five-item reporting checklist for expert-pruning comparisons (\cref{sec:discussion}).

\textbf{Scope.} Reduced-rank regression is classical \citep{izenman1975rrr}. Our contribution is the tilted novice as a replacement operator for pruned experts (its formulation as an extension of MoNE's constant, its streamed closed-form fit and parameter accounting, and its evaluation against rerouting, survivor-based operators and equal-budget copies), not a new estimator. Our claims are limited to one-shot, training-free pruning of the tested base models, with 40--64 routed experts per layer, and to the two endpoint gate scales. We do not infer preservation of open-ended generation from multiple-choice scores.

\section{Related Work}
\label{sec:related}

\textbf{Expert selection.} Training-free expert pruning scores each expert on calibration data and removes the lowest-scoring ones per layer. Criteria range from activation frequency and soft counts \citep{muzio2024seer}, reconstruction search over expert subsets \citep{lu2024naee} and expert activation norm \citep{jaiswal2025fantastic} to MoNE's product of access frequency and output variance \citep{zhang2026mone} and REAP's gate-weighted output norm, which REAP motivates with a reconstruction bound under which pruning beats merging for generation \citep{lasby2026reap}. \citet{liu2026unified} place frequency, SEER, EAN and REAP in one family and recommend mean activation norm as the most stable across calibration domains; AIMER scores experts from weights alone \citep{liu2026aimer}. Other lines search subsets evolutionarily \citep[EEP;][]{liu2024eep}, follow expert pruning with unstructured pruning \citep[STUN;][]{lee2025stun}, prune weights inside experts with router statistics \citep[MoE-Pruner;][]{xie2024moepruner}, or prune coupled channels inside experts \citep[UC-RAMP;][]{xu2026ucramp}. Second-order and interaction-aware criteria include HEAPr's output-space Hessian \citep{li2026heapr}, HOPE's quadratic program over pairwise interactions between pruned experts, which recovers REAP when the interactions are dropped \citep{tseng2026hope}, orthogonal matching pursuit \citep{li2026omp}, Shapley values \citep{huang2025shapley}, differentiable scores \citep{bai2025diep}, learned router biases \citep{sun2026mora}, and RAZOR's score of how far surviving experts can compensate a deletion \citep{song2026razor}. Removing the few \emph{super experts} that create massive activations is catastrophic \citep{su2026super,sun2024massive}. These works vary selection and fix the operator, usually to renormalized rerouting. We fix selection and vary the operator, and at 50\% we cross four masks with the operators (\cref{sec:grid}).

\textbf{Replacing a removed expert.} MoNE substitutes each pruned expert with a \emph{novice}, its mean output over calibration tokens, and its ablation shows the novice helps most at high pruning ratios \citep{zhang2026mone}. This is the per-expert form of a classical device: FLAP compensates pruned channels of dense LLMs with bias terms from their mean activations \citep{an2024flap}, and quantization uses the same mean-shift correction \citep{nagel2019dfq}. We show that the constant is the rank-0 member of a family of fitted input maps, the tilted novice, fitted by classical reduced-rank ridge regression \citep{izenman1975rrr}. MoRA maps a retained proxy expert through a channel-wise affine fit, used only when its held-out $R^2$ exceeds a threshold \citep{sun2026mora}. ConMoE builds a pool of retained experts and assigns each removed expert's routed workload to one of them, by parameter distance or a router-weighted functional cost, keeping the router and gate values \citep{yao2026conmoe}, so it already separates pool construction from workload assignment; its fixed-pool ablation compares assignment rules but not remapping against a constant, an input map or rerouting, and on OLMoE-1B-7B-0125 at 50\% it reports MoNE ahead of ConMoE (58.90 against 58.03 on six tasks), consistent with our remapping results (\cref{app:reported}). COMPEL prunes layer-adaptively and adds a Fisher-weighted share of the pruned experts' summed weights into every survivor \citep{yoon2026compel}. OptEP selects the retained set through a continuous relaxation of layer-output reconstruction and then updates the retained experts' weights against a layerwise reconstruction objective regularized toward their pretrained values \citep{markovic2026optep}; in our terms its second stage is, like COMPEL, a compensation operator that moves the pruned experts' work onto the survivors. Its gains are reported together with a new selection rule, so they do not separate the two decisions either. DERN recombines pruned experts' neuron segments into compatible survivors without adding memory \citep{zhou2025dern}; CD-MoE condenses MoE layers of shared-expert models into small dense layers \citep{cao2025cdmoe}; LightMoE replaces experts with shared bases and adapters and needs fine-tuning \citep{hao2026lightmoe}. Outside MoE, ReplaceMe \citep{shopkhoev2025replaceme} and LLM-Streamline \citep{chen2025streamline} replace pruned transformer blocks with maps fitted to calibration activations, and the near-linearity of maps between layers \citep{razzhigaev2024linear} is one reason to expect an input map to work.

Expert merging keeps the router's targets but shares parameters: experts are merged by routing statistics \citep[MC-SMoE;][]{li2024mcsmoe}, by hierarchical clustering of outputs \citep[HC-SMoE;][]{chen2025hcsmoe}, in a shared subspace \citep[Sub-MoE;][]{li2026submoe} or with Fisher weights \citep{matena2022fisher}. REAP argues that merging loses to pruning on generation because it collapses the router's independent control over the merged experts \citep{lasby2026reap}; REAM, which merges with router-weighted activations, often outperforms REAP \citep{jha2026ream}, and PRIME chooses per expert between pruning and merging and refits the router in closed form \citep{huang2026prime}. We evaluate slot remapping, the merging-free special case that maps a pruned expert's router index to a retained expert, and we do not run merging pipelines ourselves. As an indicative cross-paper reference, on MoRA's task set and with the same C4 calibration, trained-scale rerouting on MoNE's masks scores 1.6--11.3 points above the MC-SMoE and HC-SMoE results MoRA reports for \dsv{} and Moonlight in all four settings (\cref{app:merging}).

\textbf{Routing after compression.} When experts are deleted, tokens that would have used them go to the next-highest survivors, and the surviving gate weights change. REAP and HOPE decompose the resulting error into a substitution term and a renormalization term and treat the latter as small \citep{lasby2026reap,tseng2026hope}; both assume a router whose top-$k$ gates are renormalized. OLMoE and \dsv{} are trained without that renormalization: their top-$k$ gates sum to well under one, and masking pruned logits before the softmax inflates every surviving gate by $1/(1-P_{\text{pruned}})$. \citet{chen2026halving} study the compute-side counterpart on renormalizing routers: when fewer experts are activated at inference, renormalization calibrates the expert branch's gain to the training top-$k$, removing it is catastrophic, the best normalizer lies between the endpoints, and the invariant is the reference set the router was trained with rather than renormalization as such. They note, without testing such a model, that a router trained without renormalization would keep its trained gain if left unrenormalized; both of their models renormalize. Our setting is the memory side of the same principle: experts are deleted, the router refills the top-$k$ from survivors, and on routers trained without renormalization the usual implementation changes the scale implicitly. We test both router conventions and compare the two endpoint scales only (\cref{sec:renorm,sec:granite}). Recent methods each use one semantics. AIMER deletes each pruned expert together with its router row and keeps top-$k$ over the rest \citep{liu2026aimer}; on a softmax router trained without renormalization this is our renormalized rerouting, and AIMER's models include OLMoE-1B-7B-0125-Instruct, which shares the router convention on which we find that scale costs accuracy (\cref{sec:renorm}). MoRA's formulation removes pruned experts from routing with normalized selected weights \citep{sun2026mora}; in our re-implementation, MoRA's own \dsv{} masks gain 0.47 and 1.24 points on MoRA's task set (25\%, 50\%; both intervals excluding zero) under the trained scale (\cref{app:mora-semantics}). OptEP renormalizes the gates over the retained experts \citep{markovic2026optep} on Qwen3, GLM-4.5-Air and Mixtral, whose routers are trained with top-$k$ renormalization, which is the scale our results favor for such routers. Self-distillation that lets a post-trained MoE skip half its experts makes a related argument about the residual scale of the expert branch \citep{lv2026skip}, and a study of dynamic expert pruning on twelve fine-grained checkpoints finds that about two thirds of the selected experts preserve about 99\% of performance \citep{chen2026twothirds}. HARC realigns the router in closed form after model merging \citep{huang2026harc}, and Router KD distils the original model's next-token distribution into the router alone \citep{hyeon2026routerkd}; our keep-scale rerouting is the training-free counterpart, which changes only how the existing router's output is used. Dynamic methods skip experts per token \citep{lu2024naee,chen2026halving}; they save compute, not memory.

\textbf{Compressing experts.} A second line keeps every expert but stores it more cheaply, by low-rank or shared-basis decompositions \citep{yang2024moei2,li2025moesvd,gu2025d2moe,chen2026mobe,ai2025resmoe,chakrabarti2026lorexperts,cary2026gsmoe}; we use the truncation-aware, activation-whitened SVD of \citet{wang2025svdllm} to build an equal-budget compressed copy of each \emph{pruned} expert. Quantization is the main alternative at a fixed memory budget \citep{lasby2026reap,huang2025mc,frantar2023gptq}; we compare with it at matched storage and test whether each operator's gain survives it (\cref{sec:quant}).

\textbf{Calibration and evaluation.} In dense pruning, arithmetic calibration data can match pretraining text and a task's own data does not reliably help that task \citep{bandari2024c4}. For MoEs, general calibration collapses GSM8K on OLMoE and task calibration restores it on OLMoE-Instruct \citep{zhang2026mone}, REAP finds the same for code \citep{lasby2026reap}, domain-contrastive scores extract coding specialists \citep{jha2026codeprune}, EASY-EP prunes from a few domain demonstrations \citep{dong2025easyep}, and pruning sets transfer poorly across domains \citep{chen2026halving}; for domain-general pruning, diverse pretraining-like text can match test-distribution data \citep{salim2026calib}.  We add how far the domain moves the mask relative to resampling, at the full and at a matched token budget, and a crossed test of whether the mask or the fitted operator carries the domain effect. Knowledge-intensive and reasoning tasks degrade before multiple-choice benchmarks under compression \citep{jaiswal2024kick}, multiple-choice scoring is sensitive to answer-option bias \citep{zheng2024selectors}, and perplexity and accuracy can favor different settings \citep{chen2026halving}, with near-dense perplexity coexisting with collapsed MATH500 and HumanEval under micro-expert pruning \citep{xu2026ucramp}. Most MoE pruning papers report single runs; Shapley-MoE reports seed variance \citep{huang2025shapley}, and ConMoE reports paired bootstrap intervals and repeats pool construction and assignment over three calibration samples \citep{yao2026conmoe}.

\section{Setup}
\label{sec:setup}

\begin{figure*}[t]
\centering
\begin{tikzpicture}[font=\scriptsize, >={Stealth[length=4pt]},
    stage/.style={draw=black!55, rounded corners=2pt, inner sep=3pt, anchor=north west},
    hdr/.style={font=\scriptsize\bfseries, anchor=south west, inner sep=1pt}]
  \node[stage] (cal) at (0,0) {\parbox[t][3.2cm][t]{2.85cm}{\hyphenpenalty=10000\exhyphenpenalty=10000\raggedright Zyda-2, one cached stream\\[2pt]
    \textbf{A}\ \ sequences 0--999\\ \textbf{B}\ \ 1,000--1,999\\ \textbf{C}\ \ 2,000--2,999\\[2pt]
    \textcolor{black!60}{GSM8K-train: domain study only}}};
  \node[stage, right=0.38cm of cal.north east, anchor=north west] (mask) {\parbox[t][3.2cm][t]{2.85cm}{\hyphenpenalty=10000\exhyphenpenalty=10000\raggedright MoNE's released code on \textbf{A}\\
    fusion score, per layer\\[2pt]
    \textbf{25 / 50 / 75\%} of routed experts\\[2pt]
    \textcolor{black!60}{50\%: also SGE, REAP and frequency masks (grid)}\\[2pt]
    \textbf{mask fixed from here on}}};
  \node[stage, right=0.38cm of mask.north east, anchor=north west] (op) {\parbox[t][3.2cm][t]{2.85cm}{\hyphenpenalty=10000\exhyphenpenalty=10000\raggedright \textbf{seven operators}, same mask\\[2pt]
    \emph{from the pruned expert:} constant, tilt\\[1pt]
    \emph{from a survivor:} MoRA proxy, remap\\[1pt]
    \emph{reroute:} keep, renorm, COMPEL\\[2pt]
    fitted on \textbf{A}; \textbf{B}, \textbf{C} for the draw checks}};
  \node[stage, right=0.38cm of op.north east, anchor=north west] (ev) {\parbox[t][3.2cm][t]{2.85cm}{\hyphenpenalty=10000\exhyphenpenalty=10000\raggedright nine zero-shot multiple-choice tasks; big-5; 5-shot\\[2pt]
    GSM8K, HumanEval, MBPP\\[2pt]
    perplexity: WikiText, C4, held-out Zyda-2\\[2pt]
    0 tokens reach a pruned expert}};
  \node[stage, right=0.38cm of ev.north east, anchor=north west] (st) {\parbox[t][3.2cm][t]{2.85cm}{\hyphenpenalty=10000\exhyphenpenalty=10000\raggedright paired, task-stratified item bootstrap ($B=10{,}000$)\\[2pt]
    Holm within each table and over 111 specified contrasts\\[2pt]
    same sign on every draw (stability check)\\[2pt]
    exploratory contrasts identified in the appendix}};
  \foreach \a/\b in {cal/mask, mask/op, op/ev, ev/st} \draw[->, black!60, thick] ([yshift=-1.7cm]\a.north east) -- ([yshift=-1.7cm]\b.north west);
  \node[hdr] at (cal.north west) {1 \ Calibration};
  \node[hdr] at (mask.north west) {2 \ Mask (held fixed)};
  \node[hdr] at (op.north west) {3 \ Operator (varied)};
  \node[hdr] at (ev.north west) {4 \ Evaluation};
  \node[hdr] at (st.north west) {5 \ Statistics};
\end{tikzpicture}
\caption{Protocol. Selection is done once and held fixed; only the replacement operator varies. Calibration noise is measured by refitting operators on disjoint draws with the mask unchanged; test-item noise by the paired bootstrap; multiplicity by Holm.}
\label{fig:protocol}
\end{figure*}
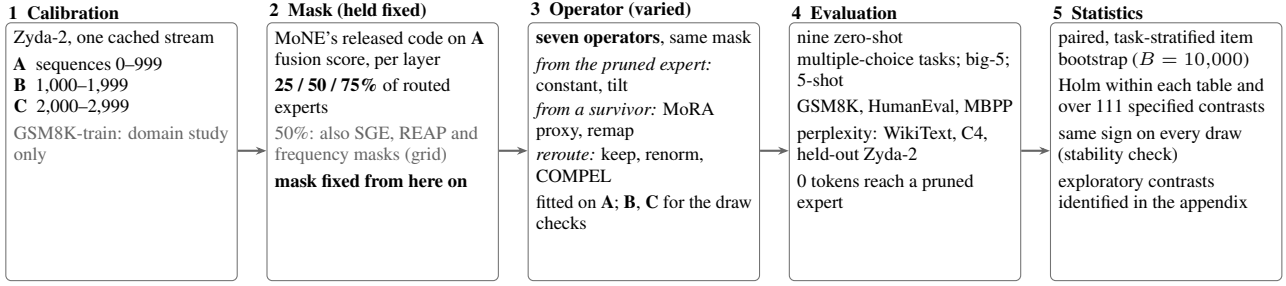

Every comparison holds the mask, the calibration tokens and the evaluation fixed and varies only what fills the pruned experts' routing slots. \Cref{fig:protocol} summarizes the protocol.

\begin{table}[t]
\caption{Models, top-$k$ routing over the routed experts of each MoE layer. Renorm.: the router is trained with its top-$k$ gates renormalized to sum to one (\cref{sec:routing} gives each router's gate equations). Granite enters only the gate-scale test (\cref{sec:granite}). Tables write DeepSeek for \dsv{}.}
\label{tab:models}
\centering\footnotesize
\setlength{\tabcolsep}{2.5pt}
\begin{adjustbox}{max width=\columnwidth}
\begin{tabular}{@{}lccccccc@{}}
\toprule
Model & Params & MoE layers & Routed & $k$ & Shared & Gate & Renorm. \\
\midrule
OLMoE-1B-7B-0125 & 6.92\,B & 16 of 16 & 64 & 8 & 0 & softmax & no \\
\dsv{} & 15.7\,B & 26 of 27 & 64 & 6 & 2 & softmax & no \\
Moonlight-16B-A3B & 16\,B & 26 of 27 & 64 & 6 & 2 & sigmoid & yes \\
Granite-3.0-3B-A800M & 3.3\,B & 32 of 32 & 40 & 8 & 0 & softmax & yes \\
\bottomrule
\end{tabular}
\end{adjustbox}
\end{table}

\textbf{Models and masks.}
\Cref{tab:models} lists the models; Moonlight's sigmoid gate uses a routed scaling factor of 2.446. Masks are MoNE's, produced by its released code with the fusion score, per-layer ranking and 1,000 calibration sequences of 4,096 tokens from Zyda-2; we prune 25\%, 50\% or 75\% of routed experts in every MoE layer. For the crossed grid (\cref{sec:grid}) we add 50\% masks from three published criteria computed exactly on the same slice: SGE ($\sum g^2\|f\|^2$), REAP and routing frequency; on \dsv{} the frequency mask keeps its super expert, without which it collapses (\cref{app:E8}), so it prunes 831 instead of 832 experts. MoNE's masks are a strong fixed reference: of seven other published criteria we computed (five on \dsv{}), evaluated under keep-scale rerouting, only SGE scores above them by more than the spread between calibration draws, and only on \dsv{} (\cref{app:E1}). Our reproduction differs from the published nine-task averages by a mean absolute 0.22 points over six OLMoE settings, and by 0.30 and 0.25 points on \dsv{} (50\%, 25\%) and 0.00 and 0.02 on Moonlight. Our \dsv{} 25\% mask computed from 100 calibration sequences is identical to the authors' released checkpoint in all 416 pruned experts. The reproduction also surfaced two details of MoNE's code that its results depend on: the score's frequency term averages routing weight over all tokens, not over routed tokens as printed, and evaluation silently re-initializes pruned experts unless the model path names the architecture (\cref{app:G2}). Every comparison in this paper uses our in-process re-evaluation of these masks (constant at 50\%: 49.09, 59.84 and 60.24), which differs from the reproduction from MoNE's saved checkpoints (49.16, 59.77 and 60.11) by at most 0.13 points. Granite-3.0-3B-A800M (base) \citep{granite2024} enters only \cref{sec:granite}: its masks come from the same MoNE scoring on its own tokenization of slice A, pruning 10, 20 or 30 of its 40 experts per layer, and its MoE blocks are converted to OLMoE's block type with top-$k$ renormalization switched on, a conversion that reproduces the original model exactly in float32 (\cref{app:granite}).

\textbf{Calibration draws.}
Calibration text is streamed from Zyda-2 once, cached, and cut into three disjoint slices of 1,000 sequences: A (sequences 0--999), B (1,000--1,999) and C (2,000--2,999). Masks always come from slice A. Fitted operators are fitted on A, B or C, so the spread across draws measures calibration noise with the mask held fixed. The slices are disjoint parts of one cached stream, not independent random draws, and the mask is not re-derived when refitting, so the spread measures the calibration noise of an operator conditional on its mask, not of the whole pruning pipeline. A GSM8K-train cache (7,473 problems, 1.17--1.42\,M tokens per tokenizer) is used only in \cref{sec:domain}.

\textbf{Replacement operators.}
Let expert $i$ be pruned, $x$ a token routed to it, and $g_i(x)$ its gate. Each operator defines what the layer computes in place of $g_i(x)\,E_i(x)$ (\cref{fig:schema}).
\begin{itemize}
\item[] \emph{Fill the slot from the pruned expert.}
\item \textbf{Constant (MoNE).} $N_i(x)=\mu_i$, the mean output of $E_i$ over the calibration tokens routed to it. The router is unchanged.
\item \textbf{Tilted novice (ours; ``tilt'').} $N_i(x)=\mu_i+U_i\clip\big(V_i(x-\bar{x}_i)\big)$, a rank-$r$ reduced-rank ridge regression \citep{izenman1975rrr} of $E_i(x)$ on $x$ over the same tokens, solved in closed form from streamed sufficient statistics ($r=32$, ridge $10^{-2}$, features clipped at $4\sigma$, touching about 0.01\% of values). It adds $2rd+2r$ parameters per pruned expert over the constant, whose mean it reuses: the input mean $\bar{x}_i$ is not stored, because it is folded into the feature offset $V_i\bar{x}_i$ ($r$ values), and the clip scale takes the other $r$ (\cref{app:B1,app:B2}). Percentages are of the pruned model: at 50\%, 67.1\,M (1.82\%) on OLMoE, 109.1\,M (1.28\%) on \dsv{} and 109.1\,M (1.24\%) on Moonlight. At equal parameter cost, a per-expert map of the full input removes more of the constant's error than one map shared by a layer or a map of the router logits (\cref{app:B3}). Fits use 1,000 calibration sequences; with 100 the gain is not stable across slices (\cref{app:B5}).
\item[] \emph{Fill the slot from a survivor.}
\item \textbf{Cross-expert proxy (MoRA).} $N_i(x)=a\odot E_j(x)+b$ for the retained expert $j$ whose output best predicts $E_i$'s on held-out tokens, used only if that $R^2$ exceeds 0.4 \citep{sun2026mora}; our re-implementation is described in \cref{app:mora}.
\item \textbf{Slot remapping (ConMoE).} The router and the gate are unchanged, and the pruned expert's slot is answered by a retained expert $\pi(i)$ of the same layer: $g_i(x)\,E_{\pi(i)}(x)$. \emph{Nearest}: $\pi(i)$ minimizes ConMoE's parameter distance (its Eq.~13). \emph{Functional}: among the four nearest, the candidate with the lowest router-weighted output error $\mathbb{E}[g_i^2\|E_i-E_p\|^2]$ on up to 64 routed calibration states, accepted only if both halves of the calibration data agree \citep{yao2026conmoe}. No parameters are stored.
\item[] \emph{Drop the slot and reroute.}
\item \textbf{Keep-scale rerouting.} Pruned experts are excluded from the top-$k$ choice, and every selected survivor gets the gate the unpruned router would give it (\cref{sec:routing}). No parameters are added; the router keeps its rows for the pruned experts.
\item \textbf{Renormalized rerouting.} The survivors' gates are rescaled as if the pruned experts did not exist: on softmax routers trained without renormalization (OLMoE, \dsv{}) this is a softmax over the retained experts, REAP's and HOPE's semantics; on routers trained with top-$k$ renormalization (Granite, Moonlight) it is the trained normalization over the selected survivors. We write \emph{trained-scale rerouting} for whichever of the two arms matches the router's training convention: keep-scale on OLMoE and \dsv{}, renormalized on Granite and Moonlight.
\item \textbf{Weight compensation (COMPEL).} Each surviving expert $j$ absorbs $\gamma\,F_j/\sum_{k}F_k$ times the summed weights of the layer's pruned experts ($\gamma=0.1$, $F$ each expert's summed diagonal Fisher, the mean squared norm of the loss gradient with respect to its parameters, on 128 calibration sequences of 1,024 tokens), and the pruned experts are then removed by keep-scale or renormalized rerouting \citep{yoon2026compel}. No parameters are added. We use COMPEL's compensation on our masks, not its layer-adaptive selection.
\end{itemize}

With rerouting, every evaluation counts tokens reaching pruned experts; the count was zero in every run. MoRA, ConMoE and COMPEL had released no code by 27 October 2026 (MoRA announces a release), so we implemented each from its paper's equations. \Cref{app:mora,app:others} list the settings the papers leave open, our choices, and the checks that each implementation computes what its equations specify. Results for these three operators are for our implementations of the operator alone: ConMoE's assignment on a MoNE pool and COMPEL's compensation on MoNE's masks are component comparisons, not evaluations of either method, whose pool construction and layer-adaptive selection may interact with the operator.

\textbf{Routing semantics in equations.}\label{sec:routing}
For a token, let $p_e$ be the router's score of expert $e$ (the softmax probability over all $E$ experts on OLMoE, \dsv{} and Granite; an independent sigmoid on Moonlight), $D$ the top-$k$ of the unpruned router, $S$ the top-$k$ among the retained experts $R$, and $c$ the routed scaling factor (2.446 on Moonlight, 1 elsewhere). On Moonlight both choices rank $p_e+b_e$ with a learned bias $b$, while the gates use $p_e$. The selection $S$ is the same in every rerouting arm; only the gate $g_j$ of a selected survivor $j\in S$ differs ($P_{\mathrm{pruned}}=\sum_{e\notin R}p_e$):
{\small
\begin{align*}
\textit{OLMoE, \dsv{}:}\qquad g^{\mathrm{keep}}_j &= p_j,\\
g^{\mathrm{renorm}}_j &= p_j/(1-P_{\mathrm{pruned}});\\
\textit{Granite, Moonlight:}\qquad g^{\mathrm{keep}}_j &= c\,p_j\big/\textstyle\sum_{e\in D}p_e,\\
g^{\mathrm{renorm}}_j &= c\,p_j\big/\textstyle\sum_{e\in S}p_e.
\end{align*}}
Keep gives every survivor the gate the unpruned router would give it, including an expert that enters the top-$k$ only because a pruned one left. On OLMoE and \dsv{} neither arm makes the selected gates sum to one; on Granite keep's gates sum to less than one whenever a pruned expert was in $D$; on Moonlight, where the bias steers the choice but not the gate, a newly admitted expert can have a larger $p$ than the one it displaced, so keep's total can exceed the trained total. The trained semantics are keep on OLMoE and \dsv{} and renorm on Granite and Moonlight. Keep needs the pruned experts' router rows ($d$ values each) to form the unpruned gate, so a checkpoint that stores only the retained experts and their router rows can express renormalized rerouting but not keep: the trained semantics on Granite and Moonlight, but not on OLMoE and \dsv{}. \Cref{tab:toy} works one token through each case. Every implementation is checked against a reference written from the model's own router code: with an empty mask the replays are bit-exact, the keep arms of Granite and Moonlight match independent references to a relative error below $10^{-6}$, and no token reached a pruned expert in any run (\cref{app:G3}). Tables that span models label the rerouting arms ``trained scale'' and ``other scale''; figures label trained-scale rerouting ``keep'' on every model, so on Moonlight a figure's ``keep'' is its trained, renormalized scale, and Moonlight's keep arm itself is reported in \cref{sec:granite}.

\begin{table}[t]
\caption{One token, four experts, top-2, expert 1 pruned: the gates of each rerouting arm (expert: gate; sum in brackets). The newly admitted expert is 3. Sigmoid row with $c=1$.}
\label{tab:toy}
\centering\footnotesize
\setlength{\tabcolsep}{2.5pt}
\begin{adjustbox}{max width=\columnwidth}
\begin{tabular}{@{}llll@{}}
\toprule
Router & Dense & Keep & Renorm \\
\midrule
Softmax, no renorm., $p=(.4,.3,.2,.1)$ & 1: .400, 2: .300 (.70) & 2: .300, 3: .200 (.50) & 2: .500, 3: .333 (.83) \\
Softmax, renorm., $p=(.4,.3,.2,.1)$ & 1: .571, 2: .429 (1) & 2: .429, 3: .286 (.71) & 2: .600, 3: .400 (1) \\
Sigmoid, renorm., $p=(.9,.6,.5,.2)$ & 1: .600, 2: .400 (1) & 2: .400, 3: .333 (.73) & 2: .545, 3: .455 (1) \\
\bottomrule
\end{tabular}
\end{adjustbox}
\end{table}

\begin{figure}[t]
\centering
\begin{adjustbox}{max width=\columnwidth}
\begin{tikzpicture}[font=\scriptsize, >={Stealth[length=4pt]},
    box/.style={draw, rounded corners=2pt, minimum height=0.48cm, minimum width=0.9cm, inner sep=2pt},
    lab/.style={inner sep=1pt}]
  \node[lab] (x) at (0,0) {$x$};
  \node[box, right=0.35cm of x] (r) {router};
  \node[box, right=1.05cm of r, yshift=0.62cm] (ej) {$E_j$};
  \node[box, right=1.05cm of r, dashed, draw=cConst, text=cConst] (ei) {$E_i$};
  \node[box, right=1.05cm of r, yshift=-0.62cm] (el) {$E_l$};
  \draw[cConst, thick] ($(ei.south west)+(2pt,2pt)$) -- ($(ei.north east)+(-2pt,-2pt)$);
  \node[draw, circle, inner sep=0.5pt, right=0.8cm of ei] (sum) {$+$};
  \node[lab, right=0.3cm of sum] (y) {$y$};
  \draw[->] (x) -- (r);
  \draw[->] (r.east) -- node[above, sloped, lab] {$g_j$} (ej.west);
  \draw[->, cConst] (r.east) -- node[above, lab] {$g_i$} (ei.west);
  \draw[->] (r.east) -- node[below, sloped, lab] {$g_l$} (el.west);
  \draw[->] (ej.east) -- (sum); \draw[->] (el.east) -- (sum);
  \draw[->, cConst, dashed] (ei.east) -- node[above, lab] {?} (sum);
  \draw[->] (sum) -- (y);
  \node[anchor=north, inner sep=0pt] (tab) at ($(ei.south)+(0,-0.95cm)$) {%
    \setlength{\tabcolsep}{3pt}\renewcommand{\arraystretch}{1.12}%
    \begin{tabular}{@{}lll@{}}
      \toprule
      Operator & Computed for the slot of $E_i$ & Stored \\
      \midrule
      \multicolumn{3}{@{}l}{\emph{Fill the slot from the pruned expert}} \\
      \textcolor{cConst}{\textbf{Constant}} (MoNE) & $g_i\,\mu_i$ & $d$ \\
      \textcolor{cTilt}{\textbf{Tilt}} (ours) & $g_i\,[\mu_i+U_i\clip(V_i\tilde{x})]$ & $+\,2rd{+}2r$ \\
      \multicolumn{3}{@{}l}{\emph{Fill the slot from a survivor}} \\
      Proxy (MoRA) & $g_i\,[a\odot E_j(x)+b]$ & $2d$ \\
      Slot remap (ConMoE) & $g_i\,E_{\pi(i)}(x)$, $\pi(i)$ retained & 0 \\
      \multicolumn{3}{@{}l}{\emph{Drop the slot and reroute}} \\
      \textcolor{cKeep}{\textbf{Keep reroute}} & next survivor, unpruned gate & 0 \\
      \textcolor{cContrast}{\textbf{Renorm reroute}} & survivors' gates rescaled (\cref{sec:routing}) & 0 \\
      Compensation (COMPEL) & reroute; survivors $+\,\gamma w_j\!\sum\theta_{\text{pruned}}$ & 0 \\
      \bottomrule
    \end{tabular}};
\end{tikzpicture}
\end{adjustbox}
\caption{Seven operators in three families act on the same mask; they differ only in what the layer computes when the router selects a pruned expert ($E_i$, dashed). Remapping, renormalization and COMPEL move the slot's work onto survivors (\cref{sec:renorm}). $\tilde{x}=x-\bar{x}_i$; $P$ is the pruned experts' softmax mass for the token. The tilt stores $\mu_i$ (already in MoNE's model), $U_i$, $V_i$ and two $r$-vectors; $\bar{x}_i$ is folded into the feature offset.}
\label{fig:schema}
\end{figure}

\textbf{Evaluation.}
Accuracy is the unweighted mean of nine zero-shot lm-eval tasks (ARC-c, ARC-e, BoolQ, COPA, MMLU, OpenBookQA, PIQA, RTE, WinoGrande; acc\_norm for ARC, OBQA and PIQA), MoNE's protocol. Because RTE (277 items) and COPA (100) weigh as much as MMLU (14,042), we also report the big-5 mean (ARC-e, ARC-c, BoolQ, PIQA, WinoGrande; 9,923 items). Generation is GSM8K (5-shot, strict match), HumanEval (pass@1) and, at 25\% pruning, MBPP (pass@1, 3-shot); language modelling is WikiText word perplexity at 2,048 tokens.

For two runs on the same items we resample each task's items with replacement ($B=10{,}000$, seed 0), recompute the nine-task difference, and report the 95\% percentile interval; * marks an interval that excludes zero. These intervals capture test-item noise only. Calibration noise is shown by repeating the comparison on disjoint draws, and we call an effect established only if it has the same sign on every draw, a stability check rather than a statistical treatment of calibration uncertainty. An interval that contains zero means that a difference is not detectable at this precision; it is not evidence that two operators are equivalent. Where the text calls two operators close, it gives the largest difference the data rule out: the bound of the 90\% interval, which is the equivalence margin two one-sided tests at $\alpha=0.05$ would accept. These are post-hoc bounds (\cref{app:tost}).

\textbf{Multiple comparisons.} To control the family-wise error we also compute two-sided bootstrap p-values from the same resamples and apply Holm's step-down correction \citep{holm1979} within each table of contrasts and over all 111 specified comparisons together; 47 survive correction over all of them (\cref{app:holm}). These are the contrasts the text rests on, not every comparison in the paper. The others (Granite's contrasts other than renorm $-$ keep, the per-mask grid contrasts, the in-domain GSM8K comparisons and the other calibration-domain results) are exploratory and are marked as outside the families. In the text, \holmt{} marks a difference that survives Holm within its table and \holma{} one that also survives over all comparisons. Holm addresses test-item noise; the same-sign rule addresses calibration noise.

\textbf{Fit share.} For a fitted operator, the \emph{fit share} is $1-\mathrm{SSE}_{\text{op}}/\mathrm{SSE}_{\text{const}}$, the share of the constant's squared output error it removes on an expert's routed tokens. We distinguish the in-sample unweighted fit from a held-out test on even/odd calibration steps, weighting experts by $n_i\,\mathbb{E}[G_i^2]\,\mathrm{tr}\,\Sigma_i$ in three OLMoE layers (\cref{app:B3}). These metrics are not interchangeable. The \emph{recovery share} of \cref{sec:ppl} is a different quantity: the share of the constant's model-level damage removed.

\section{Does the Replacement Change the Result?}
\label{sec:operators}

On MoNE's own masks, a better operator than the constant improves accuracy at 50\% and 75\% pruning on all three models; which operator is best depends on the model, and, on the two models we crossed, the best of the constant, the tilt and rerouting does not change between masks.

\begin{table}[t]
\caption{Nine-task zero-shot accuracy (\%), MoNE masks, calibration draw A. Dense: OLMoE 66.76, DeepSeek 67.15, Moonlight 71.48. Rerouting at the trained scale is keep-scale on OLMoE and DeepSeek and renormalized on Moonlight; the other scale is the opposite arm (Moonlight's from \cref{tab:moonkeep}). Bold: highest in the row, significant or not; intervals are in \cref{fig:operators}.}
\label{tab:main}
\centering\footnotesize
\setlength{\tabcolsep}{3.5pt}
\begin{tabular}{@{}llcccc@{}}
\toprule
 & & & & \multicolumn{2}{c}{Reroute, gate scale} \\
\cmidrule(l){5-6}
Model & Pruned & Constant & Tilt r32 & Trained & Other \\
\midrule
OLMoE & 25\% & 61.19 & \textbf{63.02} & 62.65 & 58.15 \\
      & 50\% & 49.09 & \textbf{52.54} & 51.48 & 48.08 \\
      & 75\% & 41.04 & \textbf{43.84} & 42.94 & 39.85 \\
\midrule
DeepSeek & 25\% & 65.40 & 65.56 & \textbf{65.66} & 64.87 \\
         & 50\% & 59.84 & 60.32 & \textbf{61.19} & 56.52 \\
         & 75\% & 47.59 & 48.82 & \textbf{51.80} & 43.33 \\
\midrule
Moonlight & 25\% & 68.66 & 68.79 & 68.71 & \textbf{68.88} \\
          & 50\% & 60.24 & \textbf{61.61} & 60.06 & 58.38 \\
          & 75\% & 43.89 & \textbf{47.16} & 45.48 & 43.88 \\
\bottomrule
\end{tabular}
\end{table}

\begin{figure*}[t]
\centering
\includegraphics[width=\textwidth]{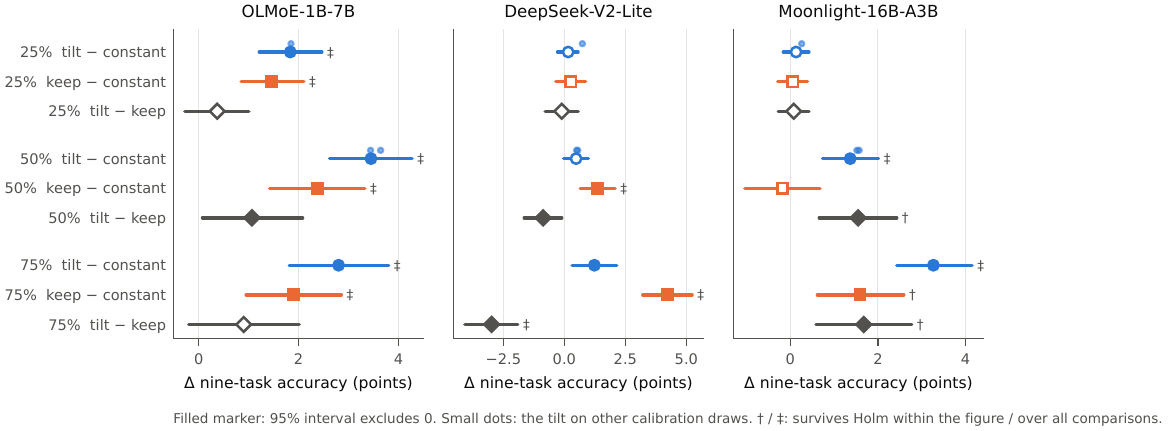}
\caption{Operator differences in nine-task accuracy with 95\% item-bootstrap intervals, MoNE masks, calibration draw A. Filled marker: the interval excludes zero. Small dots: the tilt on calibration draws B and C. \holmt{} / \holma: survives Holm correction within the figure / over all 111 comparisons (\cref{app:holm}). ``Keep'' is rerouting at the router's trained gate scale on every model (renormalized on Moonlight).}
\label{fig:operators}
\end{figure*}

\subsection{Main Results}

\begin{figure}[t]
\centering
\includegraphics[width=\columnwidth]{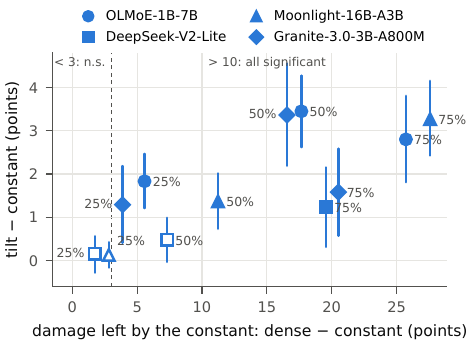}
\caption{The tilt's gain over the constant against the damage the constant leaves (dense $-$ constant), nine-task, MoNE masks, draw A, 95\% intervals; filled: the interval excludes zero. Diamonds: Granite (\cref{sec:granite}), whose contrasts are outside the Holm families. Where the constant loses more than 10 points every gain is significant; below 3 points none is.}
\label{fig:damage}
\end{figure}

\textbf{A fitted replacement helps where pruning leaves damage.} The tilt beats the constant at 75\% on all three models and at 50\% on OLMoE and Moonlight, but at 25\% only on OLMoE (\cref{tab:main,fig:operators}). Its gain follows the damage the constant leaves (\cref{fig:damage}): it is significant in all five settings where that damage exceeds 10 points (+1.23 to +3.45) and in neither where it is below 3 points (+0.16, +0.13), and all five survive Holm correction within the figure except \dsv{} at 75\% (+1.23 [+0.32, +2.15]), which we therefore do not count as established. Granite fits the pattern (+1.29, +3.36 and +1.58 at 25--75\%, outside the Holm families), although at 75\%, within 2.4 points of guessing, most of its gain is BoolQ and may reflect answer bias (\cref{app:granite}). The relation is not one curve across models: OLMoE at 25\% (damage 5.6 points) gains +1.83\holma, more than \dsv{} at 50\% (damage 7.3) gains (+0.48, n.s.), so the tilt's value also depends on the model (\cref{app:A4}). The largest gain is +3.45 [+2.62, +4.27]\holma{} on OLMoE at 50\%, and it is broad: on every draw it is positive on seven of the nine tasks, largest on ARC-e (+13.4 to +13.9) and PIQA (+6.7 to +7.1), and its one consistent loss, RTE, is on draw A a shift in answer bias rather than in ranking ability (\cref{tab:A2,sec:auc}).

\textbf{No operator is best everywhere.} Keep-scale rerouting beats the tilt on \dsv{} at 50\% and 75\%; the tilt beats rerouting on Moonlight at both ratios and on OLMoE at 50\% (the tilt $-$ keep rows of \cref{fig:operators}). The ranking follows the model, not the pruning ratio. On MoNE's mask alone, the 50\% tilt $-$ keep differences on OLMoE (+1.07) and \dsv{} ($-$0.87) do not survive Holm correction; averaged over the four masks of \cref{sec:grid} both do (+2.39\holma{} and $-$1.22\holma). The comparison is matched in memory, not in compute: rerouting evaluates $k$ real experts for every token, whereas the constant and the tilt replace pruned slots with cheaper maps, so at batch 32 rerouting keeps only 5--7\% of MoNE's prefill saving; at batch 1, where prefill is launch-bound, the order reverses (\cref{tab:cost,app:F1}).

\textbf{The gains are stable across calibration slices.} Across three disjoint calibration draws, with the mask held fixed, the tilt's 50\% gain moves by at most 0.20 points on every model (dots in \cref{fig:operators}). On \dsv{} the gain has the same sign on all three draws, but two of its three intervals touch zero, so we call it consistent rather than significant. At 25\% the gain moves by at most 0.58 across two draws. All fits use 1,000 calibration sequences, and the tilt needs them: fitted on two disjoint 100-sequence slices instead, its OLMoE 50\% gain is +3.19 on one and +0.66 on the other (\cref{app:B5}).

\subsection{Does the Ranking Depend on the Mask?}
\label{sec:grid}

\begin{table}[t]

\caption{Crossed mask $\times$ operator grid at 50\% pruning, nine-task accuracy, draw A. Overlap: share of pruned experts shared with MoNE's mask. Bottom rows: each contrast (T tilt, K keep, C constant, R renorm) averaged over the four masks in one paired test (R$-$K over the MoNE and REAP masks); \holma: survives Holm over all comparisons. Per-mask contrasts and selection effects are in \cref{app:gridfull}.}
\label{tab:grid}
\centering\footnotesize
\setlength{\tabcolsep}{2.5pt}
\begin{adjustbox}{max width=\columnwidth}
\begin{tabular}{@{}llccccc@{}}
\toprule
Model & Mask & Overlap & Const. & Tilt & Keep & Renorm \\
\midrule
OLMoE & MoNE & 100\% & 49.09 & 52.54 & 51.48 & 48.08 \\
      & SGE & 80.7\% & 49.81 & 54.34 & 53.08 & --- \\
      & REAP & 79.9\% & 47.89 & 51.78 & 49.00 & 47.12 \\
      & Frequency & 81.8\% & 44.49 & 48.35 & 43.88 & --- \\
\midrule
DeepSeek & MoNE & 100\% & 59.84 & 60.32 & 61.19 & 56.52 \\
      & SGE & 87.1\% & 60.67 & 60.62 & 61.74 & --- \\
      & REAP & 88.5\% & 58.70 & 58.73 & 60.46 & 56.58 \\
      & Freq.\ (SE kept) & 81.7\% & 59.49 & 59.49 & 60.64 & --- \\
\midrule
\multicolumn{3}{@{}l}{\emph{Mean over masks}} & {\scriptsize T$-$C} & {\scriptsize K$-$C} & {\scriptsize T$-$K} & {\scriptsize R$-$K} \\
OLMoE & & & +3.93\holma & +1.54\holma & +2.39\holma & $-$2.64\holma \\
DeepSeek & & & +0.11 & +1.33\holma & $-$1.22\holma & $-$4.28\holma \\
\bottomrule
\end{tabular}
\end{adjustbox}
\end{table}

MoNE's score multiplies routing frequency by output variance, so it tends to prune the experts for which a constant is already a good substitute. Comparing operators only on MoNE's masks could therefore favor the constant. We therefore crossed four 50\% masks with the operators on OLMoE and \dsv{} (\cref{tab:grid}): MoNE's, SGE's and REAP's, which weigh outputs differently, and routing frequency, which ignores outputs entirely.

\textbf{The highest-scoring of the constant, the tilt and rerouting is the same on every mask; the full order is not.} On OLMoE the tilt scores highest on all four masks: it beats the constant on each (+3.45 to +4.52, every interval excluding zero and surviving Holm within the per-mask family) and keep-scale rerouting on each (+1.07 to +4.47; on the MoNE and SGE masks per comparison only). On \dsv{} keep-scale rerouting scores highest on all four: it beats the constant (+1.07 to +1.76) and the tilt ($-$0.87 to $-$1.72) on each, while the tilt is never more than 0.5 points from the constant. The weaker operators change order. On OLMoE's frequency mask keep-scale rerouting falls below the constant (43.88 against 44.49; $-$0.61 [$-$1.40, +0.19]), and on \dsv{}'s SGE mask the tilt falls just below the constant (60.62 against 60.67), reversing their order on MoNE's mask. The tilt's gain is also the less mask-dependent: on OLMoE, keep-scale rerouting's gain over the constant falls from +3.27 on the best mask (SGE) to $-$0.61 on frequency's, while the tilt's stays between +3.45 and +4.52. Of the 28 per-mask contrasts, 16 survive Holm within that exploratory family (\cref{tab:H3}). Averaged over masks in one paired test per contrast, every contrast in \cref{tab:grid} survives Holm over all comparisons except \dsv{}'s tilt $-$ constant. Full per-mask contrasts and multiplicity corrections are in \cref{tab:H3,app:holm}.

\textbf{Over the tested operators and criteria, operator and selection effects are of similar size.} Holding the operator fixed and changing the mask among the three competitive criteria (MoNE, SGE, REAP) moves accuracy by 1.9--4.1 points on OLMoE and 1.3--2.0 on \dsv{}; holding the mask fixed and changing the operator among constant, tilt and keep moves it by 3.5--4.5 and 1.1--1.8. A weak criterion costs more than any operator choice: OLMoE's frequency mask is 4.2--7.6 points below MoNE's. On \dsv{} the gate-scale choice in this comparison (renorm $-$ keep, $-$3.9 to $-$4.7) exceeds both. Selection between competitive criteria is also less stable than it looks: averaged over three calibration draws, SGE's mask is not detectably better than MoNE's on OLMoE after Holm correction under any operator (+0.77, +0.94 and +0.42), although on draw A alone it is +1.79\sig{} under the tilt (\cref{app:gridfull}). In that comparison SGE's mask is re-derived on each draw while MoNE's stays at slice A; with both masks re-derived on slices A and B, SGE's advantage under keep-scale rerouting is +1.07, which is still within the spread between draws (\cref{tab:E1}). These are 50\% results on two models; we did not cross masks at 25\% or 75\%. The comparison is of observed ranges, not a decomposition of how much each decision matters in general.

\textbf{Choosing the mask for the tilt does not help.} Selecting OLMoE's 50\% mask from the tilt's residuals instead of the constant's, per expert or by a per-layer quadratic program over co-routed residuals (an operator-aware analogue of HOPE's objective), moves about 6\% of the mask, less than a new calibration draw does (8--9\%). The experts it moves are the ones the tilt reconstructs best, yet MMLU needs them: the resulting masks lower MMLU by 2.6--4.3 points to near chance while the nine-task mean stays similar. Fisher-weighted selection trades commonsense accuracy for knowledge and is unstable across draws (\cref{app:E5,app:E6}). Being easy to imitate in layer error is not the same as being safe to prune.

\subsection{Accuracy per Added Parameter}
\label{sec:copies}

\begin{figure*}[t]
\centering
\includegraphics[width=\textwidth]{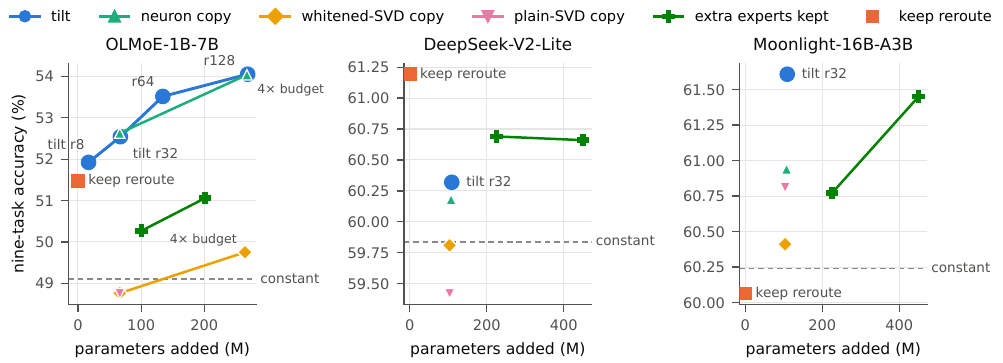}
\caption{Nine-task accuracy at 50\% pruning against parameters added to MoNE's constant model (MoNE masks, draw A); contrasts with intervals are in \cref{tab:budget}. Tilt: ranks 8, 32, 64 and 128 on OLMoE, rank 32 elsewhere. Extra experts: MoNE's mask keeping 33 and 34 of 64 experts per layer, evaluated in the same process as every other arm. Copies at the tilt's rank-32 budget, and at four times it on OLMoE.}
\label{fig:budget}
\end{figure*}

\textbf{The tilt buys accuracy cheaply on OLMoE and holds its own elsewhere.} On OLMoE it reaches 52.54 at 67.1\,M added parameters, more than keeping one or two extra experts per layer (50.27 at +100.7\,M, 51.06 at +201.3\,M; \cref{tab:budget,fig:budget}), and rank 8 already reaches 51.92 at 16.8\,M, beating one extra expert per layer with a sixth of its parameters (+1.65\holma). With half the parameters of one extra expert per layer, the rank-32 tilt is not detectably different from it on \dsv{} ($-$0.37 [$-$1.01, +0.26]; differences beyond 0.90 points are ruled out) and is ahead per comparison on Moonlight (+0.84 [+0.21, +1.48]), which does not survive Holm correction. On \dsv{} keep-scale rerouting, with no parameters, is not detectably different from keeping one extra expert per layer (+0.50 [$-$0.12, +1.13]), and a second extra expert adds nothing detectable ($-$0.04 [$-$0.43, +0.36]). On OLMoE the tilt keeps rising with rank: +0.97\holma{} from rank 32 to 64 and +1.51\holma{} to rank 128, where it scores 54.05, about three points above keeping two extra experts per layer with a third more parameters (51.06 at 201.3\,M; point estimates).

\textbf{A partial copy of the pruned expert is the other budgeted replacement that works.} Instead of fitting a map, one can store a cheaper version of the pruned expert itself, with the mean residual as a bias: its highest-energy neurons kept exactly, a truncation-aware, activation-whitened SVD of its three matrices (the SVD-LLM factorization D²-MoE applies to retained experts; \citealp{wang2025svdllm,gu2025d2moe}), or a plain SVD. The neuron copy works and the SVD copies do not. At the tilt's rank-32 budget the neuron copy is not detectably different from the tilt on OLMoE (tilt $-$ copy $-$0.09 [$-$0.95, +0.78]) or \dsv{} (+0.14 [$-$0.56, +0.83]); post-hoc bounds rule out differences larger than 0.81 and 0.73 points (\cref{app:tost}). On Moonlight the tilt is ahead (+0.66 [+0.09, +1.27]). At four times the budget the neuron copy overtakes the rank-32 tilt on OLMoE (+1.50*), and the tilt at the matching rank ($r=128$, 268.6\,M against 267.4\,M added) is not detectably different from it (+0.01 [$-$1.02, +1.05]; bound 0.89 points). The SVD copies stay near the constant at the tilt's budget and barely move at four times it; the rank-128 tilt beats the 4$\times$ whitened-SVD copy by +4.30\holma. Per-expert fit does not by itself predict accuracy: on average the tilt fits the pruned experts better than the equal-budget neuron copy on every model, yet leads in accuracy only on Moonlight (\cref{app:C3}). \Cref{tab:C2} gives every copy with its intervals.

\subsection{Robustness to the Scoring Protocol}
\label{sec:fewshot}

The tilt's gains on OLMoE and Moonlight do not depend on how multiple choice is scored: they hold under 5-shot prompting, with raw instead of length-normalized accuracy, and on tasks whose answer options change with every item. Rerouting's gains are more protocol-sensitive.

\begin{table*}[t]
\caption{Gain over the constant, zero-shot and 5-shot, nine-task, 50\%, MoNE masks, draw A, 95\% intervals. Reroute: rerouting at the router's trained gate scale. 5-shot accuracy (constant / tilt / reroute): OLMoE 52.23 / 55.24 / 54.26, DeepSeek 62.88 / 62.59 / 65.34, Moonlight 62.38 / 63.92 / 63.66.}
\label{tab:fewshot}
\centering\footnotesize
\setlength{\tabcolsep}{4pt}
\begin{tabular}{@{}lccccc@{}}
\toprule
 & \multicolumn{2}{c}{Tilt $-$ constant} & \multicolumn{2}{c}{Reroute $-$ constant} & Tilt $-$ reroute \\
\cmidrule(lr){2-3}\cmidrule(lr){4-5}\cmidrule(l){6-6}
Model & 0-shot & 5-shot & 0-shot & 5-shot & 5-shot \\
\midrule
OLMoE & +3.45 [+2.62, +4.27]* & +3.01 [+2.25, +3.85]* & +2.38 [+1.42, +3.33]* & +2.03 [+1.16, +2.90]* & +0.98 [+0.12, +1.87]* \\
DeepSeek & +0.48 [−0.03, +0.99] & −0.29 [−0.88, +0.29] & +1.35 [+0.65, +2.08]* & +2.46 [+1.76, +3.17]* & −2.75 [−3.53, −2.00]* \\
Moonlight & +1.37 [+0.73, +2.01]* & +1.54 [+0.70, +2.38]* & −0.18 [−1.04, +0.68] & +1.28 [+0.42, +2.16]* & +0.25 [−0.57, +1.08] \\
\bottomrule
\end{tabular}
\end{table*}

\textbf{The tilt's gain survives five-shot prompting on OLMoE and Moonlight; rerouting's changes.} On OLMoE and Moonlight the tilt keeps 87\% and 112\% of its zero-shot gain at five shots, with intervals excluding zero (\cref{tab:fewshot}). On \dsv{} it adds nothing at five shots while rerouting's gain nearly doubles, so rerouting is that model's operator under both protocols and on perplexity. On Moonlight rerouting moves from no gain to a significant one, and at five shots the two are not detectably different (+0.25 [$-$0.57, +1.08]).

\textbf{The gain is not an answer-bias artifact.} Scoring by raw accuracy instead of acc\_norm moves it by at most 0.5 points (\cref{tab:C5}). On the six tasks whose answer options change with every item, where a preference between fixed answer strings cannot raise accuracy, the gain excludes zero on every draw for OLMoE and Moonlight, and it carries most of OLMoE's gain. On OLMoE's three fixed-answer tasks a BoolQ gain in ranking ability is offset by an RTE loss with no detectable change in ranking ability (AUC change $-$0.008 [$-$0.064, +0.047], \cref{sec:auc}). Detailed scoring checks and representative fixed-answer results are in \cref{app:scoring,tab:aucci}.

\subsection{What the Tilt Restores}
\label{sec:restore}
 At 50\% pruning the tilt restores 14--34\% of the language-modelling damage, a share stable across calibration draws and held-out corpora but different between models.

\textbf{Share of the damage recovered.}
\label{sec:ppl}

An operator's recovery is the share of the constant's damage it removes: $(\mathrm{NLL}_{\text{const}}-\mathrm{NLL}_{\text{op}})/(\mathrm{NLL}_{\text{const}}-\mathrm{NLL}_{\text{dense}})$ for language modelling, and the same ratio for nine-task accuracy. Perplexity is measured on WikiText and on two corpora no operator in this section was fitted on: 64 blocks of 4,096 tokens each of C4 and of Zyda-2 slice C.

\begin{table}[t]
\caption{Share of the constant's damage recovered at 50\% pruning (\%), MoNE masks. Tilt cells on C4 and Zyda-2 give calibration draws A / B; all other cells use draw A. Reroute: rerouting at the trained gate scale. Copies are the equal-budget copies of \cref{sec:copies}.}
\label{tab:recovery}
\centering\footnotesize
\setlength{\tabcolsep}{2pt}
\begin{tabular}{@{}llcccc@{}}
\toprule
Model & Operator & WikiText & C4 & Zyda-2 C & Acc. \\
\midrule
OLMoE & Tilt r32 & 18.3 & 23.7 / 23.8 & 22.3 / 22.5 & 19.5 \\
      & Reroute & 11.1 & 12.4 & 9.3 & 13.5 \\
      & Neuron copy & --- & 28.0 & 26.6 & --- \\
      & W-SVD copy & --- & 16.1 & 14.2 & --- \\
\midrule
DeepSeek & Tilt r32 & 19.0 & 13.9 / 14.3 & 14.1 / 14.1 & 6.6 \\
         & Reroute & 27.4 & 29.8 & 27.2 & 18.5 \\
         & Neuron copy & --- & 13.4 & 14.6 & --- \\
         & W-SVD copy & --- & −2.7 & −3.1 & --- \\
\midrule
Moonlight & Tilt r32 & 18.7 & 33.8 / 34.2 & 31.5 / 32.7 & 12.2 \\
          & Reroute & −4.7 & 18.1 & 15.2 & −1.6 \\
          & Neuron copy & --- & 26.3 & 27.4 & --- \\
          & W-SVD copy & --- & 15.3 & 19.0 & --- \\
\bottomrule
\end{tabular}
\end{table}

\textbf{Recovery is stable within a model and differs between models.} The tilt's share agrees within 1.2 points across calibration draws and within 2.3 points across the two held-out corpora, but it ranges from 14\% on \dsv{} to 34\% on Moonlight (\cref{tab:recovery}). The near-equal shares on WikiText are a property of that corpus. Its language-modelling recovery grows with the pruning ratio on every model (\cref{tab:ratio}): at 75\% the tilt lowers OLMoE's WikiText perplexity from 1,434 under the constant to 172, while rerouting reaches 1,280 (\cref{tab:ppl}).

\textbf{Perplexity orders the tilt and rerouting as accuracy does, but converts differently.} On every corpus the tilt recovers more than rerouting on OLMoE and Moonlight and less on \dsv{}, the same order as accuracy. In our runs perplexity ordered two operators correctly whenever both moved the same way from the constant, but it can misjudge whether an operator beats the constant at all (\cref{sec:pplacc}). How much accuracy a share of perplexity buys differs by model; it is largest on OLMoE, whose accuracy damage is largest (17.7 points, against 11.2 on Moonlight and 7.3 on \dsv{}).

\textbf{Where the error reduction comes from (OLMoE, 50\%).}
\label{sec:interact}
Per expert, the tilt is a weak approximator: measured as a held-out fit share in layers 0, 8 and 15 (\cref{sec:setup}), a full-rank linear map of the expert input removes 22--42\% of the constant's error and the rank-32 tilt 8.5--30.4\% (\cref{tab:B3}). Summed over calibration tokens, the tilt removes 28\% of the constant's gate-weighted residual error, about three quarters of it from each pruned expert's own residual and a quarter from residuals shared by pruned experts routed to the same token, on both calibration slices (\cref{tab:shared}). No decomposition of layer error can show what causes an accuracy gain, and HOPE's off-diagonal shares (products of norms under deletion with renormalized gates) are a different quantity \citep{tseng2026hope}.

\section{Which Routing Semantics Matter?}
\label{sec:renorm}

Three of our operators move a pruned slot's work onto the surviving experts: renormalized rerouting scales their gates up, slot remapping gives the pruned gate value to one of them, and COMPEL adds the pruned weights into all of them. On routers trained without top-$k$ renormalization, none of them beats keeping the survivors' trained scale and dropping the slot, and they lose to it in seven of the eight operator $\times$ model cases at 50\% (\cref{fig:gatescale}). Renormalization costs 0.8--8.5 points, lowers ranking ability, and changes two comparisons on these models: whether rerouting beats the constant, and, on \dsv{}, how REAP's mask under REAP's routing semantics compares with MoNE's. On Granite, whose router is trained with renormalization, the gate-scale result reverses at 50\%, and on Moonlight, also trained with it, at 50\% and 75\%, while remapping still loses (\cref{sec:granite}).

\subsection{Renormalized Gates}

\begin{table}[t]
\caption{Renormalized minus keep-scale rerouting, same mask (MoNE), 95\% intervals for the nine-task mean. Every nine-task row survives Holm within the table; all but \dsv{} 25\% also survive over all comparisons. At 75\% both operators score below chance on MMLU (22.9--23.7\%, chance 25\%), so the positive MMLU difference there does not indicate retained knowledge.}
\label{tab:renorm}
\centering\footnotesize
\setlength{\tabcolsep}{4pt}
\begin{tabular}{@{}lccc@{}}
\toprule
Setting & Nine-task & Big-5 & MMLU \\
\midrule
OLMoE 25\% & −4.50 [−5.48, −3.57]* & −5.59* & −2.08* \\
OLMoE 50\% & −3.39 [−4.66, −2.12]* & −4.31* & −2.04* \\
OLMoE 75\% & −3.09 [−4.52, −1.68]* & −4.83* & +0.73* \\
DeepSeek 25\% & −0.79 [−1.51, −0.07]* & −0.67* & −1.40* \\
DeepSeek 50\% & −4.67 [−5.71, −3.59]* & −6.28* & −7.35* \\
DeepSeek 75\% & −8.47 [−9.98, −7.01]* & −10.84* & +0.73* \\
\bottomrule
\end{tabular}
\end{table}

\textbf{Cost.} OLMoE and \dsv{} use raw softmax probabilities as gates, so the top-$k$ sum well below one. Masking pruned logits before the softmax multiplies every surviving gate by $1/(1-P_{\text{pruned}}(x))$, a rescaling the router was not trained with. Keeping the trained scale instead is better in all six settings (\cref{tab:renorm}), and on REAP's mask as well ($-$1.88 and $-$3.88; \cref{tab:reverse}). On \dsv{} the cost grows with the pruning ratio ($-$0.79, $-$4.67, $-$8.47); on OLMoE it is large already at 25\% and stays at 3--4.5 points.

\textbf{Renormalization lowers ranking ability.} On two-choice tasks with fixed answer strings, an operator can lose accuracy by shifting the bias between the two answers or by damaging the model's ability to rank them. The area under the ROC curve of the answer margin is unaffected by any shift, so it separates the two (\cref{sec:auc}). Renormalization lowers it: in all eight two-choice task $\times$ setting cells on OLMoE and \dsv{} at 25\% and 50\%, renormalized rerouting ranks items worse than keep-scale rerouting on the same mask, with an AUC lower by 0.03--0.14 (point estimates, \cref{tab:A7}). Against the constant the largest drops have intervals: on OLMoE 50\% RTE the AUC falls from 0.639 [0.570, 0.704] to 0.498 [0.428, 0.566], an interval that contains chance; BoolQ AUC falls by 0.137 [0.111, 0.163] (OLMoE 50\%) and 0.121 [0.107, 0.136] (\dsv{} 50\%). On OLMoE 25\% BoolQ about two thirds of the $-$10.2-point loss is recovered by re-thresholding ($-$3.30 [$-$5.47, $-$1.56] after it), but the AUC still falls by 0.061 [0.041, 0.081] (\cref{tab:aucci}).

\begin{table}[t]
\caption{Nine-task accuracy of rerouting under each gate scale, against MoNE's constant on its own mask. REAP's mask was evaluated at 50\% only; ``renorm'' on the REAP mask is REAP with its own routing semantics (REAP did not evaluate these models).}
\label{tab:reverse}
\centering\footnotesize
\setlength{\tabcolsep}{3.5pt}
\begin{tabular}{@{}lccccc@{}}
\toprule
 & \multicolumn{3}{c}{MoNE mask} & \multicolumn{2}{c}{REAP mask} \\
\cmidrule(lr){2-4}\cmidrule(l){5-6}
Setting & Constant & Renorm & Keep & Renorm & Keep \\
\midrule
OLMoE 25\% & 61.19 & 58.15 & 62.65 & --- & --- \\
OLMoE 50\% & 49.09 & 48.08 & 51.48 & 47.12 & 49.00 \\
DeepSeek 25\% & 65.40 & 64.87 & 65.66 & --- & --- \\
DeepSeek 50\% & 59.84 & 56.52 & 61.19 & 56.58 & 60.46 \\
\bottomrule
\end{tabular}
\end{table}

\textbf{Renormalization changes two conclusions.} With renormalized gates, rerouting scores below MoNE's constant in all six softmax settings (25--75\%; \cref{tab:main}), and REAP's mask under REAP's semantics scores 3.3 points below MoNE's constant model on \dsv{} (\cref{tab:reverse}). With the trained gate scale, rerouting beats the constant in five of the six settings, each surviving Holm correction over all comparisons (all but \dsv{} 25\%, +0.26 n.s.), and the same REAP mask scores 0.6 points above MoNE's constant model (a point estimate, not a tested contrast). REAP itself was not evaluated on these models, so this changes a comparison made with REAP's semantics, not a number REAP published. The same concern applies to published methods that renormalize on such routers: on MoRA's own \dsv{} masks, the trained scale gains 0.47 and 1.24 points over renormalized rerouting in our re-implementation (\cref{app:mora-semantics}).

\subsection{Slot Remapping and Weight Compensation}
\label{sec:others}

\begin{figure*}[t]
\centering
\includegraphics[width=\textwidth]{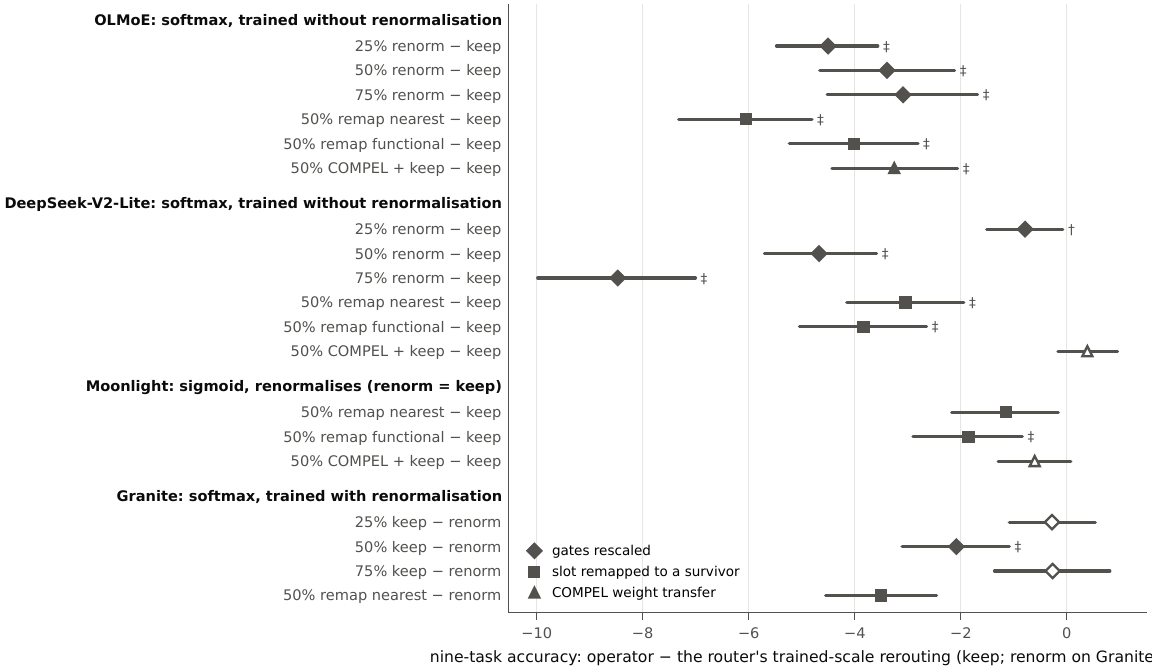}
\caption{Each operator that changes the survivors' gate scale or moves a pruned slot's work onto survivors, minus rerouting at the scale the router was trained with: keep-scale on OLMoE and \dsv{} (softmax, trained without top-$k$ renormalization) and on Moonlight (sigmoid, trained with top-$k$ renormalization: rerouting at its trained, renormalized scale; \cref{sec:routing}), renormalized on Granite (softmax, trained with renormalization; \cref{sec:granite}). Nine-task, MoNE masks, draw A, 95\% intervals; filled marker: the interval excludes zero; \holma{} / \holmt: survives Holm over all comparisons / within its table only; Granite's remap row is outside the Holm families. Rescaling changes the survivors' gates (renorm on the first two models, keep on Granite); remapping hands the slot to a survivor at the unchanged gate value; COMPEL moves pruned weights into the survivors.}
\label{fig:gatescale}
\end{figure*}

ConMoE remaps a pruned expert's routing slot to a retained expert and keeps the gate value \citep{yao2026conmoe}; COMPEL leaves the slot empty but adds a Fisher-weighted share of the pruned experts' weights into every survivor \citep{yoon2026compel}. We implemented both from their papers and applied them to MoNE's 50\% masks (\cref{fig:gatescale,tab:glance}; details and every contrast in \cref{app:others}).

\textbf{Remapping a slot to a surviving expert loses to keep-scale rerouting on every model.} On OLMoE, remapping to the nearest retained expert is 6.05 points below keep-scale rerouting and 3.66 below the constant; the functional choice recovers 2.04 of that but stays 4.01 below keep. On \dsv{} both rules are about 3--4 points below keep, and on Moonlight 1.2--1.9 points below it (the nearest rule's $-$1.15 is significant per comparison only); on these two models the functional choice is no better than the nearest one ($-$0.80 and $-$0.70, n.s.). On every model both rules are also below MoNE's constant (by 1.3--3.7 points, every interval excluding zero; \cref{tab:H8}). This agrees with ConMoE's own OLMoE results, in which MoNE scores above ConMoE at 25\% and 50\% \citep{yao2026conmoe}; our pool (MoNE's mask), calibration (general text) and routed states (the unpruned model) differ from ConMoE's, so we evaluate its assignment operator, not the complete method (\cref{app:reported}).

\textbf{Weight compensation and cross-expert proxies do not help either.} COMPEL's Fisher-weighted compensation, added to keep-scale rerouting, costs 3.26 points on OLMoE and changes \dsv{} and Moonlight by +0.39 and $-$0.61 (n.s.). MoRA's channel-wise proxies find almost nothing to fit under our implementation: on \dsv{} and Moonlight at most 3 of 832 pruned experts have a retained expert that passes MoRA's own threshold, and the approximation changes accuracy by at most +0.19 points (n.s.). Settings, checks and every contrast are in \cref{app:others,app:mora}.

\textbf{Both survivor results hold on a second calibration draw.} Refitting COMPEL's Fisher weights and the functional assignment on slice B for OLMoE gives $-$3.13 and $-$4.24 points against keep-scale rerouting (slice A: $-$3.26 and $-$4.01), each interval excluding zero; the change between draws is at most 0.23 points (\cref{tab:drawsB}).

\subsection{Routers Trained with Renormalization}
\label{sec:granite}

\textbf{Why a fourth model.} On both models without trained renormalization, the operators that hand the slot to survivors lose to the trained scale by 3.0--6.1 points in every case but one, COMPEL on \dsv{}, which is not detectably different (\cref{fig:gatescale}). These routers were never trained with their survivors taking on a slot's work. If the cost comes from moving the survivors away from the gate scale they were trained with, a router trained with renormalization should show the opposite sign. Granite-3.0-3B-A800M has a softmax router like OLMoE's that normalizes over the selected top-$k$, so the gates it was trained with always sum to one. We converted its MoE blocks to OLMoE's block with renormalization switched on, checked that the conversion is exact (next-token loss 2.082205 nats before and after in float32, top-1 agreement 1.000), and ran MoNE's masks with the constant, the tilt and both rerouting scales at 25--75\% and nearest remapping at 50\%, under the paper's protocol (\cref{tab:granite}; \cref{app:granite}). All Granite comparisons use one calibration draw; implementation checks and further contrasts are in \cref{app:granite}.

\begin{table}[t]
\caption{Granite-3.0-3B-A800M (softmax router trained with top-$k$ renormalization), MoNE masks, draw A. Top: nine-task accuracy (dense 61.82; guessing about 40.5); bold: highest in the column. Bottom: contrasts; * the 95\% interval excludes zero; \holma: survives Holm over all comparisons (the other contrasts are outside the Holm families). Every interval, with big-5 and MMLU, is in \cref{tab:granitefull}.}
\label{tab:granite}
\centering\footnotesize
\setlength{\tabcolsep}{5pt}
\begin{tabular}{@{}lccc@{}}
\toprule
Pruned & 25\% & 50\% & 75\% \\
\midrule
Constant & 57.95 & 45.27 & 41.30 \\
Tilt r32 & \textbf{59.24} & \textbf{48.64} & \textbf{42.88} \\
Keep (other scale) & 57.11 & 44.65 & 40.87 \\
Renorm (trained scale) & 57.39 & 46.74 & 41.14 \\
Remap nearest & --- & 43.23 & --- \\
\midrule
Renorm $-$ keep & +0.28 & +2.08*\holma & +0.27 \\
Tilt $-$ constant & +1.29* & +3.36* & +1.58* \\
Tilt $-$ renorm & +1.85* & +1.90* & +1.75* \\
Remap $-$ renorm & --- & $-$3.51* & --- \\
\bottomrule
\end{tabular}
\end{table}

\textbf{The gate-scale result reverses.} At 50\% renormalized rerouting beats keep-scale rerouting by +2.08 [+1.08, +3.11]\holma{} points (big-5 +2.46*), and it is ahead on all nine tasks, where OLMoE and \dsv{} lose 3.39 and 4.67 points by renormalizing. At 25\% and 75\% the two scales are not detectably different (+0.28 [$-$0.54, +1.09] and +0.27 [$-$0.81, +1.36]; the data rule out differences larger than 0.96 and 1.19 points), so the reversal is established at 50\% only.

\textbf{Remapping still loses.} Nearest remapping keeps every gate value the router computes, so on Granite the gates still sum to one, as in training. Against renormalized rerouting it scores $-$3.51 [$-$4.55, $-$2.46], against keep-scale rerouting $-$1.43 [$-$2.34, $-$0.51] and against the constant $-$2.04 [$-$3.01, $-$1.12].

\textbf{With each router at its trained scale, Granite orders the operators as OLMoE does.} At 50\% the order is the tilt, trained-scale rerouting, the constant, rerouting at the other scale and nearest remapping on both models (Granite 48.64, 46.74, 45.27, 44.65, 43.23; OLMoE 52.54, 51.48, 49.09, 48.08, 45.43), although which rerouting arm is ``trained'' is opposite on the two.

Granite therefore splits the explanation in two. Rescaling the survivors' gates is a matter of gate scale and follows how the router was trained. Remapping does not change the scale and costs on every router. This fits what our MoRA re-implementation shows on \dsv{} and Moonlight, where almost no retained expert predicts a pruned expert's output under a channel-wise affine map (\cref{sec:mora}), while remapping substitutes such an output at the full gate. Remapping costs least on the two models with shared experts (\dsv{} 3.04 and Moonlight 1.15 points, against 6.05 on OLMoE and 3.51 on Granite), which we note as a hypothesis (\cref{app:hypotheses}).

\textbf{A second router trained with renormalization: Moonlight.} Moonlight's sigmoid router also normalizes over the selected top-$k$, so keeping the unpruned normalizer (\cref{sec:routing}) tests the reverse direction a second time, on a larger model with a different gate function and two shared experts.  Keeping the unpruned normalizer is a real intervention: the unpruned top-$k$ holds a pruned expert for 55\%, 97\% and 100\% of token--gate calls at 25\%, 50\% and 75\%, and keep's routed gate mass is on average 0.96, 0.71 and 0.42 of the trained arm's. Because the bias steers the choice but not the gate, keep gives \emph{more} mass than the trained arm on 9.9\%, 5.2\% and 0.8\% of calls (\cref{sec:routing}). The reverse direction holds (\cref{tab:moonkeep}). The trained scale beats keep by +1.68 [+0.84, +2.52] at 50\% and +1.60 [+0.36, +2.85] at 75\%. Both survive Holm within their family, and the 50\% contrast also survives over all 111 comparisons. At 25\% the two are not detectably different ($-$0.17 [$-$0.46, +0.11]); keep is ahead on MMLU alone (+1.59 [+1.01, +2.16]), and big-5 shows no detectable difference (+0.02 [$-$0.34, +0.39]). The other rows of \cref{tab:moonkeep}, outside the Holm families, point the same way: at 50\% keep falls 1.86 points below the constant, and the tilt's lead over keep (+3.23) is about twice its lead over the trained scale (+1.55). Unlike Granite, Moonlight keeps the reversal at 75\%, where it is further from guessing (45.5 against a guessing level of 40.5; every Granite model at 75\% is within 2.4 points of it).

\begin{table}[t]
\caption{Moonlight-16B-A3B (sigmoid router trained with top-$k$ renormalization), MoNE masks, draw A, nine-task accuracy: rerouting at the trained scale (renormalized over the selected survivors) against keeping the unpruned normalizer. Contrasts with 95\% intervals in the text; * the interval excludes zero; \holmt{} / \holma: survives Holm within the renorm $-$ keep family / also over all 111 comparisons (the other rows are outside the Holm families).}
\label{tab:moonkeep}
\centering\footnotesize
\setlength{\tabcolsep}{5pt}
\begin{adjustbox}{max width=\columnwidth}\begin{tabular}{@{}lccc@{}}
\toprule
Pruned & 25\% & 50\% & 75\% \\
\midrule
Constant & 68.66 & 60.24 & 43.89 \\
Renorm (trained scale) & 68.71 & 60.06 & 45.48 \\
Keep (unpruned normalizer) & 68.88 & 58.38 & 43.88 \\
\midrule
Renorm $-$ keep & $-$0.17 & +1.68*\holma & +1.60*\holmt \\
Keep $-$ constant & +0.22 & $-$1.86* & $-$0.01 \\
Tilt $-$ keep & $-$0.09 & +3.23* & +3.28* \\
\bottomrule
\end{tabular}\end{adjustbox}
\end{table}

\textbf{The pattern: the trained-scale rule.} On all four routers at 50\% pruning, the surviving gates do best at the scale the router was trained with: keep-scale rerouting on OLMoE and \dsv{}, renormalized rerouting on Granite and Moonlight (\cref{tab:teaser}). At that ratio the wrong scale scores below MoNE's constant on all four routers (point estimates $-$1.01 OLMoE, $-$3.32 \dsv{}, $-$0.62 Granite, $-$1.86 Moonlight), while the trained scale is not detectably different from it or above it. On the two routers trained without renormalization the trained scale wins at every ratio; on the two trained with it, the difference is not detectable at 25\%, nor on Granite at 75\%. This is the expert-removal counterpart of the reference-set principle \citet{chen2026halving} state for top-$k$ reduction. REAP and HOPE derive their error decompositions for renormalizing routers and treat the rescaling as a small term \citep{lasby2026reap,tseng2026hope}; on such routers renormalizing is indeed the trained behavior. Remapping a slot to a survivor is not a substitute for either scale on any of the four models.

Three limits apply. We compare the two endpoint scales only, whereas \citet{chen2026halving} find interior normalizers best for top-$k$ reduction. The gate-scale contrasts are measured on multiple choice and perplexity, not on generation. And the reverse direction rests on less: two routers with one calibration draw each. Granite differs from OLMoE and \dsv{} in size, expert count and shared experts as well as in router training, and at 50\% its rerouted models are only 4--6 points above guessing; Moonlight, which shares \dsv{}'s expert layout, gives the reverse direction a second router at 50\% and 75\%.

\subsection{Across Replacement Families}
\label{sec:glance}
\begin{table*}[t]
\caption{Representative replacement operators at 50\% expert pruning, on MoNE masks and the same Zyda-2 calibration. Values are mean accuracy over nine zero-shot tasks (\%). Trained-scale reroute: keep-scale on OLMoE and DeepSeek-V2-Lite, renormalized on Moonlight. The tilt and the neuron copy add similar parameter budgets.}
\label{tab:glance}
\centering\small
\begin{tabular}{@{}lrrrrr@{}}
\toprule
Model & Dense & Constant & Tilt r32 & Trained-scale reroute & Neuron copy \\
\midrule
OLMoE & 66.76 & 49.09 & 52.54 & 51.48 & 52.63 \\
DeepSeek-V2-Lite & 67.15 & 59.84 & 60.32 & 61.19 & 60.18 \\
Moonlight & 71.48 & 60.24 & 61.61 & 60.06 & 60.94 \\
\bottomrule
\end{tabular}
\end{table*}
The comparison across replacement families supports two conclusions. First, no replacement is best on all models: on \dsv{} parameter-free trained-scale rerouting beats the tilt at 50\% pruning; the tilt beats rerouting on OLMoE and Moonlight, and is the highest-scoring operator on Moonlight. Second, two budgeted replacements work, the tilt and a partial copy of the pruned expert's top neurons; they are not detectably different on OLMoE (where the copy scores 0.09 points higher) and \dsv{}, and the tilt is ahead on Moonlight. Survivor-based operators do not reach trained-scale rerouting on any model (\cref{sec:others,fig:gatescale}).

\section{Does the Answer Survive Realistic Evaluation?}
\label{sec:eval}

Multiple-choice accuracy, the standard metric for training-free MoE pruning, hides most of the generative damage and ranks operators differently from generation; on fixed-answer tasks it also mixes calibration with ability. The calibration domain moves the mask far more than resampling. At matched storage, uniform 3-bit expert quantization is stronger on general tasks, while in-domain pruning is stronger on GSM8K, and the tilt's gain survives quantization.

\subsection{Multiple Choice against Generation}

\begin{figure*}[t]
\centering
\includegraphics[width=\textwidth]{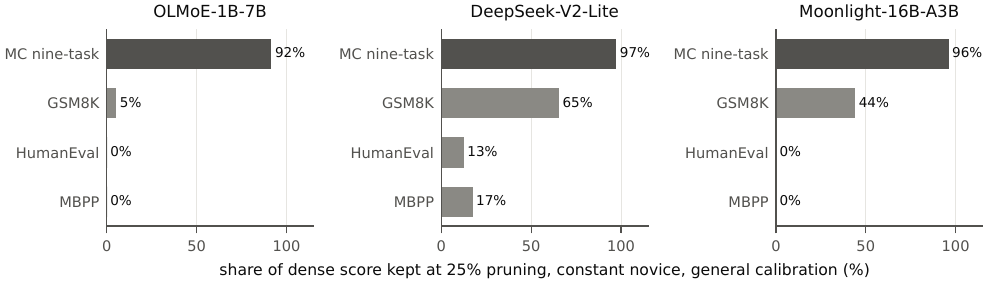}
\caption{Share of the dense model's score kept at 25\% pruning with MoNE's constant and general calibration: nine-task multiple choice, GSM8K (5-shot, strict match), HumanEval (pass@1) and MBPP (pass@1, 3-shot). Multiple-choice tasks have chance floors and generation tasks near-zero floors, so the bars are not comparable fractions of ability; above chance, multiple choice keeps 79\%, 93\% and 91\% (\cref{tab:chance}).}
\label{fig:generation}
\end{figure*}

\textbf{Multiple choice hides most of the generative damage.} At 25\% pruning every model keeps more than 90\% of its multiple-choice accuracy (79--93\% of its margin above chance; \cref{tab:chance}), yet OLMoE keeps 5\% of its GSM8K score and none of its HumanEval score (\cref{fig:generation}); its GSM8K collapse matches MoNE's own appendix. At 50\% the constant keeps about 10\% of dense GSM8K at most, and every operator scores zero on HumanEval. MBPP, with three times as many problems as HumanEval, shows code collapsing at 25\% on all three models under every operator: OLMoE and Moonlight solve at most 1.2\% of its 500 problems (dense 21.0\% and 52.0\%), and \dsv{} keeps 15--21\% of its dense 43.6\% (\cref{tab:mbpp}). This is not an extraction artifact: the pruned models write functions, cutting each generation at the end of its answer changes no OLMoE or Moonlight score, and on OLMoE and Moonlight 39--100\% of the pruned programs we classified are not valid Python, whereas the dense Moonlight model fails mostly on wrong answers (\cref{app:mbpp}). The operators break code differently: none of the programs from Moonlight's constant is valid Python, while rerouting keeps the syntax but invents APIs and signatures. Consistent with REAP's finding that code needs in-domain calibration \citep{lasby2026reap}, no operator preserves it under general calibration.

\begin{table*}[t]
\caption{GSM8K (5-shot, strict match) absolute scores and gain over the constant on the same mask, general calibration, 95\% item-bootstrap intervals. Dense: OLMoE 45.49, DeepSeek 37.53, Moonlight 73.84. At 50\% every model keeps at most 17\% of its dense score under every operator, so differences there cannot rank operators; OLMoE 50\% scores 0 under both the constant and the tilt. Reroute: rerouting at the trained gate scale. \holmt/\holma: survives Holm within the table / over all comparisons.}
\label{tab:gsm8k}
\centering\footnotesize
\setlength{\tabcolsep}{4pt}
\begin{tabular}{@{}lccccc@{}}
\toprule
Setting & Constant & Tilt & Reroute & Tilt $-$ constant & Reroute $-$ constant \\
\midrule
OLMoE 25\% & 2.43 & 1.82 & 2.73 & −0.61 [−1.59, +0.38] & +0.30 [−0.76, +1.29] \\
OLMoE 50\% & 0.00 & 0.00 & 0.68 & +0.00 [+0.00, +0.00] & +0.68 [+0.30, +1.14]*\holma \\
DeepSeek 25\% & 24.56 & 23.12 & 23.81 & −1.44 [−3.11, +0.30] & −0.76 [−2.58, +1.06] \\
DeepSeek 50\% & 3.79 & 4.62 & 6.37 & +0.83 [−0.53, +2.20] & +2.58 [+1.21, +4.02]*\holma \\
Moonlight 25\% & 32.60 & 32.75 & 42.76 & +0.15 [−1.97, +2.35] & +10.16 [+7.66, +12.66]*\holma \\
Moonlight 50\% & 0.38 & 1.36 & 1.52 & +0.99 [+0.30, +1.67]*\holmt & +1.14 [+0.38, +1.90]*\holmt \\
\bottomrule
\end{tabular}
\end{table*}

\textbf{Generation ranks the operators differently.} On zero-shot multiple choice at 50\% the tilt is Moonlight's best operator and trained-scale rerouting scores below the constant; on GSM8K at 25\% rerouting gains ten points on Moonlight and the tilt nothing (\cref{tab:gsm8k}), although on multiple choice the two are not detectably different there (+0.08 [$-$0.28, +0.43]). At 50\% generation has collapsed on every model, so differences there, some of a few items (OLMoE's +0.68 is nine of 1,319), cannot rank operators. On MBPP, where only \dsv{} keeps code ability at 25\%, keep-scale rerouting is again the best operator and the tilt the worst ($-$2.80 [$-$5.20, $-$0.40] against keep; \cref{tab:mbpp}). An operator chosen on multiple choice can be the wrong one for generation. Under general-text calibration, at 25\%, where generation has not collapsed, we find no evidence that the tilt preserves GSM8K or MBPP; with in-domain masks the operator matters again (\cref{sec:domain}).

\subsection{Calibration versus Ability on Fixed-Answer Tasks}
\label{sec:auc}

BoolQ and RTE use the same two answer strings for every item, so an operator can lose accuracy by shifting the model's bias between them without changing how well it ranks items. The AUC of the log-likelihood margin is threshold-free and separates the two; prior-matched accuracy re-thresholds so the predicted label shares equal the gold ones (\cref{tab:aucci}). On OLMoE and Moonlight at 50\%, the tilt's RTE losses ($-$2.89 and $-$3.25) leave the AUC unchanged ($-$0.008 [$-$0.064, +0.047] and +0.001 [$-$0.028, +0.030]), so they are shifts in answer bias; its OLMoE BoolQ gain raises the AUC by 0.089 [0.074, 0.104]; and Moonlight's rerouting loss on BoolQ lowers it by 0.094 [0.080, 0.108].

The constant model itself is badly miscalibrated: on OLMoE 50\% RTE it predicts ``True'' on 96\% of items (gold 53\%) yet ranks them above chance. The same holds for four-way MMLU: Moonlight's 50\%-pruned model answers ``A'' on 93\% of items, and removing each letter's mean preference raises the tilt's MMLU from 28.7\% to 37.7\%, against 28.1\% for the constant after the same correction, so the tilt retains knowledge that raw accuracy hides (\cref{app:pertask}). Reported zero-shot accuracy at high pruning ratios therefore understates what the model can still do and can move with an operator's answer bias alone.

\subsection{Perplexity versus Accuracy}
\label{sec:pplacc}

Perplexity orders two operators as accuracy does when both move the same way from the constant: the tilt against rerouting on every model, and keep-scale against renormalized rerouting (\cref{sec:ppl,tab:A5}). It is not a safe test of whether an operator beats the constant at all. Renormalized rerouting on \dsv{} lowers perplexity relative to the constant (22.74 → 19.70) but adds 45\% to its accuracy damage (\cref{sec:renorm}), and on Moonlight rerouting at the trained scale worsens WikiText perplexity but improves C4 and Zyda-2 perplexity (\cref{sec:ppl}). \citet{chen2026halving} report the same disagreement for top-$k$ reduction, and \citet{xu2026ucramp} for micro-expert pruning.

\subsection{Calibration Domain}
\label{sec:domain}

The calibration text moves the pruned set far more than resampling does. Changing it changes both which experts are removed and what the fitted operators are fitted on, so the multiple-choice comparisons below do not by themselves separate the two. Results in this section use one GSM8K calibration draw; intervals are item-level only.

\begin{table}[t]
\caption{Share of pruned experts that the GSM8K-calibrated and general (Zyda-2) masks have in common (chance: 50\% and 25\%). Last column: two disjoint general draws, both masks from MoNE's released code.}
\label{tab:overlap}
\centering\footnotesize
\setlength{\tabcolsep}{4pt}
\begin{tabular}{@{}lccc@{}}
\toprule
 & \multicolumn{2}{c}{GSM8K vs general} & General A vs B \\
\cmidrule(lr){2-3}\cmidrule(l){4-4}
Model & 50\% & 25\% & 50\% \\
\midrule
OLMoE & 65.2\% & 44.1\% & 93.6\% \\
DeepSeek & 70.3\% & 50.0\% & 95.8\% \\
Moonlight & 66.1\% & 60.8\% & 95.8\% \\
\bottomrule
\end{tabular}
\end{table}

\textbf{How far the mask moves.} We re-ran MoNE with GSM8K-train text as calibration and compared its masks with the general-text masks (\cref{tab:overlap}). Changing the domain swaps about a third of the pruned set at 50\% and 39--56\% of it at 25\%. At 50\% that is five to eight times the change between two disjoint general-text draws (34.8\% of pruned experts against 6.4\% on OLMoE, 29.7\% against 4.2\% on \dsv{}, 33.9\% against 4.2\% on Moonlight). This does not contradict \citet{salim2026calib}, who find for domain-general pruning that diverse, pretraining-like calibration text can match test-distribution data: our general masks are of that kind, and the shift we measure is what task-specific calibration does to them. Even two general corpora are not interchangeable: on \dsv{} at 50\%, MoNE's pipeline run on C4 and on Zyda-2 prunes 86.2\% of the same experts, yet MMLU with constants is 22.95 (chance) after C4 calibration against 40.98 after Zyda-2 (\cref{app:mora-tasks}). The general-text masks use about 4.1\,M calibration tokens and the GSM8K masks 1.2--1.4\,M, so the ratio above mixes a change of domain with a change of budget. The budget is not what moves the mask. With both general-text masks cut to the first $N$ blocks, where $N$ is each model's GSM8K block count (286, 346, 295), the domain still moves the pruned set 4.9, 5.9 and 5.9 times as much as resampling at 50\% (OLMoE, \dsv{}, Moonlight), and 4.9, 9.3 and 3.9 times at 25\%; cutting slice A from 1,000 blocks to $N$ moves only 2.6--4.9\% of the pruned set at 50\% (\cref{tab:budgetfull}). The mask-domain effect remains more than three times the resampling effect on each model at both ratios. A likely mechanism is routing concentration: GSM8K routes tokens to fewer experts, and MoNE's score carries routing frequency, so it prunes the experts GSM8K leaves idle (\cref{app:hypotheses}).

\textbf{In-domain calibration rescues generation.} With GSM8K-calibrated masks, 25\% pruning keeps essentially all of dense GSM8K (46.3, 37.8 and 73.8 against dense 45.5, 37.5 and 73.8), extending REAP's finding for code to mathematics on MoNE's masks. Once the mask is in-domain but damage remains, the operator matters a great deal: on OLMoE at 50\% the tilt adds +19.4 [+16.5, +22.2] and keep-scale rerouting +22.7 [+19.9, +25.4] GSM8K points over the constant (21.6 → 41.0 and 44.3; dense 45.5). Smaller remaining damage can matter too: on Moonlight at 50\% the in-domain constant is 3.5 points below dense (70.4 against 73.8), and the tilt recovers +2.96 [+1.06, +4.85] of them (to 73.3) while trained-scale rerouting's +0.83 is not significant. Elsewhere in-domain operator differences are inside their intervals (\cref{tab:A6b}); these comparisons are per comparison and outside the Holm families.

\begin{figure*}[t]
\centering
\includegraphics[width=\textwidth]{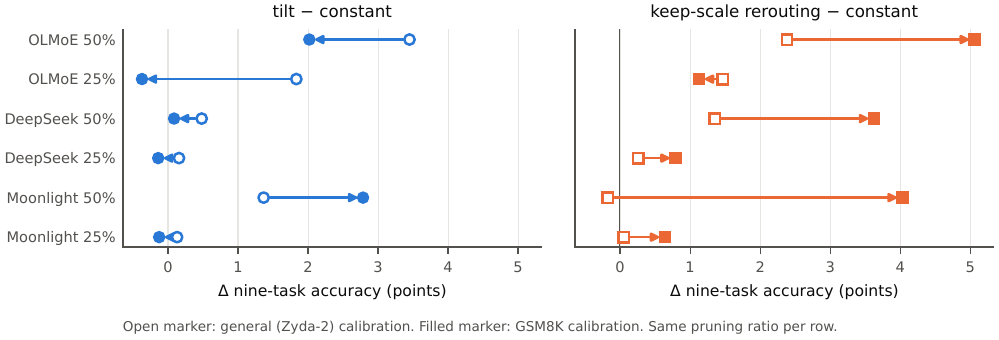}
\caption{Nine-task gain over the constant on the same mask, with general (Zyda-2) and GSM8K calibration, for the tilt and rerouting at the trained gate scale (renormalized on Moonlight; \cref{sec:routing}). One GSM8K calibration draw.}
\label{fig:domain}
\end{figure*}

\textbf{With GSM8K-calibrated masks, rerouting gains and the tilt's advantage shrinks.} With GSM8K-calibrated masks, rerouting is the best multiple-choice operator in all six settings, and on the 50\% masks its gain grows by 2.3--4.2 points (\cref{fig:domain}). The tilt's advantage shrinks or reverses everywhere except Moonlight at 50\%.  Rerouting inherits the domain too, through its mask; it simply has nothing else fitted. Crossing them on OLMoE at 50\% separates the two (\cref{tab:domainx}). The price on multiple choice depends on the model: with the constant, \dsv{} loses multiple-choice accuracy and MMLU at both ratios, while OLMoE and Moonlight lose commonsense but gain MMLU strongly at 50\% (\cref{tab:taskcal}). One configuration serves both kinds of task: OLMoE pruned by half with a GSM8K-calibrated mask and keep-scale rerouting scores 44.28 on GSM8K, 97\% of dense (45.49), and 54.00 on multiple choice (dense 66.76), with no fitted parameters.

\begin{table}[t]
\caption{Calibration domain crossed, OLMoE 50\%, nine-task accuracy: rows choose the mask, columns fit the operator. Diagonal cells are the runs of \cref{fig:domain}; keep-scale rerouting depends on the mask only. Fit effect = GSM8K $-$ general fit on the same mask; mask effect = GSM8K $-$ general mask with the same fit; * the paired 95\% interval excludes zero; \holmt{} / \holma: survives Holm within the four fit-effect rows / also over all 111 comparisons (the token-matched rows are outside the Holm families; \cref{app:holm}).}
\label{tab:domainx}
\centering\footnotesize
\setlength{\tabcolsep}{3pt}
\begin{adjustbox}{max width=\columnwidth}\begin{tabular}{@{}llccc@{}}
\toprule
Operator & Mask & Fit: general & Fit: GSM8K & Fit effect \\
\midrule
\multicolumn{5}{@{}l}{\emph{General text: Zyda-2 slice A, 1,000 blocks (4.1\,M tokens)}} \\
Constant & general & 49.09 & 48.12 & $-$0.98*\holmt \\
 & GSM8K & 51.74 & 48.93 & $-$2.80*\holma \\
 & \emph{mask effect} & +2.64* & +0.82 & \\
Tilt r32 & general & 52.54 & 51.03 & $-$1.52*\holmt \\
 & GSM8K & 53.81 & 50.95 & $-$2.86*\holma \\
 & \emph{mask effect} & +1.27 & $-$0.07 & \\
Keep & general / GSM8K & \multicolumn{2}{c}{51.48 / 54.00 (mask effect +2.52*)} & --- \\
\midrule
\multicolumn{5}{@{}l}{\emph{General text cut to the GSM8K budget: first 286 blocks of slice A, for mask and fits}} \\
Constant & general & 49.54 & 47.60 & $-$1.94* \\
 & GSM8K & 51.96 & 48.93 & $-$3.03* \\
Tilt r32 & general & 52.45 & 51.42 & $-$1.03* \\
 & GSM8K & 53.43 & 50.95 & $-$2.47* \\
Keep & general / GSM8K & \multicolumn{2}{c}{52.14 / 54.00 (mask effect +1.86*)} & --- \\
\bottomrule
\end{tabular}\end{adjustbox}
\end{table}
\textbf{Crossing the mask and the fit.} Holding the mask, fitting on GSM8K lowers nine-task accuracy for both fitted operators on both masks, by 0.98 to 2.86 points (every interval excluding zero; \cref{tab:domainx}): fitted operators inherit the domain of their fit. Holding the fit at general text, the GSM8K mask raises nine-task accuracy for every operator (+2.64, +1.27 and +2.52 [+1.09, +3.92] for the constant, the tilt and keep-scale rerouting; the tilt's +1.27 is not significant), with MMLU gains outweighing commonsense losses (\cref{tab:taskcal}). The tilt's gain over the constant shrinks mainly with the mask (+3.45 [+2.62, +4.27] on the general mask to +2.07 [+1.05, +3.14] on the GSM8K mask, both with general fits) and less with the fit (+2.91 [+2.00, +3.79] with GSM8K fits on the general mask). The best fitted cell combines the two calibrations: the GSM8K mask with general-text fits (53.81 for the tilt), 0.19 points below keep-scale rerouting on the same mask (54.00), which fits nothing. The constant's fit effect is smaller than the tilt's on the general mask ($-$0.98 against $-$1.52) but not on the GSM8K mask ($-$2.80 against $-$2.86). The general-text fits use about 4.1\,M tokens and the GSM8K fits 1.2\,M, but the budget does not explain the fit effect: with the general text cut to GSM8K's 286 blocks for both mask and fits, GSM8K fits still lower nine-task accuracy by 1.03 to 3.03 points (every interval excluding zero), and the tilt's gain over the constant still shrinks with the mask (+2.92 [+1.84, +3.95] to +1.47 [+0.50, +2.49]) and not with the fit (+3.82 [+2.81, +4.84] with GSM8K fits). One GSM8K draw; OLMoE at 50\% only.

\subsection{Pruning with Quantization at Matched Storage}
\label{sec:quant}
\textbf{The tilt's gain survives quantization; rerouting's erodes.} Expert pruning composes with quantization. Quantizing the pruned models with GPTQ (every decoder linear at 4 or 3 bits; \cref{app:F2}) does not detectably change the tilt's gain over the constant: the interaction, the gain under quantization minus the gain at 16 bits, is not detectable in 11 of 12 model $\times$ ratio $\times$ bit-width settings, with model means of $-$0.16, +0.08 and $-$0.05 points. Rerouting's gain falls in 11 of the 12 settings, significantly in 5 (\cref{tab:F3}). A candidate mechanism, which we did not test, is that rerouting sends every slot through quantized experts while the constant's and the tilt's slots stay at 16 bits.

\textbf{In-domain deployment.} At about 7\,GB, with masks, fits and the quantizer all calibrated on GSM8K, every 25\%-pruned arm with 4-bit weights scores above quantizing every expert to 3 bits (every other linear at 4 bits) on GSM8K, in point estimates on both models. On \dsv{} the tilt is ahead by +4.25* (37.15 against 32.90, 99\% of dense), and also above a Mixture-Compressor-style 2/3/4-bit allocation (35.63; point estimates). On Moonlight trained-scale rerouting is ahead by +2.58*, but there 99--132 rarely routed linears of the dense competitors fell back to round-to-nearest for lack of GSM8K tokens, which may handicap them (\cref{tab:F5}). These two arms keep 98--99\% of dense GSM8K at 23--24\% of the 16-bit size. The quantizer's own calibration domain is the largest single effect (+8.6* and +22.7* for dense 3-bit weights).

\textbf{General tasks.} With general-text calibration, at the matched storage we tested (25\% pruning with 4-bit weights against 3-bit experts with 4-bit attention and shared experts), uniform quantization scores above every pruned arm, by 1.33 and 1.72 nine-task points for the best one (the tilt) on OLMoE and Moonlight; on \dsv{} the nine-task difference is not detectable but big-5 and MMLU are lower (\cref{tab:F4}). The deficit is mostly MMLU, matching the knowledge that general-text pruning removes (\cref{tab:A1}) and MoEXBench's finding, across ten MoE models of 30--235B parameters, that expert pruning is the dominant degradation source when combined with quantization and KV-cache compression \citep{benazir2026moexbench}. At the lowest budget we ran, the comparison changes: uniform 2-bit GPTQ of every linear layer falls to 39.2--45.7 nine-task points (guessing is about 40.5), while 50\% pruning with 4-bit weights at most 5\% larger keeps 48.7--61.2; part of that margin is attention precision, and we did not run a 2-bit-expert arm with 4-bit attention (\cref{tab:F2}). Pruning is therefore most attractive for a known deployment domain or a very small memory budget, which is also where the choice of replacement operator matters most (\cref{sec:domain}). Our operator comparison is a controlled comparison at fixed masks, not a claim that pruning is the better compression.

\section{Discussion and Limitations}
\label{sec:discussion}

What fills a pruned expert's slot should be chosen as deliberately as which experts are pruned, and it should be judged on the task the model will serve.

\textbf{Practical interpretation.} Within the tested configurations: when rerouting, use the gate scale the router was trained with, which costs nothing and can change conclusions; a mean-output constant is competitive at light pruning on \dsv{} and Moonlight; a budgeted replacement, the tilted novice or a partial copy of the pruned expert, recovers part of the accuracy damage when pruning is aggressive, and the tilt also keeps most of pruning's batch-32 prefill saving and survives quantization; and trained-scale rerouting is the strongest parameter-free replacement, at the price of most of the prefill saving. Which replacement is best depends on the model, its router convention and the target task, so these are observations rather than a universal selection policy. For general tasks at a fixed memory budget, quantizing every expert to 3 bits was stronger than 25\% pruning with 4-bit weights; pruning paid off for a known deployment domain and at the smallest budget we tested.

\textbf{A reporting checklist for expert pruning.} Our results suggest that every expert-pruning comparison report:
\begin{enumerate}
\item the router's training convention and the gate semantics used after pruning, that is, which scale the selected survivors receive;
\item replacement-operator comparisons on a fixed mask, separately from selection;
\item at least one generative task next to multiple choice;
\item the calibration domain and token budget, for the mask and for any fitted operator;
\item a quantization baseline at matched storage.
\end{enumerate}

\begin{table}[t]
\caption{Weights and HuggingFace eager timings relative to the dense model at 50\% pruning (prompt 512 tokens, 64 decoded). Prefill at batch 32; decode per token at batch 1. ``Kept'': share of MoNE's prefill saving kept. Timings compare operators within a model; they are not serving-stack times. Weights are measured sizes; on \dsv{} they exceed two bytes per parameter (\cref{tab:B2}; 31.41\,GB dense, as in \cref{tab:F5}) by a constant 1.14\,GB in every row, which does not affect comparisons between operators. Batch 1 and 8: \cref{tab:F1}.}
\label{tab:cost}
\centering\footnotesize
\setlength{\tabcolsep}{3pt}
\begin{tabular}{@{}lcccc@{}}
\toprule
Operator & Weights (GB) & Prefill & Kept & Decode \\
\midrule
\multicolumn{5}{@{}l}{\emph{OLMoE}} \\
Dense & 13.84 & 1.000 & --- & 1.000 \\
Constant (MoNE) & 7.40 & 0.752 & 100\% & 0.885 \\
Tilt r32 & 7.53 & 0.837 & 66\% & 1.076 \\
Keep reroute & 7.40 & 0.982 & 7\% & 0.946 \\
\midrule
\multicolumn{5}{@{}l}{\emph{\dsv{}}} \\
Dense & 32.55 & 1.000 & --- & 1.000 \\
Constant (MoNE) & 18.15 & 0.838 & 100\% & 0.969 \\
Tilt r32 & 18.37 & 0.880 & 74\% & 1.062 \\
Keep reroute & 18.16 & 0.992 & 5\% & 1.093 \\
\bottomrule
\end{tabular}
\end{table}

\textbf{Cost: at batch 32 the tilt keeps most of pruning's prefill speed-up; rerouting gives it up.} All operators here save memory; none is a latency method as implemented. The tilt adds 0.13\,GB of weights on OLMoE and 0.22\,GB on \dsv{} at 50\% and keeps two thirds to three quarters of MoNE's batch-32 prefill saving; keep-scale rerouting stores nothing but keeps almost none of it, because every token again runs $k$ real experts. At batch 1, where even prefill is bound by kernel launches, the order reverses: rerouting keeps 85--92\% of the saving and the tilt's prefill is no faster than the dense model's (\cref{tab:F1}). In our unfused implementation the tilt also decodes 6--8\% slower per token than the dense model at batch 1, and rank 8 costs the same as rank 32, so the overhead is kernel launches, not arithmetic; we have not tested a fused kernel.

\textbf{Limitations.}
\emph{Models.} Three base models of 7--16B parameters with 64 routed experts, plus Granite-3.0-3B-A800M (40 experts) for the gate-scale test; models with 128 or more experts and instruction-tuned checkpoints are not covered. The reverse direction of the trained-scale rule rests on two routers, Granite (established at 50\%) and Moonlight (50\% and 75\%), each with one calibration draw and without the mask grid or generation for the keep arm, and we compare only the two endpoint gate scales. External results on Qwen3-30B-A3B suggest that the constant remains competitive there: ConMoE reports MoNE 0.3 points behind its method at 50\% \citep{yao2026conmoe}. Our protocol follows MoNE's (base models, nine multiple-choice tasks, GSM8K and code), so our absolute numbers are not comparable with work on instruct and larger checkpoints \citep{liu2026aimer,markovic2026optep}.
\emph{Training.} Everything is training-free; whether a short recovery fine-tune or router-only distillation \citep{hyeon2026routerkd} erases the differences between operators is open.
\emph{Draws and masks.} Every mask comes from slice A, so refits on slices B and C change only what an operator fits; rerouting and nearest remapping fit nothing and are identical on every slice. Results fitted on slice A only remain exposed to calibration noise for some operators and for every Granite result (\cref{app:mv-draws}). The calibration-domain results use one GSM8K draw; the mask grid covers 50\% on two models.
\emph{Re-implementations.} MoRA, ConMoE and COMPEL are our implementations of their papers (none had public code by 27 October 2026), each validated by a check that it computes what its equations specify (\cref{app:mora,app:others}). They are component comparisons under our masks and calibration, not reproductions of the complete published pipelines, and our MoRA re-implementation scores 1.2--4.4 points below MoRA's published numbers (\cref{app:mora-repro}). We did not run merging methods ourselves.
\emph{Generation and cost.} Our generative suite is GSM8K, HumanEval and MBPP (25\% only); under general-text calibration, code is largely lost at 25\% and GSM8K at 50\% under every operator, so operator differences in generation rest on few settings. MATH and long-form tasks are not covered. Operator comparisons are matched in stored parameters, not in compute.
\emph{Statistics.} Holm correction controls the family-wise error for test-item noise across 111 specified contrasts; other comparisons are exploratory. Calibration-draw noise is handled only by the same-sign rule, and only where several draws exist: a stability check on disjoint slices of one stream with the mask held fixed, not a statistical treatment of calibration uncertainty.

\FloatBarrier
\section*{Impact Statement}
Training-free pruning can lower the memory needed to store MoE models, but memory reduction is not automatically an inference-energy saving. We measured selected memory and latency configurations, not energy consumption across serving systems. A compressed model can retain most of its multiple-choice score while losing much of its mathematical and coding capability. For deployment, we recommend task-relevant generative evaluation, explicit reporting of calibration distributions and routing semantics, and comparison with quantization at matched storage.

\section*{AI Use Statement}
We use AI tools to assist with literature research, methodological discussions, and manuscript polishing. All results have been manually reviewed by the authors.

\bibliography{references}
\bibliographystyle{icml2026}

\newpage
\appendix
\onecolumn

\input{appendix}

\end{document}

%% file: authors.tex
\icmlsetsymbol{equal}{*}

\begin{icmlauthorlist}
\icmlauthor{Ezza Ali}{seecs,ncai}
\icmlauthor{Sehrish Bibi}{ncai}
\icmlauthor{Momina Moetesum}{seecs,ncai}
\icmlauthor{Faisal Shafait}{seecs,ncai}
\end{icmlauthorlist}

\icmlaffiliation{seecs}{School of Electrical Engineering and Computer Science (SEECS), National University of Sciences and Technology (NUST)}
\icmlaffiliation{ncai}{Deep Learning Laboratory,
National Center of Artificial Intelligence (NCAI), NUST}

\icmlcorrespondingauthor{Ezza Ali}{eali.phdai25seecs@seecs.edu.pk}

%% file: appendix.tex
\counterwithin{table}{section}
\renewcommand{\thetable}{\thesection.\arabic{table}}
\let\appolditemize\itemize
\renewcommand{\itemize}{\appolditemize\setlength{\itemsep}{1pt}\setlength{\parskip}{0pt}\setlength{\topsep}{2pt}}

Conventions are the same as in the main text:

\begin{itemize}
\item \textbf{Models.} ``DeepSeek'' in tables is DeepSeek-V2-Lite.
\item \textbf{Masks and data.} Masks are MoNE's (fusion score, per-layer ranking, Zyda-2 slice A) unless stated otherwise. A, B and C are the three disjoint calibration draws.
\item \textbf{Units.} Accuracy and its differences are in points.
\item \textbf{Intervals.} Brackets give paired, task-stratified item-bootstrap intervals (B = 10,000, seed 0). * marks an interval that excludes zero.
\item \textbf{``Nine-task'' average.} The unweighted mean of ARC-c, ARC-e, BoolQ, COPA, MMLU, OpenBookQA, PIQA, RTE and WinoGrande.
\item \textbf{``Big-5'' average.} The mean of ARC-e, ARC-c, BoolQ, PIQA and WinoGrande (9,923 items).
\end{itemize}

\section{Per-task and per-setting results}
\label{app:tables}

\subsection{Every operator at every ratio}
\label{app:A1}

\begin{table}[H]
\centering
\caption{Nine-task / big-5 / MMLU accuracy, MoNE masks, draw A. Rerouting at the trained gate scale is keep-scale on OLMoE and DeepSeek and renormalized on Moonlight; the other-scale cells give nine-task / big-5 only, and nine-task only on Moonlight (from \cref{tab:moonkeep}).}
\label{tab:A1}
\footnotesize
\begin{adjustbox}{max width=\linewidth}
\begin{tabular}{@{}lccccc@{}}
\toprule
Model & Pruned & Constant (MoNE) & Tilt r32 & Reroute, trained scale & Reroute, other scale \\
\midrule
OLMoE & 0\% (dense) & 66.76 / 68.69 / 53.62 & & & \\
 & 25\% & 61.19 / 63.62 / 41.64 & 63.02 / 65.40 / 42.45 & 62.65 / 64.10 / 42.94 & 58.15 / 58.51 \\
 & 50\% & 49.09 / 50.97 / 26.39 & 52.54 / 56.66 / 26.50 & 51.48 / 53.73 / 26.14 & 48.08 / 49.42 \\
 & 75\% & 41.04 / 42.69 / 22.97 & 43.84 / 46.67 / 23.02 & 42.94 / 43.91 / 22.87 & 39.85 / 39.07 \\
\midrule
DeepSeek & 0\% (dense) & 67.15 / 71.20 / 55.01 & & & \\
 & 25\% & 65.40 / 69.51 / 49.32 & 65.56 / 69.48 / 49.29 & 65.66 / 69.76 / 49.66 & 64.87 / 69.08 \\
 & 50\% & 59.84 / 64.06 / 40.98 & 60.32 / 64.51 / 41.63 & 61.19 / 65.39 / 42.81 & 56.52 / 59.12 \\
 & 75\% & 47.59 / 49.89 / 23.00 & 48.82 / 52.32 / 22.96 & 51.80 / 55.80 / 22.96 & 43.33 / 44.96 \\
\midrule
Moonlight & 0\% (dense) & 71.48 / 74.56 / 67.36 & & & \\
 & 25\% & 68.66 / 73.57 / 54.70 & 68.79 / 73.14 / 57.09 & 68.71 / 73.70 / 54.48 & 68.88 \\
 & 50\% & 60.24 / 66.19 / 23.80 & 61.61 / 67.64 / 28.67 & 60.06 / 64.54 / 27.30 & 58.38 \\
 & 75\% & 43.89 / 45.15 / 22.96 & 47.16 / 49.86 / 23.11 & 45.48 / 46.38 / 23.66 & 43.88 \\
\bottomrule
\end{tabular}
\end{adjustbox}
\end{table}

\textbf{With constants, MMLU sits at chance (23–27\%) in five settings, and only on Moonlight 50\% does an operator lift it clearly.} The five settings are OLMoE at 50\% and 75\%, DeepSeek at 75\%, and Moonlight at 50\% and 75\%. On Moonlight 50\% the tilt adds +4.9 and rerouting +3.5. Every other tilt or keep-reroute change to MMLU in those settings is within 0.8 points.

\subsection{Per-task change over the constant}
\label{app:pertask}\label{app:A2}

\begin{table}[H]
\centering
\caption{Per-task change (second run minus first), MoNE masks, Zyda-2 calibration. Reroute: rerouting at the trained gate scale (renormalized on Moonlight); renorm $-$ keep is given for the two routers trained without renormalization. The last column is the nine-task difference, which is the unweighted mean of the nine per-task changes (checked to 10$^{-14}$).}
\label{tab:A2}
\footnotesize
\begin{adjustbox}{max width=\linewidth}
\begin{tabular}{@{}llcccccccccc@{}}
\toprule
Setting & Comparison & ARC-c & ARC-e & BoolQ & COPA & MMLU & OBQA & PIQA & RTE & Wino & Nine \\
\midrule
OLMoE 50\% & tilt − const (A) & +4.18 & +13.85 & +2.91 & +4.00 & +0.11 & +1.40 & +6.86 & −2.89 & +0.63 & +3.45 \\
 & tilt − const (B) & +4.10 & +13.80 & +3.79 & +4.00 & +0.66 & +1.60 & +6.69 & −2.17 & +0.32 & +3.64 \\
 & tilt − const (C) & +3.92 & +13.38 & +3.70 & +4.00 & +0.05 & +1.00 & +7.07 & −1.44 & −0.71 & +3.44 \\
 & reroute − const (A) & +3.41 & +4.71 & −1.22 & +1.00 & −0.24 & +4.00 & +5.01 & +2.89 & +1.89 & +2.38 \\
 & renorm − keep & −2.56 & −3.11 & −4.46 & +3.00 & −2.04 & −5.60 & −3.21 & −4.33 & −8.21 & −3.39 \\
\midrule
OLMoE 25\% & tilt − const (A) & +2.05 & +4.46 & +1.41 & +1.00 & +0.81 & +1.80 & +0.76 & +3.97 & +0.24 & +1.83 \\
 & tilt − const (B) & +0.34 & +4.29 & +1.13 & +2.00 & +1.13 & +0.20 & +1.14 & +5.05 & +1.34 & +1.85 \\
 & reroute − const (A) & +0.17 & +1.14 & −0.03 & +2.00 & +1.30 & +1.00 & +0.16 & +6.50 & +0.95 & +1.46 \\
 & renorm − keep & −3.16 & −9.47 & −10.21 & −4.00 & −2.08 & −2.20 & −1.85 & −4.33 & −3.24 & −4.50 \\
\midrule
DeepSeek 50\% & tilt − const (A) & +0.17 & +1.22 & +0.34 & −1.00 & +0.65 & +1.00 & +0.27 & +1.44 & +0.24 & +0.48 \\
 & tilt − const (B) & +0.17 & +1.05 & −0.34 & +1.00 & +0.25 & +1.00 & +0.54 & +0.00 & +1.26 & +0.55 \\
 & tilt − const (C) & +0.17 & +0.29 & −0.12 & +0.00 & +1.18 & +0.80 & +0.33 & +1.08 & +0.79 & +0.50 \\
 & reroute − const (A) & +2.22 & +2.57 & +0.61 & +2.00 & +1.83 & +0.60 & +1.20 & +1.08 & +0.08 & +1.35 \\
 & renorm − keep & −1.54 & −4.80 & −18.53 & +4.00 & −7.35 & +0.60 & −0.44 & −7.94 & −6.08 & −4.67 \\
\midrule
DeepSeek 25\% & tilt − const (A) & +0.43 & +0.17 & −0.09 & −1.00 & −0.02 & +0.40 & +0.00 & +2.17 & −0.63 & +0.16 \\
 & tilt − const (B) & +0.60 & +0.00 & −0.55 & +2.00 & +0.40 & +0.60 & +0.00 & +3.61 & +0.00 & +0.74 \\
 & reroute − const (A) & +1.54 & +0.38 & −0.12 & +0.00 & +0.34 & +0.40 & +0.16 & +0.36 & −0.71 & +0.26 \\
 & renorm − keep & +0.09 & −0.51 & −5.08 & +1.00 & −1.40 & +1.00 & +0.00 & −4.33 & +2.13 & −0.79 \\
\midrule
Moonlight 50\% & tilt − const (A) & +1.54 & +2.10 & +1.62 & +3.00 & +4.87 & +0.40 & +1.14 & −3.25 & +0.87 & +1.37 \\
 & tilt − const (B) & +1.37 & +1.73 & +0.86 & +3.00 & +5.08 & +1.00 & +0.87 & −1.08 & +0.87 & +1.52 \\
 & tilt − const (C) & +1.45 & +1.39 & +1.25 & +4.00 & +6.96 & −0.60 & +0.87 & −2.53 & +1.34 & +1.57 \\
 & reroute − const (A) & +2.13 & −0.13 & −10.28 & +1.00 & +3.50 & +1.40 & +0.82 & +0.72 & −0.79 & −0.18 \\
\midrule
Moonlight 25\% & tilt − const (A) & −0.26 & −0.51 & −0.09 & +0.00 & +2.39 & +0.60 & −0.38 & +0.36 & −0.95 & +0.13 \\
 & tilt − const (B) & −0.17 & −0.59 & −0.06 & +0.00 & +2.10 & +0.80 & −0.27 & +0.36 & +0.16 & +0.26 \\
 & reroute − const (A) & +0.60 & −0.04 & +0.09 & +0.00 & −0.22 & +0.40 & +0.00 & −0.36 & +0.00 & +0.05 \\
\bottomrule
\end{tabular}
\end{adjustbox}
\end{table}

\textbf{The tilt's OLMoE gain is broad and repeats task by task across draws.} It is largest on ARC-e (+13.4 to +13.9) and PIQA (+6.7 to +7.1). Its only loss on every draw, RTE, is on draw A, where we measured ranking ability, a calibration shift (\cref{sec:auc}, \cref{tab:A7}).

\textbf{On Moonlight the tilt's largest single contribution is MMLU, which the constant model answers at chance} (\cref{tab:A1}). The tilt lifts it by 4.9–7.0 points on every draw. Part of that is letter bias (\cref{sec:auc}), and part is real: removing each letter's mean preference raises the tilt's MMLU to 37.7\% against 28.1\% for the constant.

\textbf{On OLMoE and \dsv{}, renormalization hurts almost every task,} most of all BoolQ (−18.5 on DeepSeek 50\%, −10.2 on OLMoE 25\%) and WinoGrande (−8.2 on OLMoE 50\%). COPA (100 items) and OpenBookQA (500) are the only tasks where it gains in more than one setting.

\subsection{75\% pruning}
\label{app:A3}

\begin{table}[H]
\centering
\caption{Paired differences at 75\% pruning (MoNE masks, draw A): nine-task [95\% CI] | big-5 | MMLU. Reroute: rerouting at the trained gate scale.}
\label{tab:A3}
\footnotesize
\begin{adjustbox}{max width=\linewidth}
\begin{tabular}{@{}lccc@{}}
\toprule
Model & Tilt − constant & Reroute − constant & Tilt − reroute \\
\midrule
OLMoE & +2.80 [+1.81, +3.81]* $|$ +3.99* $|$ +0.04 & +1.90 [+0.95, +2.86]* $|$ +1.22* $|$ −0.11 & +0.90 [−0.20, +2.01] $|$ +2.77* $|$ +0.15 \\
DeepSeek & +1.23 [+0.32, +2.15]* $|$ +2.43* $|$ −0.04 & +4.21 [+3.21, +5.24]* $|$ +5.91* $|$ −0.04 & −2.98 [−4.08, −1.91]* $|$ −3.48* $|$ +0.00 \\
Moonlight & +3.27 [+2.43, +4.15]* $|$ +4.71* $|$ +0.15 & +1.59 [+0.62, +2.59]* $|$ +1.23* $|$ +0.71* & +1.68 [+0.58, +2.77]* $|$ +3.47* $|$ −0.56* \\
\bottomrule
\end{tabular}
\end{adjustbox}
\end{table}

\textbf{At 75\% both operators beat the constant on every model, and the model-dependent ranking of \cref{sec:operators} persists.} Rerouting leads on DeepSeek, the tilt on Moonlight; on OLMoE the two are not detectably different (tilt $-$ keep +0.90 [$-$0.20, +2.01]).

\subsection{Calibration draws}
\label{app:A4}

\begin{table}[H]
\centering
\caption{Tilt r32 minus constant on each disjoint calibration draw (nine-task / big-5 / MMLU; the constant's own score in brackets).}
\label{tab:A4}
\footnotesize
\begin{adjustbox}{max width=\linewidth}
\begin{tabular}{@{}lccccc@{}}
\toprule
Setting & Draw A & Draw B & Draw C & Mean & Spread \\
\midrule
OLMoE 50\% & +3.45 / +5.68 / +0.11 (49.09) & +3.64 / +5.74 / +0.66 (49.18) & +3.44 / +5.47 / +0.05 (49.26) & +3.51 & 0.20 \\
OLMoE 25\% & +1.83 / +1.78 / +0.81 (61.19) & +1.85 / +1.65 / +1.13 (61.18) & — & +1.84 & 0.01 \\
DeepSeek 50\% & +0.48 / +0.45 / +0.65 (59.84) & +0.55 / +0.54 / +0.25 (59.77) & +0.50 / +0.29 / +1.18 (59.96) & +0.51 & 0.07 \\
DeepSeek 25\% & +0.16 / −0.03 / −0.02 (65.40) & +0.74 / +0.01 / +0.40 (65.01) & — & +0.45 & 0.58 \\
Moonlight 50\% & +1.37 / +1.45 / +4.87 (60.24) & +1.52 / +1.14 / +5.08 (60.21) & +1.57 / +1.26 / +6.96 (60.33) & +1.49 & 0.20 \\
Moonlight 25\% & +0.13 / −0.44 / +2.39 (68.66) & +0.26 / −0.19 / +2.10 (68.38) & — & +0.19 & 0.13 \\
\bottomrule
\end{tabular}
\end{adjustbox}
\end{table}

\textbf{Within a model the gain grows with the damage the constant leaves, up to 50\% pruning everywhere and to 75\% on two models; across models it does not follow one curve.}

\begin{itemize}
\item The 50\% gain exceeds the 25\% gain on every model. At 75\% (draw A, \cref{tab:main}) it rises further on \dsv{} (+1.23, damage 19.56) and Moonlight (+3.27, damage 27.59) but falls on OLMoE (+2.80, damage 25.72), where the tilt's share of the accuracy damage also falls (\cref{tab:ratio}).
\item Over the nine draw-A settings, the rank correlation between the constant's damage and the tilt's gain is 0.73 (descriptive; nine settings from three models).
\item Across models, OLMoE at 25\% (damage 5.57) gains +1.84, while DeepSeek at 50\% (damage 7.31) gains +0.51. The tilt's value has a model-specific component.
\end{itemize}
\subsection{Language modelling at every ratio}
\label{app:A5}

\begin{table}[H]
\centering
\caption{WikiText word perplexity (max length 2,048). Bracketed values are the share of the constant's log-perplexity damage recovered, ln(c/op)/ln(c/dense), computed from unrounded perplexities; where the perplexities differ by hundredths (Moonlight 25\%), the shares cannot be recomputed from the rounded values shown.}
\label{tab:A5}\label{tab:ppl}
\footnotesize
\begin{adjustbox}{max width=\linewidth}
\begin{tabular}{@{}lccccc@{}}
\toprule
Model & Pruned & Constant & Tilt r32 & Reroute, trained & Reroute, other \\
\midrule
OLMoE (dense 8.31) & 25\% & 14.64 & 14.36 (3.4\%) & 14.34 (3.6\%) & — \\
 & 50\% & 51.65 & 36.95 (18.3\%) & 42.20 (11.1\%) & 98.03 (−35.1\%) \\
 & 75\% & 1434.38 & 171.93 (41.2\%) & 1279.67 (2.2\%) & — \\
\midrule
DeepSeek (dense 9.44) & 25\% & 10.97 & 11.05 (−4.9\%) & 10.70 (16.7\%) & — \\
 & 50\% & 22.74 & 19.25 (19.0\%) & 17.88 (27.4\%) & 19.70 (16.3\%) \\
 & 75\% & 105.96 & 56.16 (26.3\%) & 61.64 (22.4\%) & — \\
\midrule
Moonlight (dense 9.31) & 25\% & 9.50 & 9.46 (17.7\%) & 9.51 (−10.4\%) & — \\
 & 50\% & 14.92 & 13.66 (18.7\%) & 15.26 (−4.7\%) & — \\
 & 75\% & 221.90 & 67.09 (37.7\%) & 102.79 (24.3\%) & — \\
\bottomrule
\end{tabular}
\end{adjustbox}
\end{table}

\textbf{At 75\% the tilt cuts OLMoE's perplexity by a factor of eight, while rerouting barely moves it.}

\begin{itemize}
\item Constant: 1434. Rerouting: 1280. Tilt: 172.
\item Yet on accuracy the two operators are within one point of each other (+0.90, n.s.; \cref{tab:A3}).
\end{itemize}
\textbf{Perplexity and accuracy part ways most when damage is extreme.}

\textbf{As pruning deepens, the tilt restores more language modelling but not more accuracy.} Its perplexity share rises with the pruning ratio on every model, while its accuracy share does not (\cref{tab:ratio}): it falls on OLMoE (32.9 → 19.5 → 10.9\%) and \dsv{} (9.0 → 6.6 → 6.3\%) and peaks at 50\% on Moonlight (4.6 → 12.2 → 11.8\%). On OLMoE the two move in opposite directions, from 3.4\% of the log-perplexity damage and 32.9\% of the accuracy damage at 25\% to 41.2\% and 10.9\% at 75\%.

\begin{table}[H]
\caption{The tilt's share of the constant's damage recovered by pruning ratio (\%): WikiText log-perplexity / nine-task accuracy, MoNE masks, draw A.}
\label{tab:ratio}
\centering\footnotesize
\begin{tabular}{@{}lccc@{}}
\toprule
Model & 25\% & 50\% & 75\% \\
\midrule
OLMoE & 3.4 / 32.9 & 18.3 / 19.5 & 41.2 / 10.9 \\
DeepSeek & −4.9 / 9.0 & 19.0 / 6.6 & 26.3 / 6.3 \\
Moonlight & 17.7 / 4.6 & 18.7 / 12.2 & 37.7 / 11.8 \\
\bottomrule
\end{tabular}
\end{table}

\subsection{Generation}
\label{app:A6}

\begin{table}[H]
\centering
\caption{GSM8K (5-shot; strict / flexible extraction) and HumanEval pass@1. Zyda-2 calibration unless marked GSM8K. Reroute: rerouting at the trained gate scale.}
\label{tab:A6}
\footnotesize
\begin{adjustbox}{max width=\linewidth}
\begin{tabular}{@{}lcclc@{}}
\toprule
Model & Setting & Constant & Tilt r32 & Reroute \\
\midrule
OLMoE (dense 45.49 / 52.39; HumanEval 11.59) & 25\%, Zyda & 2.43 / 3.03; 0 & 1.82 / 2.20; 0 & 2.73 / 2.96; 0 \\
 & 50\%, Zyda & 0.00 / 1.59; 0 & 0.00 / 1.59; 0 & 0.68 / 2.05; 0 \\
 & 25\%, GSM8K & 46.32 / 52.16 & 47.54 / 52.84 & 48.45 / 52.24 \\
 & 50\%, GSM8K & 21.61 / 22.74 & 41.02 / 44.58 & 44.28 / 44.66 \\
\midrule
DeepSeek (dense 37.53 / 37.91; HumanEval 29.27) & 25\%, Zyda & 24.56 / 24.79; 3.66 & 23.12 / 23.43; 4.27 & 23.81 / 24.03; 5.49 \\
 & 50\%, Zyda & 3.79 / 4.32; 0 & 4.62 / 4.85; 0 & 6.37 / 6.67; 0 \\
 & 25\%, GSM8K & 37.76 / 37.83 & 37.30 / 37.83 & 37.76 / 38.29 \\
 & 50\%, GSM8K & 34.80 / 35.18 & 34.65 / 35.10 & 35.41 / 35.71 \\
\midrule
Moonlight (dense 73.84 / 73.92; HumanEval 48.78) & 25\%, Zyda & 32.60 / 32.75; 0 & 32.75 / 32.75; 0 & 42.76 / 42.84; 0 \\
 & 50\%, Zyda & 0.38 / 1.74; 0 & 1.36 / 1.82; 0 & 1.52 / 2.20; 0 \\
 & 25\%, GSM8K & 73.84 / 73.77 & 73.39 / 73.24 & 74.37 / 74.45 \\
 & 50\%, GSM8K & 70.36 / 70.36 & 73.31 / 73.24 & 71.19 / 71.19 \\
\bottomrule
\end{tabular}
\end{adjustbox}
\end{table}

\begin{table}[H]
\centering
\caption{Paired GSM8K (strict) differences over the constant not shown in the main text.}
\label{tab:A6b}
\footnotesize
\begin{adjustbox}{max width=\linewidth}
\begin{tabular}{@{}llc@{}}
\toprule
Setting & Tilt − constant & Reroute − constant \\
\midrule
OLMoE 25\%, GSM8K & +1.21 [−1.06, +3.41] & +2.12 [−0.23, +4.47] \\
DeepSeek 50\%, GSM8K & −0.15 [−2.35, +2.05] & +0.61 [−1.59, +2.73] \\
DeepSeek 25\%, GSM8K & −0.45 [−2.12, +1.21] & +0.00 [−1.59, +1.67] \\
Moonlight 50\%, GSM8K & +2.96 [+1.06, +4.85]* & +0.83 [−1.14, +2.81] \\
Moonlight 25\%, GSM8K & −0.45 [−1.74, +0.76] & +0.53 [−0.99, +1.97] \\
\bottomrule
\end{tabular}
\end{adjustbox}
\end{table}

\textbf{Once the mask is calibrated in-domain, operators matter only where damage remains.}

\begin{itemize}
\item OLMoE at 50\% has the largest remaining damage (21.61 against dense 45.49): tilt +19.4*, keep +22.7* (\cref{sec:domain}).
\item Moonlight at 50\% (70.36 against dense 73.84): tilt +2.96* (to 73.31), rerouting +0.83 (n.s.).
\item Elsewhere every in-domain operator difference is inside its interval.
\item Every 50\% model scores zero on HumanEval under every operator.
\end{itemize}

\begin{table}[H]
\centering
\caption{Share of the dense nine-task score kept, raw / chance-adjusted, $(\text{op}-\text{chance})/(\text{dense}-\text{chance})$, with chance 40.55 (uniform guessing; the majority answer on BoolQ and RTE). Generation tasks have chance near zero, so their raw shares need no adjustment. MoNE masks, draw A.}
\label{tab:chance}
\footnotesize
{\begin{adjustbox}{max width=\linewidth}
\begin{tabular}{@{}llccc@{}}
\toprule
Model & Pruned & Constant & Tilt r32 & Reroute \\
\midrule
OLMoE & 25\% & 91.7 / 78.7 & 94.4 / 85.7 & 93.8 / 84.3 \\
 & 50\% & 73.5 / 32.6 & 78.7 / 45.8 & 77.1 / 41.7 \\
 & 75\% & 61.5 / \phantom{0}1.9 & 65.7 / 12.6 & 64.3 / \phantom{0}9.1 \\
DeepSeek & 25\% & 97.4 / 93.4 & 97.6 / 94.0 & 97.8 / 94.4 \\
 & 50\% & 89.1 / 72.5 & 89.8 / 74.3 & 91.1 / 77.6 \\
 & 75\% & 70.9 / 26.5 & 72.7 / 31.1 & 77.1 / 42.3 \\
Moonlight & 25\% & 96.1 / 90.9 & 96.2 / 91.3 & 96.1 / 91.1 \\
 & 50\% & 84.3 / 63.7 & 86.2 / 68.1 & 84.0 / 63.1 \\
 & 75\% & 61.4 / 10.8 & 66.0 / 21.4 & 63.6 / 16.0 \\
\bottomrule
\end{tabular}
\end{adjustbox}}
\end{table}

\textbf{Chance-adjusted, the multiple-choice share still dwarfs the generative one.} At 25\% the constant keeps 79\%, 93\% and 91\% of the margin above chance (OLMoE, \dsv{}, Moonlight) against 5\%, 65\% and 44\% of GSM8K and 0\%, 17\% and 0\% of MBPP. At 75\% OLMoE's constant model is 1.9\% of the way from chance to dense: its nine-task accuracy carries almost no information about retained ability.

\subsection{MBPP at 25\% pruning}
\label{app:mbpp}

\begin{table}[H]
\centering
\caption{MBPP pass@1 (3-shot, lm-eval, 500 problems), 25\% pruning, MoNE masks, Zyda-2 calibration, draw A. Strict: lm-eval's score, which executes the whole generation; lenient: the generation cut at its first answer boundary ([DONE], [BEGIN], the few-shot preamble or a top-level assert / print), then executed with the tests (3 s timeout, as lm-eval). Last column: generations that ran past their answer. Keep: rerouting at the trained gate scale. Contrasts use strict scores; \holmt: survives Holm within the MBPP family (tilt and keep against the constant).}
\label{tab:mbpp}
\footnotesize
\begin{adjustbox}{max width=\linewidth}
\begin{tabular}{@{}llcccc@{}}
\toprule
Model & Operator & Strict & Lenient & Share of dense (strict) & Ran past answer \\
\midrule
OLMoE & dense & 21.00 & 21.00 & --- & 0 \\
 & constant / tilt / keep & 0.00 / 0.00 / 0.00 & 0.00 / 0.00 / 0.00 & 0\% & 4 / 0 / 1 \\
DeepSeek & dense & 43.60 & 43.60 & --- & 0 \\
 & constant / tilt / keep & 7.60 / 6.40 / 9.20 & 8.20 / 6.40 / 11.20 & 17.4 / 14.7 / 21.1\% & 52 / 9 / 58 \\
Moonlight & dense & 52.00 & 52.00 & --- & 2 \\
 & constant / tilt / keep & 0.00 / 0.00 / 1.20 & 0.00 / 0.00 / 1.20 & 0 / 0 / 2.3\% & 0 / 1 / 0 \\
\midrule
\multicolumn{6}{@{}l}{\emph{Contrasts (strict; lenient in brackets where it differs)}} \\
DeepSeek & tilt $-$ constant & \multicolumn{4}{l}{$-$1.20 [$-$3.20, +0.80] (lenient $-$1.80 [$-$3.80, +0.40])} \\
 & keep $-$ constant & \multicolumn{4}{l}{+1.60 [$-$0.20, +3.40] (lenient +3.00 [+1.40, +4.80]*)} \\
 & tilt $-$ keep & \multicolumn{4}{l}{$-$2.80 [$-$5.20, $-$0.40]* (lenient $-$4.80 [$-$7.20, $-$2.40]*)} \\
Moonlight & keep $-$ constant & \multicolumn{4}{l}{+1.20 [+0.40, +2.20]*\holmt} \\
 & tilt $-$ keep & \multicolumn{4}{l}{$-$1.20 [$-$2.20, $-$0.40]*} \\
OLMoE & all & \multicolumn{4}{l}{0 (no problem solved by any pruned model)} \\
\bottomrule
\end{tabular}
\end{adjustbox}
\end{table}

\textbf{The collapse is broken code, not an extraction artifact.}

\begin{itemize}
\item \textbf{Answer format.} lm-eval stops at ``[DONE]'' and executes the whole generation, so a model that writes a function and then invents the next few-shot task fails on the invented prose. Cutting every generation at its first answer boundary never turned a pass into a fail, left every dense score and every OLMoE and Moonlight score unchanged, and raised only \dsv{}'s constant and keep (52 and 58 run-on generations) by 0.6 and 2.0 points.
\item \textbf{Function names.} The pruned models name the required function in 472 (Moonlight keep) and 498 (\dsv{} keep) of 500 generations; OLMoE keep does so in 298 and Moonlight's constant in 376.
\item \textbf{Failure types} (executing the cut generations): the dense Moonlight model fails mainly with wrong answers (206 assertion errors, 14 undefined names). Pruned, Python rejects 381 of OLMoE keep's 500 programs, 193 of Moonlight keep's and all 500 of Moonlight constant's as syntax errors; \dsv{} keep fails mainly on wrong answers (243) and type errors (98).
\end{itemize}

\textbf{The operators damage code differently.} Moonlight on one problem (sort a matrix by row sums; first lines of each generation):

\begin{center}
\footnotesize
\begin{adjustbox}{max width=\linewidth}
\begin{tabular}{@{}ll@{}}
\toprule
Model & Generation \\
\midrule
dense (passes) & \texttt{def sort\_matrix(matrix): matrix.sort(key = lambda x: sum(x)); return matrix} \\
25\%, keep (valid Python, fails) & \texttt{import numpy as np; def sort\_matrix(nums,n): sort\_matrix = np.sorted(n); \ldots} \\
25\%, constant (not Python) & \texttt{def sort\_matrix(test matrix)  result = matrix  return result} \\
\bottomrule
\end{tabular}
\end{adjustbox}
\end{center}

Lines are joined with ``;'' or spaces for print. In the three generations we inspected, Moonlight's constant writes the same template (no colon after the signature, names with spaces), and all 500 of its programs are syntax errors; keep-scale rerouting keeps the syntax but invents APIs and signatures. Neither passes the tests.

\subsection{Multiple-choice cost of task calibration}
\label{app:taskcal}

\begin{table}[H]
\caption{Change in multiple-choice accuracy from GSM8K calibration (GSM8K minus general calibration), constant novice. * = 95\% item-bootstrap interval excludes zero; WinoGrande is a point estimate.}
\label{tab:taskcal}
\centering\footnotesize
\begin{tabular}{@{}lccc@{}}
\toprule
Setting & Nine-task & MMLU & WinoGrande \\
\midrule
OLMoE 25\% & +0.06 & +3.81* & −3.39 \\
OLMoE 50\% & −0.16 & +9.61* & −6.95 \\
DeepSeek 25\% & −1.90* & −5.55* & −2.53 \\
DeepSeek 50\% & −3.89* & −6.53* & −6.08 \\
Moonlight 25\% & +1.06* & +5.70* & −3.00 \\
Moonlight 50\% & −3.86* & +17.38* & −8.21 \\
\bottomrule
\end{tabular}
\end{table}

\textbf{The price of task calibration on multiple choice depends on the model.} With the constant, \dsv{} loses multiple-choice accuracy and MMLU at both ratios (\cref{tab:taskcal}). OLMoE and Moonlight lose commonsense, WinoGrande in every setting, but gain MMLU strongly at 50\%. One configuration serves both kinds of task: OLMoE with a GSM8K-calibrated 50\% mask and keep-scale rerouting scores 54.00 on multiple choice and 44.28 on GSM8K (dense 66.76 and 45.49), with no fitted parameters.

\subsection{Mask drift at a matched token budget}
\label{app:maskdrift}

\begin{table}[H]
\centering
\caption{Token-matched mask drift: share of pruned experts that differ between two MoNE masks (\%). $N$ = each model's GSM8K-train cache in 4,096-token blocks (OLMoE 286, DeepSeek 346, Moonlight 295); A($n$), B($n$) = the first $n$ blocks of Zyda-2 slices A and B. Ratios: domain / resampling drift. Slice-B masks at 1,000 blocks exist at 50\% only.}
\label{tab:budgetfull}
\footnotesize
\begin{adjustbox}{max width=\linewidth}
\begin{tabular}{@{}llcccccccc@{}}
\toprule
Model & Pruned & GSM8K vs A(1000) & B(1000) vs A(1000) & B($N$) vs A(1000) & GSM8K vs A($N$) & B($N$) vs A($N$) & A($N$) vs A(1000) & Ratio, full budget & Ratio, both at $N$ \\
\midrule
OLMoE & 50\% & 34.8 & 6.4 & 6.2 & 35.7 & 7.2 & 4.9 & 5.4 & 4.9 \\
 & 25\% & 55.9 & --- & 10.9 & 55.1 & 11.3 & 7.8 & --- & 4.9 \\
DeepSeek & 50\% & 29.7 & 4.2 & 4.1 & 29.7 & 5.0 & 2.6 & 7.1 & 5.9 \\
 & 25\% & 50.0 & --- & 5.3 & 49.3 & 5.3 & 3.1 & --- & 9.3 \\
Moonlight & 50\% & 33.9 & 4.2 & 4.7 & 33.8 & 5.8 & 3.6 & 8.1 & 5.9 \\
 & 25\% & 39.2 & --- & 7.7 & 39.7 & 10.1 & 4.3 & --- & 3.9 \\
\bottomrule
\end{tabular}
\end{adjustbox}
\end{table}

\subsection{Calibration versus ability, every arm}
\label{app:A7}

\begin{table}[H]
\centering
\caption{Two-choice tasks with fixed answer strings, draw A, each operator against its constant; keep is rerouting at the trained gate scale (renormalized on Moonlight), renorm the other scale on OLMoE and DeepSeek. acc = accuracy as evaluated (threshold 0 on the log-likelihood margin). AUC = threshold-free ranking ability. prior = re-thresholded so predicted label shares match the gold ones. best = oracle threshold. Gaps are the operator's change as evaluated and after prior matching.}
\label{tab:A7}
\footnotesize
\begin{adjustbox}{max width=\linewidth}
\begin{tabular}{@{}llclccccc@{}}
\toprule
Setting & Task & Constant: acc / AUC / prior / best & Operator & acc & AUC & Gap raw & Gap prior & $\Delta$AUC \\
\midrule
OLMoE 50\% & BoolQ & 65.5 / 0.6875 / 65.6 / 67.9 & tilt & 68.4 & 0.7767 & +2.91 & +5.57 & +0.0892 \\
 & & & keep & 64.3 & 0.6413 & −1.22 & −3.88 & −0.0462 \\
 & & & renorm & 59.8 & 0.5507 & −5.69 & −9.97 & −0.1368 \\
\midrule
OLMoE 50\% & RTE & 55.6 / 0.6389 / 62.5 / 65.0 & tilt & 52.7 & 0.6308 & −2.89 & −1.44 & −0.0082 \\
 & & & keep & 58.5 & 0.6322 & +2.89 & −2.53 & −0.0067 \\
 & & & renorm & 54.2 & 0.4976 & −1.44 & −13.72 & −0.1413 \\
\midrule
OLMoE 25\% & BoolQ & 66.5 / 0.7188 / 67.6 / 68.3 & tilt & 68.0 & 0.7426 & +1.41 & +2.23 & +0.0238 \\
 & & & keep & 66.5 & 0.6958 & −0.03 & −1.74 & −0.0230 \\
 & & & renorm & 56.3 & 0.6576 & −10.24 & −3.30 & −0.0612 \\
\midrule
OLMoE 25\% & RTE & 63.5 / 0.7137 / 66.1 / 67.1 & tilt & 67.5 & 0.7666 & +3.97 & +4.33 & +0.0529 \\
 & & & keep & 70.0 & 0.7413 & +6.50 & +2.53 & +0.0276 \\
 & & & renorm & 65.7 & 0.7130 & +2.17 & −1.44 & −0.0007 \\
\midrule
DeepSeek 50\% & BoolQ & 73.1 / 0.8002 / 72.7 / 73.9 & tilt & 73.5 & 0.7964 & +0.34 & −0.06 & −0.0038 \\
 & & & keep & 73.8 & 0.8190 & +0.61 & +2.02 & +0.0188 \\
 & & & renorm & 55.2 & 0.6792 & −17.92 & −8.53 & −0.1211 \\
\midrule
DeepSeek 50\% & RTE & 56.7 / 0.6168 / 58.1 / 59.9 & tilt & 58.1 & 0.6463 & +1.44 & +1.81 & +0.0296 \\
 & & & keep & 57.8 & 0.5924 & +1.08 & −1.44 & −0.0244 \\
 & & & renorm & 49.8 & 0.5221 & −6.86 & −8.30 & −0.0946 \\
\midrule
DeepSeek 25\% & BoolQ & 78.6 / 0.8594 / 77.6 / 79.0 & tilt & 78.5 & 0.8559 & −0.09 & +0.00 & −0.0035 \\
 & & & keep & 78.5 & 0.8619 & −0.12 & +0.89 & +0.0024 \\
 & & & renorm & 73.4 & 0.8252 & −5.20 & −3.46 & −0.0342 \\
\midrule
DeepSeek 25\% & RTE & 59.6 / 0.6551 / 59.6 / 61.7 & tilt & 61.7 & 0.6587 & +2.17 & +2.17 & +0.0036 \\
 & & & keep & 59.9 & 0.6471 & +0.36 & +0.36 & −0.0080 \\
 & & & renorm & 55.6 & 0.6152 & −3.97 & −1.44 & −0.0399 \\
\midrule
Moonlight 50\% & BoolQ & 76.3 / 0.8406 / 76.1 / 76.6 & tilt & 78.0 & 0.8547 & +1.62 & +1.59 & +0.0141 \\
 & & & keep & 66.1 & 0.7468 & −10.28 & −6.42 & −0.0938 \\
\midrule
Moonlight 50\% & RTE & 58.8 / 0.6799 / 61.0 / 66.8 & tilt & 55.6 & 0.6813 & −3.25 & +0.72 & +0.0015 \\
 & & & keep & 59.6 & 0.6844 & +0.72 & +4.33 & +0.0045 \\
\midrule
Moonlight 25\% & BoolQ & 79.4 / 0.8690 / 79.1 / 79.7 & tilt & 79.4 & 0.8691 & −0.09 & −0.31 & +0.0001 \\
 & & & keep & 79.5 & 0.8668 & +0.09 & −0.28 & −0.0022 \\
\midrule
Moonlight 25\% & RTE & 59.6 / 0.7604 / 67.5 / 71.1 & tilt & 59.9 & 0.7602 & +0.36 & −1.08 & −0.0002 \\
 & & & keep & 59.2 & 0.7653 & −0.36 & +2.17 & +0.0049 \\
\bottomrule
\end{tabular}
\end{adjustbox}
\end{table}

\textbf{Renormalization lowers ranking ability against keep-scale rerouting in all eight OLMoE and \dsv{} cells}, by 0.028--0.140 in AUC (point estimates; e.g.\ OLMoE 50\% RTE 0.498 against 0.632, \dsv{} 50\% BoolQ 0.679 against 0.819).

\textbf{On fixed-answer tasks, accuracy understates ability even at light pruning.}

\begin{itemize}
\item Moonlight at 25\% evaluates RTE at 59.6\% with AUC 0.760 and best-threshold accuracy 71.1\%.
\item The method was validated on synthetic margins first: a pure shift leaves AUC unchanged and prior matching closes the gap exactly; a shift plus reordering lowers AUC and leaves a gap.
\item On Moonlight 50\% BoolQ, rerouting flips 858 items relative to the constant: 466 more ``no'' predictions on net, but 196 flips go the other way. So the loss is not a pure threshold shift.
\end{itemize}

\begin{table}[H]
\centering
\caption{Fixed-answer two-choice tasks with paired item-bootstrap 95\% intervals for the largest effects ($B=10{,}000$; items resampled jointly for both runs). AUC reading: ``not detectable'' = the AUC-change interval contains 0; ``chance'' = the operator's AUC interval contains 0.5.}
\label{tab:aucci}
\footnotesize
\begin{adjustbox}{max width=\linewidth}
\begin{tabular}{@{}llcccccl@{}}
\toprule
Comparison & Task & $n$ & Acc.\ change & AUC (ref.) & AUC (op.) & AUC change & After re-thresholding; reading \\
\midrule
OLMoE 50\%, tilt vs constant & RTE & 277 & $-$2.89 [$-$5.42, $-$0.72] & 0.639 [0.570, 0.704] & 0.631 [0.562, 0.695] & $-$0.008 [$-$0.064, +0.047] & $-$1.44 [$-$8.30, +3.97]; not detectable \\
Moonlight 50\%, tilt vs constant & RTE & 277 & $-$3.25 [$-$6.50, +0.00] & 0.680 [0.616, 0.741] & 0.681 [0.617, 0.743] & +0.001 [$-$0.028, +0.030] & +0.72 [$-$3.97, +4.69]; not detectable \\
OLMoE 50\%, tilt vs constant & BoolQ & 3,270 & +2.91 [+2.08, +3.76] & 0.688 [0.669, 0.706] & 0.777 [0.760, 0.792] & +0.089 [+0.074, +0.104] & +5.57 [+3.94, +7.16]; higher \\
Moonlight 50\%, keep vs constant & BoolQ & 3,270 & $-$10.28 [$-$11.96, $-$8.56] & 0.841 [0.827, 0.854] & 0.747 [0.730, 0.763] & $-$0.094 [$-$0.108, $-$0.080] & $-$6.42 [$-$8.47, $-$5.14]; lower \\
OLMoE 50\%, renorm vs constant & RTE & 277 & $-$1.44 [$-$6.50, +3.61] & 0.639 [0.570, 0.704] & 0.498 [0.428, 0.566] & $-$0.141 [$-$0.232, $-$0.051] & $-$13.72 [$-$22.02, $-$5.05]; chance \\
OLMoE 50\%, renorm vs constant & BoolQ & 3,270 & $-$5.69 [$-$7.37, $-$3.98] & 0.688 [0.669, 0.706] & 0.551 [0.531, 0.571] & $-$0.137 [$-$0.163, $-$0.111] & $-$9.97 [$-$11.99, $-$7.52]; lower \\
DeepSeek 50\%, renorm vs constant & BoolQ & 3,270 & $-$17.92 [$-$19.85, $-$15.96] & 0.800 [0.785, 0.815] & 0.679 [0.660, 0.698] & $-$0.121 [$-$0.136, $-$0.107] & $-$8.53 [$-$10.03, $-$6.24]; lower \\
OLMoE 25\%, renorm vs constant & BoolQ & 3,270 & $-$10.24 [$-$12.72, $-$7.71] & 0.719 [0.701, 0.737] & 0.658 [0.638, 0.677] & $-$0.061 [$-$0.081, $-$0.041] & $-$3.30 [$-$5.47, $-$1.56]; lower \\
\bottomrule
\end{tabular}
\end{adjustbox}
\end{table}

\section{The tilt: fitting details}
\label{app:tilt}

\subsection{What is fitted}
\label{app:B1}

For a pruned expert i with routed calibration tokens $\Omega_i$:

\begin{itemize}
\item \textbf{Inputs.} x is the expert's input, taken from a forward pass of the \emph{unpruned} model, so every expert sees dense-model inputs. Its output is $E_i(x)$.
\item \textbf{Statistics.} We accumulate, in float64 and streaming, the count, means and centered second moments $C_{xx}$ and $C_{xy}$ over $\Omega_i$. Tokens are unweighted; gate weights do not enter.
\item \textbf{Ridge.} The ridge is relative: $C_{xx}+\lambda\,\mathrm{diag}(C_{xx})$, with $\lambda=10^{-2}$.
\item \textbf{Reduced rank.} With L the Cholesky factor of the ridged Gram matrix and A the full-rank ridge solution, we take the SVD $Q\Sigma W^\top$ of $L^\top A$ and keep the top $r$ directions: $V_i=\Sigma_r Q_r^\top L^{-1}$ ($r\times d$) and $U_i=W_r$ ($d\times r$), the maps of the main text. This is the rank-$r$ ridge regression that is optimal in the ridged metric.
\end{itemize}
The deployed operator is

\[N_i(x)=\mu_i+U_i\,\mathrm{clip}\big(V_i x-\bar{\phi}_i,\ \pm c_i\big),\]

where:

\begin{itemize}
\item $\bar{\phi}_i=V_i\bar{x}_i$ is the mean of the projected features over $\Omega_i$, and $c_i$ is four times their standard deviation;
\item the input mean $\bar{x}_i$ is folded into $\bar{\phi}_i$, so the tilt stores nothing of dimension $d$ beyond the constant's mean $\mu_i$;
\item the clip touches about 0.01\% of feature values;
\item the deployed module reproduces its fitted map to machine precision, and its squared error equals the fitted residual.
\end{itemize}
An expert with at most 4$\cdot$r routed calibration tokens falls back to the constant. \cref{tab:B1} shows this never happens with general calibration. With GSM8K calibration it happens for 5 experts on Moonlight.

\begin{table}[H]
\centering
\caption{Routed calibration tokens per pruned expert (rank-32 fits). ``Fallback'' = experts left with the constant.}
\label{tab:B1}
\footnotesize
\begin{adjustbox}{max width=\linewidth}
\begin{tabular}{@{}lccccc@{}}
\toprule
Fit set & Pruned & Min tokens & 5th pct & Median & Fallback \\
\midrule
OLMoE 50\%, Zyda A & 512 & 82,999 & 282,715 & 433,398 & 0 \\
OLMoE 50\%, GSM8K & 512 & 859 & 12,001 & 73,577 & 0 \\
DeepSeek 50\%, Zyda A & 832 & 203,810 & 276,833 & 326,865 & 0 \\
DeepSeek 50\%, GSM8K & 832 & 7,133 & 27,489 & 81,906 & 0 \\
Moonlight 50\%, Zyda A & 832 & 3,303 & 14,180 & 225,235 & 0 \\
Moonlight 50\%, GSM8K & 832 & 39 & 646 & 22,840 & 5 (layers 5, 6, 8, 9, 13) \\
Moonlight 25\%, GSM8K & 416 & 39 & 336 & 7,059 & 5 (the same five experts) \\
\bottomrule
\end{tabular}
\end{adjustbox}
\end{table}

\subsection{Parameter cost}
\label{app:B2}

A rank-r tilt adds the two maps ($2rd$) and the feature mean and clip scale ($2r$) to the pruned expert. Its output mean is the constant's own (d values, already in the pruned model), and the input mean is folded into the feature mean. So it adds $2rd+2r$ parameters per pruned expert, exactly what our implementation allocates.

\begin{table}[H]
\centering
\caption{Parameters the tilt adds to MoNE's constant-novice model (share of the pruned model).}
\label{tab:B2}
\footnotesize
\begin{adjustbox}{max width=\linewidth}
\begin{tabular}{@{}lcccc@{}}
\toprule
Model (dense) & Pruned & Pruned model & Tilt r8 & Tilt r32 \\
\midrule
OLMoE (6.919 B) & 25\% (256 experts) & 5.309 B & +8.4 M (0.16\%) & +33.6 M (0.63\%) \\
 & 50\% (512) & 3.699 B & +16.8 M (0.45\%) & +67.1 M (1.82\%) \\
 & 75\% (768) & 2.089 B & +25.2 M (1.21\%) & +100.7 M (4.82\%) \\
\midrule
DeepSeek (15.706 B) & 25\% (416) & 12.109 B & +13.6 M (0.11\%) & +54.6 M (0.45\%) \\
 & 50\% (832) & 8.511 B & +27.3 M (0.32\%) & +109.1 M (1.28\%) \\
 & 75\% (1,248) & 4.913 B & +40.9 M (0.83\%) & +163.7 M (3.33\%) \\
\midrule
Moonlight (15.960 B) & 25\% (416) & 12.362 B & +13.6 M (0.11\%) & +54.6 M (0.44\%) \\
 & 50\% (832) & 8.764 B & +27.3 M (0.31\%) & +109.1 M (1.24\%) \\
 & 75\% (1,248) & 5.167 B & +40.9 M (0.79\%) & +163.7 M (3.17\%) \\
\bottomrule
\end{tabular}
\end{adjustbox}
\end{table}

\textbf{The tilt adds 0.44–1.82\% of the pruned model's parameters at 25–50\% pruning and 3.2–4.8\% at 75\%.} The equal-budget comparison of Appendix C counts everything stored for a pruned expert: the tilt's maps plus its mean ($2rd+d+2r=133{,}184$ values) against a copy's factors plus its bias (\cref{tab:C1}).

\subsection{Why a per-expert map of the full input}
\label{app:B3}

Before building any operator, one calibration pass measured how much of the constant's error each family could remove at best. Error weights were $n_i\,\mathbb{E}[G_i^2]\,\mathrm{tr}\,\Sigma_i$; R$^{2}$ is held out, fitted on even calibration steps and scored on odd ones.

\begin{table}[H]
\centering
\caption{Held-out share of the constant's error removed by a linear map, OLMoE 50\% mask, 1,000 sequences, by feature set and rank.}
\label{tab:B3}
\footnotesize
\begin{adjustbox}{max width=\linewidth}
\begin{tabular}{@{}llccccc@{}}
\toprule
Layer & Features & r = 1 & r = 8 & r = 32 & Full rank & Full rank, in-sample \\
\midrule
0 & router logits (64) & 0.019 & 0.046 & 0.072 & 0.084 & — \\
0 & full input (2,048) & 0.028 & 0.080 & 0.147 & 0.412 & 0.471 \\
8 & router logits & 0.009 & 0.026 & 0.040 & 0.045 & — \\
8 & full input & 0.013 & 0.043 & 0.085 & 0.222 & 0.272 \\
15 & router logits & 0.093 & 0.161 & 0.181 & 0.185 & — \\
15 & full input & 0.129 & 0.243 & 0.304 & 0.419 & 0.486 \\
\bottomrule
\end{tabular}
\end{adjustbox}
\end{table}

\textbf{Expert outputs are partly linear in the input; the structure is high-rank and lives in the full input.}

\begin{itemize}
\item A full-rank map removes 22–42\% of the constant's error held out. The router-logit basis saturates below 19\%.
\item A tenfold ridge changes these values by less than 0.02.
\end{itemize}

\begin{table}[H]
\centering
\caption{Shared-error decomposition, OLMoE 50\%, MoNE's mask: sums over the calibration tokens of the gate-weighted residual inner products, diagonal (each pruned expert with itself) and off-diagonal (pairs co-routed to the same token, both orders), one aggregation for every operator.}
\label{tab:shared}
\footnotesize
{\begin{adjustbox}{max width=\linewidth}
\begin{tabular}{@{}llcccccc@{}}
\toprule
Slice & Operator & Total & Diagonal & Off-diagonal & Off / diag. & Removed (of constant's total) & Of which shared \\
\midrule
A & deletion & 5.742e7 & 4.961e7 & 7.82e6 & 0.158 & --- & --- \\
A & constant & 5.079e7 & 4.650e7 & 4.29e6 & 0.092 & --- & --- \\
A & tilt r8 & 4.038e7 & 3.915e7 & 1.23e6 & 0.031 & 20.5\% & 29.4\% \\
A & tilt r32 & 3.680e7 & 3.594e7 & 8.57e5 & 0.024 & 27.6\% & 24.5\% \\
B & constant & 5.109e7 & 4.680e7 & 4.29e6 & 0.092 & --- & --- \\
B & tilt r8 & 3.998e7 & 3.876e7 & 1.22e6 & 0.031 & 21.7\% & 27.6\% \\
B & tilt r32 & 3.650e7 & 3.564e7 & 8.59e5 & 0.024 & 28.5\% & 23.5\% \\
\bottomrule
\end{tabular}
\end{adjustbox}}
\end{table}
Against the constant, the rank-32 tilt multiplies the summed diagonal by 0.773 (A) and 0.762 (B) and the off-diagonal by 0.200 on both slices. A per-expert median of the diagonal fall (13\%) is a different aggregation from these summed ratios. 

\begin{table}[H]
\centering
\caption{At equal parameter cost, per-expert maps against one map shared by a layer (share of the constant's error removed, held out; percentages in the header are extra parameters per pruned expert as a share of one expert).}
\label{tab:B4}
\footnotesize
\begin{adjustbox}{max width=\linewidth}
\begin{tabular}{@{}lccccc@{}}
\toprule
Layer & Per-expert r8 (0.52\%) & Per-expert r32 (2.08\%) & Shared, full rank (2.08\%) & Shared + private r8 (2.6\%) & Share of the per-expert gain the shared map keeps \\
\midrule
0 & 8.0\% & 14.7\% & 5.0\% & 11.1\% & 12\% \\
8 & 4.3\% & 8.5\% & 2.0\% & 5.2\% & 9\% \\
12 & 7.1\% & 11.6\% & 2.4\% & 7.3\% & 12\% \\
15 & 24.3\% & 30.4\% & 13.3\% & 24.4\% & 32\% \\
\bottomrule
\end{tabular}
\end{adjustbox}
\end{table}

\textbf{Sharing loses in every layer.}

\begin{itemize}
\item A layer-shared map keeps 9–32\% of the per-expert gain.
\item Projecting the constants onto a shared subspace makes the constant's error worse at every dimension tried (+1.82\% at K = 1 to +0.35\% at K = 16).
\item So neither shared dictionaries nor ``eigen-novices'' (constants projected onto a shared subspace) improved on per-expert maps in our measurements.
\end{itemize}
\subsection{Fit quality by layer}
\label{app:B4}

\begin{table}[H]
\centering
\caption{Mean in-sample R$^{2}$ of the rank-32 tilt over each layer's pruned experts, OLMoE (share of the constant's squared error removed).}
\label{tab:B5}
\footnotesize
\begin{adjustbox}{max width=\linewidth}
\begin{tabular}{@{}lllllllllllllllll@{}}
\toprule
Setting & L0 & L1 & L2 & L3 & L4 & L5 & L6 & L7 & L8 & L9 & L10 & L11 & L12 & L13 & L14 & L15 \\
\midrule
50\%, draw A & .200 & .095 & .100 & .117 & .094 & .100 & .103 & .114 & .101 & .115 & .141 & .125 & .135 & .122 & .161 & .337 \\
50\%, draw B & .202 & .094 & .098 & .117 & .091 & .098 & .102 & .113 & .100 & .113 & .140 & .121 & .133 & .119 & .159 & .338 \\
50\%, 100 sequences & .219 & .124 & .124 & .145 & .119 & .131 & .130 & .131 & .122 & .132 & .167 & .153 & .166 & .166 & .191 & .354 \\
25\%, draw A & .234 & .086 & .110 & .094 & .102 & .108 & .110 & .124 & .098 & .111 & .145 & .112 & .109 & .117 & .137 & .334 \\
\bottomrule
\end{tabular}
\end{adjustbox}
\end{table}

\textbf{The last MoE layer is the most predictable in every model, but on OLMoE it is not where the gain comes from.}

\begin{itemize}
\item At 50\% the mean in-sample R$^{2}$ is:
\begin{itemize}
\item OLMoE 0.135, lowest 0.094 (L4), highest 0.337 (L15);
\item DeepSeek 0.146, lowest 0.105 (L8), highest 0.303 (L26);
\item Moonlight 0.131, lowest 0.084 (L11), highest 0.308 (L26).
\end{itemize}
\item On OLMoE, tilting layer 15 alone gains +0.35 of the full +3.45 (4.2 M parameters).
\item Draws A and B agree within 0.003 in almost every layer.
\end{itemize}
\subsection{Calibration size}
\label{app:B5}

\begin{table}[H]
\centering
\caption{OLMoE 50\%, MoNE mask fixed: two disjoint 100-sequence slices against the 1,000-sequence draws.}
\label{tab:B6}
\footnotesize
\begin{adjustbox}{max width=\linewidth}
\begin{tabular}{@{}lccc@{}}
\toprule
Calibration & Constant & Tilt r8 (gain) & Tilt r32 (gain) \\
\midrule
sequences 0–99 & 50.45 & 52.40 (+1.95) & 53.64 (+3.19) \\
sequences 900–999 & 49.79 & 49.70 (−0.09) & 50.45 (+0.66) \\
spread of the gain & & 2.04 & 2.52 \\
draw A (1,000) & 49.09 & 51.92 (+2.83) & 52.54 (+3.45) \\
draw B (1,000) & 49.18 & 51.98 (+2.81) & 52.82 (+3.64) \\
spread of the gain & & 0.02 & 0.19 \\
\bottomrule
\end{tabular}
\end{adjustbox}
\end{table}

\begin{table}[H]
\centering
\caption{Held-out and in-sample R$^{2}$ at 100 and 1,000 sequences, OLMoE.}
\label{tab:B7}
\footnotesize
\begin{adjustbox}{max width=\linewidth}
\begin{tabular}{@{}lccccc@{}}
\toprule
Layer & Rank & Held-out @100 & In-sample @100 & Held-out @1000 & In-sample @1000 \\
\midrule
0 & 8 / 32 & 0.065 / 0.114 & 0.096 / 0.185 & 0.080 / 0.147 & 0.091 / 0.169 \\
8 & 8 / 32 & 0.033 / 0.060 & 0.060 / 0.120 & 0.043 / 0.085 & 0.051 / 0.102 \\
12 & 8 / 32 & 0.058 / 0.086 & 0.106 / 0.184 & 0.071 / 0.116 & 0.087 / 0.149 \\
15 & 8 / 32 & 0.217 / 0.262 & 0.267 / 0.351 & 0.243 / 0.304 & 0.263 / 0.337 \\
\bottomrule
\end{tabular}
\end{adjustbox}
\end{table}

\textbf{The fit needs 1,000 sequences, and not because of overfitting.}

\begin{itemize}
\item At 100 sequences the held-out R$^{2}$ is already 78–89\% of its 1,000-sequence value. Yet the accuracy gain varies by 2.5 points between two slices.
\item 100 sequences of 4,096 tokens behave like a few hundred independent observations per expert: measured split-half noise of the constants is \textasciitilde{}100× the 1/n prediction.
\item At 1,000 sequences both the per-task pattern and the total replicate (\cref{tab:A2,tab:A4}).
\end{itemize}
\subsection{Rank 8 against rank 32}
\label{app:B6}

\begin{table}[H]
\centering
\caption{Nine-task accuracy with rank-8 tilts (gain over the constant on the same draw).}
\label{tab:B8}
\footnotesize
\begin{adjustbox}{max width=\linewidth}
\begin{tabular}{@{}lcc@{}}
\toprule
Setting & Rank 8 & Rank 32 \\
\midrule
OLMoE 50\%, A / B & 51.92 (+2.83) / 51.98 (+2.81) & 52.54 (+3.45) / 52.82 (+3.64) \\
OLMoE 25\%, A & 62.70 (+1.52) & 63.02 (+1.83) \\
DeepSeek 50\%, A / B & 59.42 (−0.42) / 59.66 (−0.11) & 60.32 (+0.48) / 60.32 (+0.55) \\
DeepSeek 25\%, A & 65.59 (+0.18) & 65.56 (+0.16) \\
Moonlight 50\%, A / B & 60.87 (+0.63) / 60.62 (+0.41) & 61.61 (+1.37) / 61.73 (+1.52) \\
\bottomrule
\end{tabular}
\end{adjustbox}
\end{table}

\textbf{Rank 8 captures most of the gain on OLMoE and little elsewhere.}

\begin{itemize}
\item On OLMoE, 82\% of the rank-32 gain at a quarter of the parameters.
\item On DeepSeek, a small loss at 50\%. On Moonlight, under half of the rank-32 gain.
\end{itemize}
\subsection{Correctness checks}
\label{app:B7}

The fitting mathematics was checked by a line-for-line NumPy port against brute force. The tests show that:

\begin{itemize}
\item streaming statistics equal the direct covariance under large offsets;
\item the full-rank fit equals OLS;
\item rank truncation equals optimal reduced-rank regression;
\item the held-out residual formula equals direct evaluation;
\item pooled analysis reduces to per-expert fits.
\end{itemize}
A further check:

\begin{itemize}
\item \textbf{Coverage.} Every evaluation first checks that every pruned expert has a fitted operator (512/512, 832/832 or 416/416) and refuses to run otherwise.
\end{itemize}
\section{Equal-budget compressed copies and protocol robustness}
\label{app:copies}

\subsection{Construction}
\label{app:C1}

Each copy replaces a pruned expert by a cheaper SwiGLU expert that keeps the nonlinearity. Each also stores a bias $b=\mathbb{E}[E(x)-E'(x)]$ over routed calibration tokens (d values), matching the tilt's stored mean.

\begin{itemize}
\item \textbf{Plain SVD:} rank-k truncated SVD of gate, up and down projections; $3k(d+m)$ values.
\item \textbf{Whitened SVD (truncation-aware, as in SVD-LLM / D$^{2}$-MoE):} SVD of $WS$ with $SS^\top=\mathbb{E}[xx^\top]$ over the expert's routed tokens (gate and up: expert input; down: the original intermediate activation). Keep $U_k\Sigma_k$ and $V_k^\top S^{-1}$. This minimizes $\mathbb{E}\|(W-W')x\|^2$ per matrix; $3k(d+m)$ values.
\item \textbf{Top neurons:} keep the k intermediate neurons with the largest $\mathbb{E}[h_j^2]\,\|W_{\mathrm{down}}[:,j]\|^2$ exactly; $3kd$ values.
\end{itemize}
\begin{table}[H]
\centering
\caption{Budgets, counting everything stored for one pruned expert: the tilt's maps, feature mean, clip scale and output mean; a copy's factors and bias. The copy's rank is the largest that fits under the tilt's $2rd$ (r = 32); the 4× arms quadruple it. Over MoNE's constant model, which already stores the mean, each figure is d = 2,048 lower.}
\label{tab:C1}
\footnotesize
\begin{adjustbox}{max width=\linewidth}
\begin{tabular}{@{}lcllll@{}}
\toprule
Model (d, m) & Tilt r32 & SVD / whitened SVD, 1× & Neurons, 1× & Whitened SVD, 4× & Neurons, 4× \\
\midrule
OLMoE (2,048, 1,024) & 133,184 & k = 14: 131,072 & k = 21: 131,072 & k = 56: 518,144 & k = 85: 524,288 \\
DeepSeek, Moonlight (2,048, 1,408) & 133,184 & k = 12: 126,464 & k = 21: 131,072 & k = 50: 520,448 & k = 85: 524,288 \\
\bottomrule
\end{tabular}
\end{adjustbox}
\end{table}

\subsection{Accuracy}
\label{app:C2}

\begin{table}[H]
\centering
\caption{Equal-budget copies at 50\% pruning (MoNE masks, draw A): differences against the constant, the tilt and rerouting at the trained gate scale (``keep''; renormalized on Moonlight). ``Extra'' is what the copies store, bias included; on that basis the tilt stores 68.2 M / 110.8 M / 110.8 M.}
\label{tab:C2}
\footnotesize
\begin{adjustbox}{max width=\linewidth}
\begin{tabular}{@{}llccccc@{}}
\toprule
Model & Copy & Extra & Acc. & − constant & − tilt & − keep \\
\midrule
OLMoE & Top 21 neurons (1×) & 67.1 M & 52.63 & +3.54 [+2.75, +4.34]* & +0.09 [−0.78, +0.95] & +1.15 [+0.16, +2.16]* \\
 & Whitened SVD k14 (1×) & 67.1 M & 48.76 & −0.33 [−0.99, +0.30] & −3.78 [−4.65, −2.96]* & −2.72 [−3.71, −1.72]* \\
 & Plain SVD k14 (1×) & 67.1 M & 48.75 & −0.35 [−0.89, +0.20] & −3.80 [−4.58, −3.04]* & −2.73 [−3.72, −1.73]* \\
 & Top 85 neurons (4×) & 268.4 M & 54.04 & +4.95 [+3.94, +5.94]* & +1.50 [+0.44, +2.55]* & +2.56 [+1.44, +3.70]* \\
 & Whitened SVD k56 (4×) & 265.3 M & 49.75 & +0.66 [−0.12, +1.47] & −2.79 [−3.68, −1.91]* & −1.72 [−2.75, −0.67]* \\
\midrule
DeepSeek & Top 21 neurons (1×) & 109.1 M & 60.18 & +0.34 [−0.27, +0.93] & −0.14 [−0.83, +0.56] & −1.02 [−1.84, −0.22]* \\
 & Whitened SVD k12 (1×) & 105.2 M & 59.81 & −0.02 [−0.47, +0.48] & −0.51 [−1.08, +0.10] & −1.38 [−2.11, −0.66]* \\
 & Plain SVD k12 (1×) & 105.2 M & 59.42 & −0.42 [−0.77, −0.06]* & −0.90 [−1.42, −0.37]* & −1.78 [−2.52, −1.04]* \\
\midrule
Moonlight & Top 21 neurons (1×) & 109.1 M & 60.94 & +0.70 [+0.15, +1.25]* & −0.66 [−1.27, −0.09]* & +0.88 [−0.02, +1.77] \\
 & Plain SVD k12 (1×) & 105.2 M & 60.81 & +0.57 [−0.04, +1.19] & −0.80 [−1.53, −0.07]* & +0.75 [−0.20, +1.68] \\
 & Whitened SVD k12 (1×) & 105.2 M & 60.41 & +0.17 [−0.32, +0.70] & −1.19 [−1.84, −0.55]* & +0.35 [−0.56, +1.27] \\
\bottomrule
\end{tabular}
\end{adjustbox}
\end{table}

\textbf{No copy beats keep-scale rerouting on DeepSeek, which stores nothing.} On OLMoE, the neuron copies beat rerouting at both budgets.

 At the 1$\times$ budget the tilt and the neuron copy are not detectably different on OLMoE and DeepSeek, where post-hoc bounds rule out differences beyond 0.81 and 0.73 points (\cref{tab:tost}); the tilt is ahead on Moonlight.

\subsection{Fit quality}
\label{app:C3}

\begin{table}[H]
\centering
\caption{Mean in-sample R$^{2}$ against the constant ($1-\mathrm{SSE}/\mathrm{SSE}_{\mathrm{const}}$) over all pruned experts at 50\%.}
\label{tab:C3}
\footnotesize
\begin{adjustbox}{max width=\linewidth}
\begin{tabular}{@{}lccccccc@{}}
\toprule
Model & Tilt r32 & Whitened SVD 1× & Plain SVD 1× & Neurons 1× & Whitened SVD 4× & Plain SVD 4× & Neurons 4× \\
\midrule
OLMoE & 0.135 & 0.048 & 0.006 & 0.091 & 0.130 & 0.041 & 0.203 \\
DeepSeek & 0.146 & 0.045 & −0.001 & 0.099 & 0.129 & 0.022 & 0.216 \\
Moonlight & 0.131 & 0.046 & 0.002 & 0.065 & 0.118 & 0.014 & 0.152 \\
\bottomrule
\end{tabular}
\end{adjustbox}
\end{table}

\textbf{Truncating an expert's weights without regard to its inputs preserves almost nothing of its output.}

\begin{itemize}
\item Plain SVD copies explain $\le$ 0.6\% of the constant's error at 1× and $\le$ 4.1\% at 4×.
\item Whitening multiplies that several-fold.
\item At 1× the order is tilt > neurons > whitened SVD > plain SVD on every model.
\item At 4× the neuron copy fits better than the rank-32 tilt on every model.
\end{itemize}
Caveat:

\begin{itemize}
\item \textbf{Different fit sets.} The tilt's R$^{2}$ is computed over all fit batches, while the copies' R$^{2}$ is measured on the bias-pass batches of the same cache. Both are in-sample.
\item \textbf{Different objectives.} The tilt minimizes exactly this error, while the copies approximate the weights.
\end{itemize}

\textbf{A better fit to the expert does not by itself buy accuracy.} Within one kind of copy, more fit brings more accuracy. Across kinds it does not: on average the tilt fits the pruned experts better than the equal-budget neuron copy on every model, yet it wins on accuracy only on Moonlight, where its fit lead is largest (2.0× against 1.5×). Part of the tilt's lead in fit is by construction, since it minimizes exactly this squared error while the copies approximate the weights.
\subsection{Language modelling}
\label{app:C4}

\begin{table}[H]
\centering
\caption{Share of the constant's log-perplexity damage recovered on held-out text (64 × 4,096 tokens of C4 and of Zyda-2 slice C), 50\% pruning. Tilt shares on draws A / B.}
\label{tab:C4}
\footnotesize
\begin{adjustbox}{max width=\linewidth}
\begin{tabular}{@{}llcccc@{}}
\toprule
Model & Corpus (dense NLL) & Tilt r32 & Reroute & Whitened SVD 1× & Neurons 1× \\
\midrule
OLMoE & C4 (2.5113) & 23.7\% / 23.8\% & 12.4\% & 16.1\% & 28.0\% \\
 & Zyda-2 C (2.2660) & 22.3\% / 22.5\% & 9.3\% & 14.2\% & 26.6\% \\
\midrule
DeepSeek & C4 (2.3604) & 13.9\% / 14.3\% & 29.8\% & −2.7\% & 13.4\% \\
 & Zyda-2 C (2.3114) & 14.1\% / 14.1\% & 27.2\% & −3.1\% & 14.6\% \\
\midrule
Moonlight & C4 (2.5604) & 33.8\% / 34.2\% & 18.1\% & 15.3\% & 26.3\% \\
 & Zyda-2 C (2.5052) & 31.5\% / 32.7\% & 15.2\% & 19.0\% & 27.4\% \\
\bottomrule
\end{tabular}
\end{adjustbox}
\end{table}

\textbf{On perplexity, too, the neuron copy is close to the tilt.} It is ahead on OLMoE, similar on \dsv{} and behind on Moonlight. The whitened-SVD copy is behind the tilt everywhere. \textbf{Even the sign of a perplexity effect can depend on the text:} rerouting worsens Moonlight's WikiText perplexity but recovers 15--18\% of the damage on C4 and Zyda-2.

\subsection{Scoring protocol}
\label{app:scoring}\label{app:C5}

\begin{table}[H]
\centering
\caption{Operator gain over the constant under four scoring choices (all from the same per-item dumps). Columns: as reported (MoNE's metrics: acc\_norm for ARC, OBQA, PIQA); acc on every task; the six tasks whose answer options change per item (ARC-c, ARC-e, COPA, OBQA, PIQA, WinoGrande); the three fixed-answer tasks (BoolQ, MMLU, RTE).}
\label{tab:C5}
\footnotesize
\begin{adjustbox}{max width=\linewidth}
\begin{tabular}{@{}llcccc@{}}
\toprule
Setting & Comparison & As reported & acc everywhere & Six per-item-choice & Three fixed-answer \\
\midrule
OLMoE 50\% & tilt (A) & +3.45 [+2.62, +4.27]* & +3.66 [+2.85, +4.47]* & +5.15 [+4.02, +6.32]* & +0.04 [−0.89, +0.92] \\
 & tilt (B) & +3.64 [+2.86, +4.46]* & +3.89 [+3.10, +4.68]* & +5.08 [+3.96, +6.23]* & +0.76 [−0.04, +1.54] \\
 & tilt (C) & +3.44 [+2.65, +4.25]* & +3.71 [+2.94, +4.51]* & +4.78 [+3.65, +5.92]* & +0.77 [−0.07, +1.56] \\
 & reroute (A) & +2.38 [+1.42, +3.33]* & +2.48 [+1.53, +3.41]* & +3.34 [+2.26, +4.42]* & +0.47 [−1.26, +2.31] \\
\midrule
OLMoE 25\% & tilt (A) & +1.83 [+1.21, +2.47]* & +1.40 [+0.79, +2.03]* & +1.72 [+1.08, +2.39]* & +2.06 [+0.77, +3.38]* \\
 & tilt (B) & +1.85 [+1.19, +2.55]* & +1.68 [+1.01, +2.37]* & +1.55 [+0.85, +2.32]* & +2.44 [+0.99, +3.86]* \\
 & reroute (A) & +1.46 [+0.85, +2.11]* & +1.31 [+0.72, +1.95]* & +0.90 [+0.25, +1.62]* & +2.59 [+1.38, +3.89]* \\
\midrule
DeepSeek 50\% & tilt (A) & +0.48 [−0.03, +0.99] & +0.31 [−0.21, +0.82] & +0.32 [−0.29, +0.87] & +0.81 [−0.18, +1.83] \\
 & tilt (B) & +0.55 [+0.06, +1.06]* & +0.20 [−0.29, +0.69] & +0.84 [+0.28, +1.43]* & −0.03 [−0.97, +0.92] \\
 & tilt (C) & +0.50 [−0.04, +1.06] & +0.28 [−0.27, +0.84] & +0.40 [−0.28, +1.06] & +0.71 [−0.27, +1.71] \\
 & reroute (A) & +1.35 [+0.65, +2.08]* & +1.39 [+0.69, +2.11]* & +1.44 [+0.63, +2.32]* & +1.17 [−0.24, +2.58] \\
\midrule
DeepSeek 25\% & tilt (A) & +0.16 [−0.28, +0.56] & −0.04 [−0.48, +0.37] & −0.11 [−0.61, +0.32] & +0.68 [−0.15, +1.57] \\
 & tilt (B) & +0.74 [+0.16, +1.36]* & +0.59 [−0.00, +1.21] & +0.53 [−0.16, +1.31] & +1.15 [+0.14, +2.21]* \\
 & reroute (A) & +0.26 [−0.35, +0.87] & +0.18 [−0.43, +0.79] & +0.29 [−0.45, +1.06] & +0.19 [−0.86, +1.28] \\
\midrule
Moonlight 50\% & tilt (A) & +1.37 [+0.73, +2.01]* & +1.04 [+0.41, +1.71]* & +1.51 [+0.76, +2.31]* & +1.08 [−0.09, +2.20] \\
 & tilt (B) & +1.52 [+0.90, +2.17]* & +1.05 [+0.41, +1.71]* & +1.47 [+0.73, +2.30]* & +1.62 [+0.53, +2.72]* \\
 & tilt (C) & +1.57 [+0.91, +2.24]* & +1.41 [+0.74, +2.10]* & +1.41 [+0.61, +2.29]* & +1.89 [+0.82, +2.97]* \\
 & reroute (A) & −0.18 [−1.04, +0.68] & −0.14 [−1.01, +0.75] & +0.74 [−0.23, +1.76] & −2.02 [−3.65, −0.39]* \\
\midrule
Moonlight 25\% & tilt (A) & +0.13 [−0.16, +0.42] & +0.15 [−0.13, +0.45] & −0.25 [−0.57, +0.07] & +0.89 [+0.30, +1.48]* \\
 & tilt (B) & +0.26 [−0.07, +0.58] & +0.19 [−0.11, +0.49] & −0.01 [−0.38, +0.35] & +0.80 [+0.12, +1.48]* \\
 & reroute (A) & +0.05 [−0.29, +0.39] & +0.00 [−0.34, +0.34] & +0.16 [−0.23, +0.55] & −0.16 [−0.86, +0.51] \\
\bottomrule
\end{tabular}
\end{adjustbox}
\end{table}

\textbf{On Moonlight, rerouting loses on the fixed-answer tasks ($-$2.02*), while its per-item-choice tasks show no detectable difference.} The loss is mainly BoolQ (\cref{tab:A2}).

\textbf{At 25\% on Moonlight the tilt's only significant effect is on fixed-answer tasks.} That effect is mostly MMLU.

\subsection{Few-shot prompting}
\label{app:C6}

\begin{table}[H]
\centering
\caption{Gains at 0-shot and 5-shot (nine tasks, 50\%, draw A) and their ratio.}
\label{tab:C6}
\footnotesize
\begin{adjustbox}{max width=\linewidth}
\begin{tabular}{@{}lccc@{}}
\toprule
Model & Tilt − const, 0 / 5-shot (ratio) & Reroute − const, 0 / 5-shot (ratio) & Tilt − reroute, 5-shot \\
\midrule
OLMoE & +3.45* / +3.01 [+2.25, +3.85]* (0.87) & +2.38* / +2.03 [+1.16, +2.90]* (0.85) & +0.98 [+0.12, +1.87]* \\
DeepSeek & +0.48 / −0.29 [−0.88, +0.29] (−0.60) & +1.35* / +2.46 [+1.76, +3.17]* (1.82) & −2.75 [−3.53, −2.00]* \\
Moonlight & +1.37* / +1.54 [+0.70, +2.38]* (1.12) & −0.18 / +1.28 [+0.42, +2.16]* (−7.14) & +0.25 [−0.57, +1.08] \\
\bottomrule
\end{tabular}
\end{adjustbox}
\end{table}

\section{MoRA re-implementation}
\label{app:mora}

\subsection{What was re-implemented, and the settings the paper leaves open}
\label{app:D1}

MoRA \citep{sun2026mora} learns one router bias per routed expert on $L_{\mathrm{LM}}+\lambda_d L_{\mathrm{Div}}$ with all weights frozen. Each layer keeps its top-B experts by bias. Each pruned expert can then be answered by a retained proxy expert passed through a channel-wise affine map, $a\odot f_j(x)+b$. The map is fitted by weighted ridge regression and used only if its held-out R$^{2}$ (q) exceeds $q_{\min}=0.4$. Its code is announced but was not released by 27 October 2026, so we re-implemented it from the paper and unit-tested each component.

\textbf{Settings the paper does not give, and our values:}

\begin{itemize}
\item $\alpha$ = $\tau$ = 1, so that s = 0 reproduces the pretrained router;
\item Adam with learning rate 10$^{-2}$, 1,000 steps of 4 × 1,024 C4 tokens;
\item the normalization of $L_{\mathrm{Div}}$ (\cref{app:D2});
\item token weights $\omega$ = the pretrained router's probability for the target expert, with two variants checked;
\item the fallback for experts without a proxy: renormalized, plus keep-scale as a variant.
\end{itemize}
\textbf{Self-test.} With s = 0, the differentiable MoE forward reproduces the original blocks exactly on DeepSeek-V2-Lite (V2 gate) and Moonlight (V3 gate): identical expert choice and output error 0.00. A finite-difference check of the bias gradient passes.

We did not run the MC-SMoE and HC-SMoE baselines; \cref{app:merging} compares our operators with the numbers MoRA reports for them.

\subsection{The diversity loss's normalization decides the method}
\label{app:D2}

\begin{table}[H]
\centering
\caption{Gradient balance at step 1, $|\lambda\nabla L_{\mathrm{Div}}|/|\nabla L_{\mathrm{LM}}|$, and accuracy on MoRA's nine tasks (MoRA-9, which has HellaSwag instead of COPA) for three normalizations of Eq. 6. All arms run without approximation, with renormalized fallback.}
\label{tab:D1}
\footnotesize
\begin{adjustbox}{max width=\linewidth}
\begin{tabular}{@{}lccccc@{}}
\toprule
Normalization of $L_{\mathrm{Div}}$ & Gradient ratio, DeepSeek / Moonlight & DeepSeek 25\% & DeepSeek 50\% & Moonlight 25\% & Moonlight 50\% \\
\midrule
pair-mean per layer, summed over layers (primary) & 0.316 / 0.647 & 62.14 & 51.16 & 66.32 & 50.98 \\
sum, as Eq. 6 is written & 1.27 × 10$^{3}$ / 2.61 × 10$^{3}$ & 59.46 & 42.85 & 62.20 & 49.91 \\
mean over layers too & 0.0121 / 0.0249 & 60.40 & 45.60 & 64.05 & 53.17 \\
\bottomrule
\end{tabular}
\end{adjustbox}
\end{table}

\textbf{Summed literally, the diversity term is about 10$^{3}$ times the language-modelling gradient, so the learned biases almost entirely encode router geometry.}

\begin{itemize}
\item \textbf{The best normalization differs between settings,} and the choice moves MoRA-9 by 1–8 points: pair-mean − sum is +2.68*, +8.31*, +4.12* and +1.08*; pair-mean − mean is +1.74*, +5.56*, +2.27* and −2.18*.
\item \textbf{The rankings had not settled at step 1,000.} The share of the final kept set already present at step 900 was 0.942 / 0.797 on DeepSeek at 25\% / 50\% under the primary normalization. It was lower still when the LM term dominates: 0.871 / 0.691.
\end{itemize}
\subsection{In our re-implementation, almost no retained expert predicts a pruned expert's output}
\label{app:D3}\label{sec:mora}

\begin{table}[H]
\centering
\caption{Proxy fits on held-out C4 tokens, our re-implementation: pruned experts whose best retained-expert proxy passes MoRA's threshold ($q>0.4$), and the distribution of $q$. Fit on blocks 1,000–1,015, scored on 1,016–1,031 (2 × 64 × 1,024 tokens).}
\label{tab:D2}\label{tab:mora}
\footnotesize
\begin{adjustbox}{max width=\linewidth}
\begin{tabular}{@{}lccccc@{}}
\toprule
Mask & Targets & q > 0.4 & q at 10 / 50 / 90\% & Best in-sample R$^{2}$, median & Constant novice R$^{2}$, median \\
\midrule
DeepSeek 25\%, MoRA & 416 & 1 & 0.000 / 0.000 / 0.003 & 0.003 & −0.003 \\
DeepSeek 50\%, MoRA & 832 & 3 & 0.000 / 0.000 / 0.001 & 0.003 & −0.004 \\
DeepSeek 25\%, MoNE-C4 & 416 & 0 & 0.000 / 0.000 / 0.001 & 0.002 & −0.003 \\
DeepSeek 50\%, MoNE-C4 & 832 & 3 & 0.000 / 0.000 / 0.001 & 0.002 & −0.003 \\
Moonlight 25\% / 50\%, MoRA & 416 / 832 & 0 / 0 & 0.000 / 0.000 / $\le$ 0.001 & 0.002 & −0.004 \\
Moonlight 25\% / 50\%, MoNE-C4 & 416 / 832 & 0 / 0 & 0.000 / 0.000 / 0.001 & 0.002 & −0.003 / −0.004 \\
\bottomrule
\end{tabular}
\end{adjustbox}
\end{table}

\textbf{Positive control.}

\begin{itemize}
\item In layers 1, 14 and 26 of each model, four retained experts were overwritten with copies of pruned experts whose output channels were rescaled by known factors s $\in$ [0.5, 2].
\item The unchanged fitting pipeline recovered all 24 planted pairs at q = 1.0000, each with the planted expert chosen as proxy:
\begin{itemize}
\item median |a$\cdot$s − 1| = 0.0016–0.0019 (bf16 precision);
\item median |b| $\le$ 9.3 × 10$^{-4}$.
\end{itemize}
\item The best real candidate for the same targets scored at most 0.0164.
\end{itemize}
\textbf{The near-zero fits are not a failure of the fitting pipeline.} Because we report the best retained expert rather than one chosen by a rule, they bound what a channel-wise proxy can do under our fitting protocol, not its fidelity to every setting MoRA leaves unstated (\cref{app:D1}).

\textbf{Effect on accuracy.}

\begin{itemize}
\item \textbf{MoRA mask, with against without approximation (MoRA-9).} +0.19 [−0.02, +0.41] at DeepSeek 25\% and +0.09 [−0.38, +0.56] at 50\%. On Moonlight the two models are identical, since no proxy is accepted.
\item \textbf{MoNE-C4 mask (DeepSeek).} The ``proxy'' arm loses to the constant (−0.83* at 25\%, −0.79 at 50\%). This is not the proxies' doing: at 25\% no proxy is accepted and at 50\% three are. The arm is therefore renormalized rerouting with at most three proxies, and its loss comes from renormalization, not from the proxies.
\end{itemize}
\subsection{Reproduction of the paper's rows}
\label{app:D4}\label{app:mora-repro}

\begin{table}[H]
\centering
\caption{MoRA-9 accuracy, ours / paper.}
\label{tab:D3}
\footnotesize
\begin{adjustbox}{max width=\linewidth}
\begin{tabular}{@{}lccccccc@{}}
\toprule
Setting & Dense & MoNE (C4) & Activation count & Avg. gate mass & r$'$ gate mass & MoRA & MoRA w/o approx. \\
\midrule
DeepSeek 25\% & 66.03 / 66.05 & 64.17 / 63.25 & 61.41 / 62.22 & 62.21 / 62.60 & 62.10 / 62.79 & 62.33 / 63.68 & 62.14 / 63.25 \\
DeepSeek 50\% & 66.03 / 66.05 & 53.37 / 53.64 & 52.54 / 46.27 & 51.33 / 50.98 & 52.08 / 51.38 & 51.25 / 54.40 & 51.16 / 54.02 \\
Moonlight 25\% & 69.95 / 70.07 & 66.74 / 66.23 & 65.37 / 65.50 & 66.78 / 66.47 & 66.64 / 67.07 & 66.32 / 67.48 & 66.32 / 67.19 \\
Moonlight 50\% & 69.95 / 70.07 & 55.78 / 54.39 & 53.72 / 53.54 & 56.36 / 54.33 & 53.08 / 54.58 & 50.98 / 55.36 & 50.98 / 54.82 \\
\bottomrule
\end{tabular}
\end{adjustbox}
\end{table}

\textbf{The pipeline reproduces the paper's dense and simple-baseline rows; MoRA's own rows fall short by 1–4 points, and our runs of two baselines score higher than MoRA's rows for them; we cannot explain the gap.}

\begin{itemize}
\item Dense matches to within 0.12, with no task off by more than 1.1.
\item Activation count matches to within 0.8 in three of four settings. The exception is DeepSeek 50\% (+6.28), where the paper's value is far below ours on ARC-e, ARC-c, HellaSwag and PIQA.
\item Our MoNE-C4 matches MoNE's own published C4 table but exceeds MoRA's MoNE row (+0.92 at DeepSeek 25\%, +1.39 at Moonlight 50\%).
\item Average gate mass exceeds MoRA's row at Moonlight 50\% (+2.03).
\item MoRA itself reproduces the paper's per-task pattern at 50\%, but not its level (−1.35 / −3.16 on DeepSeek, −1.16 / −4.38 on Moonlight). We cannot tell what explains the gap; the settings the paper leaves unstated (\cref{app:D1}) are candidates.
\end{itemize}
\subsection{Gate scale on MoRA's masks}
\label{app:mora-semantics}
MoRA's method section assumes a softmax router with normalized selected weights and removes pruned experts from routing \citep{sun2026mora}. In our re-implementation, which follows that formulation, MoRA's own \dsv{} masks score higher with keep-scale rerouting than with renormalized rerouting on MoRA's nine tasks (C4 calibration): +0.47* at 25\% and +1.24* at 50\%. On Moonlight, whose router is trained with renormalization, that formulation's renormalized gates are the trained scale. COMPEL's method section writes the gates of \dsv{} and OLMoE as a softmax renormalized over the top-$k$ \citep{yoon2026compel}, which neither router does; without its code we cannot tell which semantics its experiments used.

\subsection{Against our operators on MoRA's tasks}
\label{app:D6}\label{app:mora-tasks}

\begin{table}[H]
\centering
\caption{MoRA-9 for our Zyda-calibrated operators (HellaSwag added to the runs of \cref{sec:operators}) against C4-calibrated runs and MoRA's published numbers. Keep: rerouting at the trained gate scale.}
\label{tab:D6}
\footnotesize
\begin{adjustbox}{max width=\linewidth}
\begin{tabular}{@{}lccccc@{}}
\toprule
Setting & MoNE + constant (Zyda) & MoNE + tilt r32 (Zyda) & MoNE + keep (Zyda) & MoNE-C4 + keep & MoRA (paper) \\
\midrule
DeepSeek 25\% & 63.79 & 64.04 & 64.13 & 64.29 & 63.68 \\
DeepSeek 50\% & 57.52 & 58.00 & 59.16 & 55.41 & 54.40 \\
Moonlight 25\% & 67.32 & 67.47 & 67.40 & 66.90 & 67.48 \\
Moonlight 50\% & 58.19 & 59.54 & 58.40 & 55.89 & 55.36 \\
\bottomrule
\end{tabular}
\end{adjustbox}
\end{table}

\textbf{On MoRA's own task set our best configuration matches or beats the paper's MoRA in every setting (point estimates, one run each); at 50\% most of the margin is the calibration corpus.}

\begin{itemize}
\item At 50\%, MoNE with constants gains 2.4 (Moonlight) and 4.2 (DeepSeek) MoRA-9 points from Zyda-2 instead of C4.
\item With the corpus held fixed, the operator's own contribution is tilt +1.35 over the constant on Moonlight 50\% and keep +1.64 on DeepSeek 50\%.
\item C4 and Zyda-2 masks share 86.2\% of pruned experts at DeepSeek 50\%. Yet MMLU with constants is 22.95 (C4, chance) against 40.98 (Zyda-2).
\end{itemize}
\textbf{Caveats.}

\begin{itemize}
\item \textbf{One run per setting.} Each setting has one bias-learning run and one C4 draw, so intervals are item-level only and the two-draw rule is not met.
\item \textbf{Our hyperparameter values.} Every hyperparameter MoRA leaves unstated is our choice.
\item \textbf{Qwen3 not run.} Qwen3-30B-A3B, where the paper reports its largest margin, was not run.
\end{itemize}
\section{Selection}
\label{app:selection}

All arms in this appendix are on OLMoE or DeepSeek at 50\% unless stated otherwise.

\subsection{MoNE's mask against seven other published criteria}
\label{app:E1}

Each criterion is computed exactly, per token: a pre-hook re-derives each block's routing (slots per token 8.000 / 6.000) and runs every expert on its routed tokens. Every mask is evaluated with keep-scale rerouting, which fits nothing, so the spread between draws is mask variance alone.

\begin{table}[H]
\centering
\caption{Nine-task accuracy, slices A / B, with keep-scale rerouting. We call a criterion better (worse) than MoNE's only if its two-draw mean exceeds (falls below) MoNE's by more than both spreads.}
\label{tab:E1}
\footnotesize
\begin{adjustbox}{max width=\linewidth}
\begin{tabular}{@{}llcccc@{}}
\toprule
Criterion & Score (family $(b,\alpha,\beta)$ of \citet{liu2026unified}) & OLMoE A / B (mean) & $\Delta$ vs MoNE & DeepSeek A / B (mean) & $\Delta$ vs MoNE \\
\midrule
MoNE & frequency × output std & 51.48 / 50.59 (51.03) & — & 61.19 / 61.21 (61.20) & — \\
SGE & $\Sigma$ g$^{2}$$\|$f$\|$² (0, 2, 2) & 53.08 / 51.13 (52.11) & +1.07, not resolved & 61.74 / 62.02 (61.88) & +0.68, better \\
REAP & mean $\hat{g}$$\|$f$\|$ (1, 1, 1) & 49.00 / 49.42 (49.21) & −1.82, worse & 60.46 / 61.06 (60.76) & −0.44, not resolved \\
MSAN & mean $\|$f$\|$² (1, 0, 2) & 48.04 / 47.79 (47.92) & −3.11, worse & 60.40 / 60.01 (60.21) & −0.99, worse \\
MAN & mean $\|$f$\|$ (1, 0, 1) & 47.70 / 47.36 (47.53) & −3.50, worse & 59.98 / 60.85 (60.41) & −0.79, not resolved \\
EAN & $\Sigma$ $\|$f$\|$ (0, 0, 1) & 49.39 / — & — & — & — \\
SEER & $\Sigma$ g (0, 1, 0) & 45.28 / — & — & — & — \\
Frequency & $\Sigma$ 1 (0, 0, 0) & 43.88 / 42.99 (43.44) & −7.60, worse & 55.18 / 54.83 (55.00) & −6.19, worse \\
\bottomrule
\end{tabular}
\end{adjustbox}
\end{table}

\textbf{Among published criteria, MoNE's mask is hard to beat.}

\begin{itemize}
\item Only SGE beats it, and only on DeepSeek (+0.68; big-5 +1.22).
\item The criteria recommended for task-agnostic pruning (MAN, MSAN) lose to it on OLMoE by 3.1–3.5 points. Our protocol and task set differ from those of \citet{liu2026unified} (50\% here, multiple choice only).
\item On DeepSeek, frequency fails because it prunes a super expert (\cref{app:E8}).
\item Scoring SGE on both draws together (2,000 batches) lands below the better single draw on OLMoE (51.42 against 53.08 / 51.13) and below both on DeepSeek (61.44 against 61.74 / 62.02). More calibration data does not average the variance away.
\end{itemize}
\subsection{Stability}
\label{app:E2}

\begin{table}[H]
\centering
\caption{Share of pruned experts identical across draws A and B (50\%).}
\label{tab:E2}
\footnotesize
\begin{adjustbox}{max width=\linewidth}
\begin{tabular}{@{}llcc@{}}
\toprule
Criterion & Family & OLMoE & DeepSeek \\
\midrule
MAN & mean over routed tokens & 98.2\% & 98.2\% \\
MSAN & mean over routed tokens & 97.7\% & 97.4\% \\
REAP & mean over routed tokens & 97.1\% & 97.6\% \\
EAN & carries frequency & 94.1\% & 95.9\% \\
MoNE & carries frequency & 93.6\% & 95.8\% \\
Frequency & carries frequency & 92.6\% & 94.2\% \\
SGE & carries frequency & 91.6\% & 95.6\% \\
SEER & carries frequency & 89.5\% & 95.0\% \\
\bottomrule
\end{tabular}
\end{adjustbox}
\end{table}

\textbf{Routing frequency is the draw-dependent part of a selection score.}

\begin{itemize}
\item Criteria that average over routed tokens keep 97–98\% of their pruned experts across draws. Those that carry frequency keep 89–96\%.
\item On OLMoE the stability order almost exactly matches the order of accuracy spreads.
\item Stability does not mean accuracy: the most stable criteria select worse masks here.
\item \citet{liu2026unified} report the same instability across calibration domains.
\end{itemize}
\subsection{REAP with its own semantics}
\label{app:E3}

\textbf{REAP's own routing semantics understate its own mask on these routers.} On the same REAP mask, renormalized rerouting (REAP's semantics; REAP did not evaluate these models) loses to keep-scale:

\begin{itemize}
\item OLMoE: 1.88 points (big-5 2.21; 47.12 against 49.00);
\item DeepSeek: 3.88 points (big-5 5.87; 56.58 against 60.46).
\end{itemize}
\subsection{SGE mask with the tilt}
\label{app:E4}

\begin{table}[H]
\centering
\caption{SGE mask + tilt minus MoNE mask + tilt, OLMoE 50\%, three draws.}
\label{tab:E3}
\footnotesize
\begin{adjustbox}{max width=\linewidth}
\begin{tabular}{@{}lccccc@{}}
\toprule
Draw & MoNE + tilt & SGE + tilt & Nine-task [95\% CI] & Big-5 & MMLU \\
\midrule
A & 52.54 & 54.34 & +1.79 [+0.73, +2.86]* & +2.00* & +2.61* \\
B & 52.82 & 53.33 & +0.51 [−0.60, +1.63] & −0.16 & +3.58* \\
C & 52.70 & 53.21 & +0.51 [−0.52, +1.51] & −0.25 & +5.03* \\
\bottomrule
\end{tabular}
\end{adjustbox}
\end{table}

\textbf{What survives three draws is knowledge.}

\begin{itemize}
\item MMLU is significant every time.
\item The nine-task advantage is significant only on draw A, which was the outlier.
\end{itemize}
\subsection{Operator- and interaction-aware selection}
\label{app:E5}

We scored each pruned expert's replacement residual, and every pair of co-routed pruned experts' residual inner products, under four operators: deletion, constant, rank-8 and rank-32 tilt. This gives a per-layer matrix F. We then selected masks per expert (diagonal) or by a per-layer quadratic program (QP). All results are for OLMoE 50\% on two draws.

\begin{table}[H]
\centering
\caption{Interaction structure and the resulting masks.}
\label{tab:E4}
\footnotesize
\begin{adjustbox}{max width=\linewidth}
\begin{tabular}{@{}lcc@{}}
\toprule
Quantity & Draw A & Draw B \\
\midrule
Off-diagonal / diagonal of F on MoNE's mask: deletion / constant / r8 / r32 & 0.158 / 0.092 / 0.031 / 0.024 & 0.155 / 0.092 / 0.031 / 0.024 \\
Mask overlap, constant-optimal vs tilt-optimal (diagonal / QP) & 93.9\% / 94.5\% & 94.1\% / 93.8\% \\
Consistency of the experts the operator moves & 0.44 & 0.46 \\
QP-constant mask + constant (SGE mask + constant) & 50.74 (49.81) & 50.74 (49.99) \\
\bottomrule
\end{tabular}
\end{adjustbox}
\end{table}

\textbf{Interaction-aware selection for the constant gives a strong mask.} The QP-constant mask scores 50.74 with the constant on both draws, above SGE's mask with the constant (49.81 and 49.99) and MoNE's (49.09 and 49.18; point estimates).

\textbf{By layer error, choosing the mask for the tilt instead changes little, and what it changes does not replicate.}

\begin{itemize}
\item The operator moves about 6\% of the mask, against 8–9\% from a new calibration draw.
\end{itemize}
\textbf{Layer error is a weak guide to model-level damage.} Boundary swaps measured against the change in KL to the dense model gave a mean within-layer Spearman of:

\begin{itemize}
\item 0.13 for the constant (retest 0.23);
\item 0.50 for the tilt (retest 0.50).
\end{itemize}
Layer 10 reached about 0.95 under both operators, and layer 0 was negative under both.

\textbf{Selecting the mask for the tilt costs knowledge.}

\begin{itemize}
\item The QP-tilt mask lowers MMLU by 2.6–4.3 points to chance level, under both operators, while the nine-task average stays similar.
\item The 28–32 experts it prunes differently are those the tilt reconstructs best (median residual ratio 0.79 against 0.87).
\item Being easy to imitate by layer error is not the same as being safe to prune.
\end{itemize}
\subsection{Fisher-weighted selection}
\label{app:E6}

Scoring residuals by the model's own sampled-label Fisher sensitivity instead of Euclidean layer error was tested on OLMoE 50\% with two draws.

\begin{table}[H]
\centering
\caption{Fisher-weighted selection, OLMoE 50\%, two draws.}
\label{tab:E5}
\footnotesize
\begin{adjustbox}{max width=\linewidth}
\begin{tabular}{@{}ll@{}}
\toprule
Question & Result \\
\midrule
Does Fisher sensitivity predict the measured $\Delta$KL of boundary swaps better than layer error? & No. Spearman, constant: $-$0.04 vs 0.16; tilt: 0.20 vs 0.52 (repeatability ceiling 0.94 / 0.83) \\
Does it rank the experts behind the QP-tilt MMLU loss as costly? & Yes, on both draws: median within-layer percentile 81 / 66 vs 52 / 46; layer error ranks them cheap (44 / 44) \\
Do Fisher masks avoid pruning those experts? & Partly: they still prune 10 of 28 (A) and 14 of 32 (B) \\
Does a Fisher mask beat simple masks on both draws? & No. Best: Fisher QP-tilt + tilt 53.45 / 51.07 (spread 2.38), against QP-constant + tilt 54.25 \\
\bottomrule
\end{tabular}
\end{adjustbox}
\end{table}

\textbf{Fisher masks trade general accuracy for knowledge.}

\begin{itemize}
\item MMLU reaches 32.7–33.8 on four of five Fisher arms, against 27.5–30.8 for the other masks.
\item BoolQ, COPA, OBQA, PIQA and WinoGrande fall (big-5 −2.2 against SGE + tilt).
\item The masks are draw-unstable (spreads 2.0–4.6).
\end{itemize}
\subsection{Where the knowledge-critical experts sit}
\label{app:E7}

\begin{table}[H]
\centering
\caption{MMLU with constants on the QP-constant mask when one layer group is swapped in from the QP-tilt mask.}
\label{tab:E6}
\footnotesize
\begin{adjustbox}{max width=\linewidth}
\begin{tabular}{@{}lcc@{}}
\toprule
Mask & Draw A & Draw B \\
\midrule
QP-constant & 29.21 & 28.71 \\
\dots{} with the early group (L0–L5) from QP-tilt & 26.93 & 26.24 \\
\dots{} with the middle group (L6–L10) from QP-tilt & 26.31 & 27.23 \\
\dots{} with the late group (L11–L15) from QP-tilt & 29.20 & 28.56 \\
QP-tilt & 24.88 & 26.12 \\
\bottomrule
\end{tabular}
\end{adjustbox}
\end{table}

\textbf{The experts MMLU needs sit in the early and middle layer groups: swapping either group in lowers MMLU, while swapping the late group does not.} Fourteen of the moved experts move on both draws, spread over all layers:

\begin{itemize}
\item L0 E36, E52; L2 E24; L3 E23, E54; L5 E15;
\item L6 E40, E50; L8 E24;
\item L10 E60; L11 E5; L12 E44; L14 E38; L15 E12.
\end{itemize}
Their output magnitudes are ordinary (median percentile 52), so magnitude-based protection cannot find them.

\subsection{Super experts}
\label{app:E8}

We used Algorithm 1 of \citet{su2026super} on 32 × 4,096 tokens of Zyda-2 and of GSM8K. Massive-activation layers were required for a super expert.

\textbf{DeepSeek-V2-Lite has super experts; OLMoE and Moonlight do not.}

\begin{itemize}
\item \textbf{DeepSeek.} We recover the paper's L3E54 (maximum activation 1,432; paper 1,424; 1,304 on GSM8K) and L4E38.
\item \textbf{OLMoE.} The residual stream is \textasciitilde{}100× smaller than DeepSeek's, so the absolute test cannot fire. A scale-free variant flags L1E9, but removing it costs nothing (below).
\item \textbf{Moonlight.} Activations grow smoothly with no jump.
\end{itemize}
\begin{table}[H]
\centering
\caption{Routing away only the listed experts from the dense model (keep-scale), against a same-size random control from the same layers.}
\label{tab:E7}
\footnotesize
\begin{adjustbox}{max width=\linewidth}
\begin{tabular}{@{}lccccc@{}}
\toprule
Removed (control) & Nine-task: control $\rightarrow$ removed & $\Delta$ nine-task & $\Delta$ big-5 & $\Delta$ MMLU & WikiText perplexity \\
\midrule
DeepSeek L3E54 + L4E38 (L3E55 + L4E49) & 66.83 $\rightarrow$ 57.97 & −8.86 [−9.93, −7.82]* & −14.65* & −2.51* & 9.45 $\rightarrow$ 11.36 \\
DeepSeek L3E54 alone (L3E55) & 67.24 $\rightarrow$ 62.07 & −5.17 [−5.95, −4.39]* & −8.25* & −1.71* & 9.45 $\rightarrow$ 9.76 \\
OLMoE L1E9 (L1E55) & 66.88 $\rightarrow$ 66.71 & −0.17 [−0.57, +0.24] & −0.10 & −0.30 & 8.33 $\rightarrow$ 8.36 \\
\bottomrule
\end{tabular}
\end{adjustbox}
\end{table}

\textbf{Two of DeepSeek's 1,664 routed experts are worth more accuracy than MoNE's entire 50\% pruning damage with rerouting (−5.96), and one carries most of it with little perplexity cost.}

\textbf{Frequency's failure on DeepSeek is one expert.}

\begin{itemize}
\item Frequency is the only criterion whose masks prune L3E54.
\item Keeping that expert inside the frequency masks gains +5.46 [+4.48, +6.45]* and +5.28 [+4.16, +6.38]* on the two draws, closing 91\% and 83\% of the deficit to MoNE's mask.
\end{itemize}
\section{Cost and quantization}
\label{app:cost}

\subsection{Memory and latency}
\label{app:F1}

\begin{table}[H]
\centering
\caption{Deployed weights and HuggingFace eager timings relative to the dense model, at 50\% pruning (prompt 512 tokens, 64 decoded; median of 3 repeats). The bracketed figure is the share of MoNE's prefill saving kept. Timings compare operators within a model; they are not serving-stack times.}
\label{tab:F1}
\footnotesize
\begin{adjustbox}{max width=\linewidth}
\begin{tabular}{@{}llccc@{}}
\toprule
Model & Operator & Weights GB (vs dense) & Prefill / dense, batch 1 / 8 / 32 & Decode per token / dense, batch 1 / 8 / 32 \\
\midrule
OLMoE & dense & 13.84 & 1 / 1 / 1 & 1 / 1 / 1 \\
 & constant & 7.40 (−46.5\%) & 0.810 / 0.771 / 0.752 & 0.885 / 0.913 / 0.911 \\
 & tilt r8 & 7.43 (−46.3\%) & 1.007 (−3.8\%) / 0.899 (44.0\%) / 0.832 (67.8\%) & 1.060 / 1.104 / 1.036 \\
 & tilt r32 & 7.53 (−45.6\%) & 1.024 (−12.8\%) / 0.909 (39.6\%) / 0.837 (65.7\%) & 1.076 / 1.132 / 1.070 \\
 & keep reroute & 7.40 (−46.5\%) & 0.825 (92.1\%) / 0.925 (32.8\%) / 0.982 (7.3\%) & 0.946 / 0.967 / 0.916 \\
\midrule
DeepSeek & dense & 32.55 & 1 / 1 / 1 & 1 / 1 / 1 \\
 & constant & 18.15 (−44.2\%) & 0.711 / 0.803 / 0.838 & 0.969 / 0.987 / 0.770 \\
 & tilt r8 & 18.21 (−44.0\%) & 0.996 (1.4\%) / 0.880 (60.7\%) / 0.876 (76.4\%) & 1.062 / 1.321 / 1.103 \\
 & tilt r32 & 18.37 (−43.5\%) & 1.004 (−1.2\%) / 0.880 (61.1\%) / 0.880 (73.9\%) & 1.062 / 1.232 / 1.091 \\
 & keep reroute & 18.16 (−44.2\%) & 0.754 (85.2\%) / 0.942 (29.7\%) / 0.992 (4.6\%) & 1.093 / 1.162 / 0.931 \\
\bottomrule
\end{tabular}
\end{adjustbox}
\end{table}

\textbf{Which operator keeps MoNE's speed-up depends on batch size.}

\begin{itemize}
\item In compute-bound prefill (batch 32) the tilt keeps two thirds to three quarters of the saving, and rerouting almost none.
\item At batch 1 the order reverses, because even prefill is launch-bound there.
\end{itemize}
\textbf{Decode is launch-bound in our unfused implementation.} Rank 32 costs the same per token as rank 8 (ratio 0.93–1.03), despite four times the arithmetic.

\subsection{Quantization setup}
\label{app:F2}

\textbf{GPTQ setup.}

\begin{itemize}
\item \textbf{Pass.} GPTQ is sequential: inputs are propagated through layers already quantized.
\item \textbf{Format.} Asymmetric integer weights, per-row groups of 128 with fp16 scale and zero; damping 1\% of the mean diagonal; no act-order.
\item \textbf{Fallback.} Round-to-nearest for any linear that sees fewer than 128 calibration tokens.
\item \textbf{What is quantized.} Every decoder linear: attention, routed and shared experts, dense MLPs. Embeddings, lm\_head, norms, routers and novices stay at 16 bit and are counted in the size.
\item \textbf{Hessians.} Each Hessian comes from the tokens that reach that linear under the deployed operator.
\item \textbf{Calibration.} 64 × 4,096 tokens of Zyda-2 slice A.
\item \textbf{Simulation.} Quantization is simulated: accuracy is measured, not speed.
\end{itemize}
\textbf{Checks.}

\begin{itemize}
\item A synthetic self-test halves relative output error against round-to-nearest (w4 0.1065 $\rightarrow$ 0.0544).
\item 60 of 60 runs completed with zero tokens reaching a pruned expert.
\end{itemize}
\subsection{Accuracy against deployed size}
\label{app:F3}

\begin{table}[H]
\centering
\caption{Nine-task accuracy (deployed GB).}
\label{tab:F2}
\footnotesize
\begin{adjustbox}{max width=\linewidth}
\begin{tabular}{@{}lccc@{}}
\toprule
Configuration & OLMoE & DeepSeek & Moonlight \\
\midrule
dense bf16 & 66.76 (13.84) & 67.15 (31.41) & 71.48 (31.92) \\
dense GPTQ w4 & 65.18 (3.98) & 66.68 (8.97) & 70.68 (9.47) \\
dense GPTQ w3 & 62.81 (3.14) & 63.12 (7.05) & 65.02 (7.56) \\
dense GPTQ w2 & 45.71 (2.30) & 41.10 (5.14) & 39.19 (5.65) \\
dense RTN w4 / w3 & 64.79 / 60.55 & 65.69 / 60.46 & 69.97 / 61.26 \\
25\% const / tilt / reroute, w4 & 60.17 / 61.99 / 60.66 (3.13–3.19) & 64.15 / 64.36 / 64.31 (7.05–7.16) & 68.15 / 68.55 / 67.95 (7.56–7.67) \\
25\% const / tilt / reroute, w3 & 58.79 / 59.02 / 57.29 (2.49–2.56) & 60.09 / 60.93 / 61.04 (5.59–5.70) & 62.98 / 63.18 / 62.58 (6.10–6.21) \\
50\% const / tilt / reroute, w4 & 48.68 / 52.44 / 50.13 (2.27–2.41) & 59.08 / 59.89 / 60.41 (5.14–5.36) & 59.79 / 61.23 / 59.47 (5.65–5.87) \\
50\% const / tilt / reroute, w3 & 46.35 / 50.46 / 48.07 (1.83–1.97) & 56.98 / 56.72 / 56.93 (4.13–4.35) & 57.48 / 58.21 / 54.60 (4.64–4.86) \\
\bottomrule
\end{tabular}
\end{adjustbox}
\end{table}

\textbf{At the smallest budget, pruning with 4-bit weights is far above uniform 2-bit GPTQ.} Dense GPTQ w2 scores 45.71, 41.10 and 39.19 (guessing is about 40.5), while every 50\%-pruned w4 arm at the same or at most 5\% larger size scores 48.68--61.23 (point estimates). Part of this margin is attention precision, which w2 quantizes to 2 bits; we did not run a 2-bit-expert arm with 4-bit attention.

\subsection{Do the operators' gains survive quantization?}
\label{app:F4}

\begin{table}[H]
\centering
\caption{Change in each operator's gain over the constant caused by quantization, $[\mathrm{op}_w-\mathrm{const}_w]-[\mathrm{op}_{\mathrm{bf16}}-\mathrm{const}_{\mathrm{bf16}}]$, nine-task, jointly bootstrapped over the items common to the four runs. Reroute: rerouting at the trained gate scale.}
\label{tab:F3}
\footnotesize
\begin{adjustbox}{max width=\linewidth}
\begin{tabular}{@{}lcccc@{}}
\toprule
Setting & Tilt r32, w4 & Tilt r32, w3 & Reroute, w4 & Reroute, w3 \\
\midrule
OLMoE 50\% & +0.30 [−0.57, +1.16] & +0.66 [−0.30, +1.62] & −0.94 [−1.85, −0.03]* & −0.66 [−1.90, +0.59] \\
OLMoE 25\% & −0.02 [−0.88, +0.86] & −1.60 [−2.75, −0.49]* & −0.98 [−1.94, −0.03]* & −2.96 [−4.13, −1.83]* \\
DeepSeek 50\% & +0.33 [−0.45, +1.09] & −0.75 [−1.71, +0.22] & −0.02 [−0.99, +0.92] & −1.40 [−2.63, −0.17]* \\
DeepSeek 25\% & +0.05 [−0.71, +0.79] & +0.68 [−0.45, +1.82] & −0.10 [−0.94, +0.71] & +0.69 [−0.62, +2.00] \\
Moonlight 50\% & +0.07 [−0.78, +0.94] & −0.63 [−1.98, +0.69] & −0.14 [−1.14, +0.82] & −2.69 [−4.11, −1.25]* \\
Moonlight 25\% & +0.28 [−0.43, +0.97] & +0.07 [−0.89, +1.02] & −0.25 [−0.92, +0.44] & −0.44 [−1.46, +0.58] \\
\bottomrule
\end{tabular}
\end{adjustbox}
\end{table}

\textbf{The tilt's gain survives 4- and 3-bit quantization; rerouting's erodes.}

\begin{itemize}
\item \textbf{Tilt.} 1 of 12 interactions is significant; the model means are −0.16 / +0.08 / −0.05. The one significant case (OLMoE 25\% w3) comes from COPA, OBQA and RTE; its big-5 interaction is +0.07.
\item \textbf{Rerouting.} 11 of 12 are negative and 5 are significant.
\item \textbf{A candidate mechanism, untested.} Rerouting sends every slot through quantized experts, while the constant's and the tilt's slots stay at 16 bit.
\end{itemize}
\subsection{Pruning against quantizing at matched size}
\label{app:F5}

Quantization is the main competitor at a fixed memory budget \citep[App.~E]{lasby2026reap}. We compared 25\% pruning + w4 against uniform 3-bit experts with every other linear at 4 bit (``e3a4'') at about the same size. We also compared it against a Mixture-Compressor-style allocation of 2/3/4-bit experts (mean 3 bits; MC defaults $\alpha$ = 1, $\beta$ = 1.5).

\begin{table}[H]
\centering
\caption{Paired differences against e3a4 (nine-task [95\% CI]; big-5; MMLU), Zyda-2 calibration.}
\label{tab:F4}
\footnotesize
\begin{adjustbox}{max width=\linewidth}
\begin{tabular}{@{}lccc@{}}
\toprule
Comparison & OLMoE & DeepSeek & Moonlight \\
\midrule
e3a4 − dense w3 (value of 4-bit attention and shared experts) & +0.50 [−0.46, +1.45]; +0.03; +0.94* & +2.02 [+0.97, +3.08]*; +2.85*; +4.08* & +5.26 [+4.14, +6.41]*; +4.22*; +8.94* \\
MC-style − e3a4 & −3.11 [−4.21, −2.01]*; −3.46*; −3.30* & −0.26 [−1.23, +0.69]; −0.64; −0.09 & −2.02 [−2.99, −1.07]*; −1.98*; −6.00* \\
25\% tilt w4 − e3a4 (size × 1.000–1.006) & −1.33 [−2.42, −0.19]*; −2.34*; −10.87* & −0.78 [−1.70, +0.15]; −1.43*; −2.09* & −1.72 [−2.54, −0.91]*; −0.04; −7.18* \\
25\% reroute w4 − e3a4 & −2.65 [−3.81, −1.50]*; −4.35*; −10.05* & −0.83 [−1.79, +0.13]; −1.25*; −2.29* & −2.32 [−3.18, −1.48]*; −0.62; −9.24* \\
25\% const w4 − e3a4 & −3.14 [−4.31, −1.96]*; −4.01*; −10.60* & −0.98 [−2.00, +0.06]; −1.72*; −2.63* & −2.13 [−2.96, −1.31]*; −0.51; −10.10* \\
\bottomrule
\end{tabular}
\end{adjustbox}
\end{table}

\textbf{At this budget and with general calibration, quantizing every expert scores above removing experts on general and knowledge tasks.}

\begin{itemize}
\item Every pruned arm scores below uniform bits on every model: significantly on OLMoE and Moonlight, and on DeepSeek significantly on big-5 and MMLU. The loss is mostly MMLU (−2 to −11).
\item Keeping attention and shared experts at 4 bits is worth up to +5.3 points on its own (first row), so the uniform baseline is e3a4 rather than a fully 3-bit model.
\item The MC-style allocation gives 2 bits to 79–100\% of the experts MoNE prunes and 4 bits to none, so the two importance rankings agree on the least important experts.
\end{itemize}
\begin{table}[H]
\centering
\caption{GSM8K (strict) at about 7 GB with everything calibrated on GSM8K-train: masks, fits and GPTQ statistics.}
\label{tab:F5}
\footnotesize
\begin{adjustbox}{max width=\linewidth}
\begin{tabular}{@{}lcc@{}}
\toprule
Configuration & DeepSeek (GB) & Moonlight (GB) \\
\midrule
dense bf16 & 37.53 (31.41) & 73.84 (31.92) \\
dense w3, Zyda-2 GPTQ & 20.85 (7.05) & 34.72 (7.56) \\
dense w3, GSM8K GPTQ & 29.49 (7.05) & 57.39 (7.56) \\
e3a4, GSM8K GPTQ & 32.90 (7.17) & 69.75 (7.67) \\
MC-style, GSM8K & 35.63 (7.17) & 69.07 (7.67) \\
25\% const / tilt / reroute, w4, GSM8K & 34.42 / 37.15 / 35.18 (7.05–7.16) & 70.81 / 70.05 / 72.33 (7.56–7.67) \\
\bottomrule
\end{tabular}
\end{adjustbox}
\end{table}

\textbf{For in-domain generation, no task-calibrated pruned arm is behind uniform bits, and the tilt on DeepSeek and trained-scale rerouting on Moonlight are significantly ahead.}

\begin{itemize}
\item The tilt beats e3a4 by +4.25* on DeepSeek, and trained-scale rerouting by +2.58* on Moonlight. On DeepSeek the tilt is also above the MC-style allocation (37.15 against 35.63, point estimates).
\item At 23–24\% of the 16-bit size, these arms keep 98–99\% of dense GSM8K.
\item The quantizer's own calibration domain is the largest single effect: +8.6* and +22.7* for dense w3.
\end{itemize}
\textbf{Caveats.}

\begin{itemize}
\item \textbf{One draw.} There is one calibration draw and one MC hyperparameter setting.
\item \textbf{Moonlight fallbacks.} On Moonlight, 99–132 rarely routed GSM8K-calibrated linears fell back to round-to-nearest in the dense competitors.
\item \textbf{Not size-exact.} The pruned arms are within 1.6\% of the size of the arms they are compared with, not exactly size-matched.
\end{itemize}
\section{Reproduction}
\label{app:repro}

\subsection{Environment and data}
\label{app:G1}

\textbf{Environment.}

\begin{itemize}
\item \textbf{Hardware.} Four NVIDIA RTX A6000 (48 GB).
\item \textbf{Software.} torch 2.5.1 + CUDA 12.4, transformers 4.48.2, datasets 3.3.2, flash-attn 2.7.1.post4. lm-evaluation-harness is pinned at commit f724be6.
\item \textbf{Offline evaluation.} All evaluations ran with the Hugging Face hub offline.
\end{itemize}
\textbf{Calibration text.}

\begin{itemize}
\item \textbf{Source.} Zyda-2 (sample-100BT), streamed from the Hub using the complete sorted file listing, cut into 4,096-token blocks, and cached once per tokenizer. Every pruning, fitting, scoring and quantization run reads the same cached tokens.
\item \textbf{Slices.} A, B and C are blocks 0–999, 1,000–1,999 and 2,000–2,999 of one stream. The first 2,000 blocks of the 3,000-block build were verified identical to the 2,000-block caches on all three models.
\item {\sloppy\textbf{OLMoE fingerprints.} 1b03d571e9ecc036 (100 blocks), 55735574fca7c8c5 (500), ca1468cc319e46a2 (1,000), b5829c9a3f16a4f7 (2,000).\par}
\item \textbf{GSM8K-train caches.} ``Question: \dots{} Answer: \dots{}'' format, EOS-separated:
\begin{itemize}
\item OLMoE: 286 blocks, 1,174,978 tokens, cfe5c294832f2d99;
\item DeepSeek: 346 blocks, 1,417,614 tokens, 9909d868d9cfa8ea;
\item Moonlight: 295 blocks, 1,209,959 tokens, 6c1ce42e56dcf9b1.
\end{itemize}
\item \textbf{C4 caches for Appendix D.} DeepSeek f569217d49f4dfc7; Moonlight 47280beabdbcdf94 (1,000 blocks).
\end{itemize}
\subsection{MoNE reproduced}
\label{app:G2}

\begin{table}[H]
\centering
\caption{OLMoE-1B-7B-0125, nine-task accuracy, ours (from MoNE's released code) against the paper.}
\label{tab:G1}
\footnotesize
\begin{adjustbox}{max width=\linewidth}
\begin{tabular}{@{}lccccl@{}}
\toprule
Calibration sequences & Pruned & Ours & Paper & $\Delta$ & Largest single-task gap \\
\midrule
100 & 50\% & 50.28 & 50.52 & −0.23 & RTE −1.81 (5 of 277 items) \\
100 & 25\% & 61.20 & 61.04 & +0.16 & COPA +1.00 \\
500 & 50\% & 49.11 & 48.80 & +0.30 & COPA +1.00 \\
500 & 25\% & 61.70 & 61.48 & +0.21 & COPA +2.00 \\
1,000 & 50\% & 49.16 & 49.48 & −0.32 & OpenBookQA −1.00 \\
1,000 & 25\% & 61.32 & 61.44 & −0.11 & RTE −0.72 \\
\bottomrule
\end{tabular}
\end{adjustbox}
\end{table}

\begin{table}[H]
\centering
\caption{DeepSeek-V2-Lite and Moonlight-16B-A3B per task, 1,000 sequences, ours / paper.}
\label{tab:G2}
\footnotesize
\begin{adjustbox}{max width=\linewidth}
\begin{tabular}{@{}lcccc@{}}
\toprule
Task & DeepSeek 50\% & DeepSeek 25\% & Moonlight 50\% & Moonlight 25\% \\
\midrule
ARC-c & 36.95 / 37.71 & 45.22 / 44.71 & 39.42 / 39.42 & 55.12 / 55.55 \\
ARC-e & 65.57 / 65.36 & 73.70 / 73.99 & 65.82 / 65.82 & 78.91 / 79.29 \\
BoolQ & 73.33 / 73.49 & 78.13 / 78.35 & 76.48 / 76.15 & 79.57 / 79.66 \\
COPA & 83.00 / 84.00 & 89.00 / 88.00 & 88.00 / 88.00 & 90.00 / 90.00 \\
MMLU & 41.12 / 41.22 & 49.10 / 49.19 & 23.85 / 23.81 & 54.78 / 54.79 \\
OpenBookQA & 36.60 / 36.60 & 42.40 / 42.20 & 40.40 / 40.40 & 45.40 / 45.60 \\
PIQA & 75.30 / 75.30 & 78.94 / 79.27 & 78.45 / 78.40 & 81.01 / 80.69 \\
RTE & 56.68 / 57.40 & 60.65 / 59.93 & 58.12 / 58.12 & 59.57 / 59.21 \\
WinoGrande & 69.38 / 69.53 & 71.67 / 70.88 & 70.40 / 70.88 & 72.38 / 72.14 \\
Average & 59.77 / 60.07 & 65.42 / 65.17 & 60.11 / 60.11 & 68.53 / 68.55 \\
\bottomrule
\end{tabular}
\end{adjustbox}
\end{table}

\textbf{MoNE reproduces to within 0.32 points on every average.}

\begin{itemize}
\item \textbf{Mean absolute differences.} OLMoE 0.22 over six settings, with deltas summing to +0.01. DeepSeek 0.30 and 0.25. Moonlight 0.00 and 0.02.
\item \textbf{External check.} Our DeepSeek 25\% mask at 100 sequences equals the authors' released checkpoint in all 416 pruned experts.
\item \textbf{Dense baselines (in house).} 66.76 / 67.15 / 71.48 against the paper's 66.63 / 67.16 / 71.56.
\end{itemize}
\textbf{Details of MoNE's code that matter for reproduction}:

\begin{itemize}
\item \textbf{Frequency term.} The score's frequency term is the mean routing weight over \emph{all} tokens, not over routed tokens as printed. It therefore carries access frequency.
\item \textbf{Running variance.} The running variance skips the first batch.
\item \textbf{Model version.} OLMoE-1B-7B-0125 is used throughout.
\item \textbf{Model path.} Evaluation needs the architecture name in the model path. Otherwise pruned experts are silently re-initialized.
\end{itemize}
\subsection{Integrity checks}
\label{app:G3}

\begin{itemize}
\item \textbf{Constants.} The in-process ``constant'' arm rebuilds MoNE's constants from our statistics on the same mask. It matched the saved checkpoints to within 0.07 (OLMoE), 0.07 (DeepSeek) and 0.17 (OLMoE, 100 sequences).
\item \textbf{Routing.}
\begin{itemize}
\item Collectors assert token alignment bit for bit.
\item Scorers verify slots per token = top-k exactly (8.000 / 6.000).
\item DeepSeek and Moonlight gate replays are verified bit-exact against the real gates with an empty mask.
\item {\textbf{Rerouting semantics against the equations of \cref{sec:routing}.} On Granite, with 0 and 20 experts pruned in layers 0, 16 and 31, both rerouting arms match references written from Granite's router definition exactly (\cref{app:granite}). On Moonlight, the keep arm (unpruned normalizer) is checked before every evaluation on the gates of the first, middle and last pruned layer: with an empty mask it equals the original replay bit for bit; with the real mask it selects exactly the trained arm's experts, every expert kept from the unpruned top-$k$ keeps its unpruned gate, the per-token ratio to the trained arm is one constant over the token's $k$ experts, and the weights match an independent reference written from the model's gate code to a relative error of at most $2.6\times10^{-7}$. On random inputs the ratio exceeds 1 on 8.8--19.7\% of tokens, which the bias-steered choice allows (\cref{sec:routing}); over the full evaluations it does on 9.9\%, 5.2\% and 0.8\% of token--gate calls at 25\%, 50\% and 75\%, and no token reached a pruned expert.}
\item Every rerouting evaluation counts tokens reaching a pruned expert; the count was zero in every run.
\end{itemize}
\item \textbf{Regression.} Before any number from the evaluation pipeline was used, it reproduced four stored results exactly (49.09, 52.54, 59.84, 61.19; $\Delta$ 0.00). A further 27 re-run arms matched their stored scores within 0.10.
\item \textbf{Grid, rank and operator runs.} The tilt fits for the grid masks reused existing fits for every expert already fitted and refitted only the rest; canary refits of 30 (OLMoE) and 32 (DeepSeek) already-fitted experts reproduced their stored fits exactly. The rank-128 fit pass reproduced the paper's rank-32 fits for all 512 experts exactly. Every new rerouting run routed zero tokens to a pruned expert, and every slot-remapping run sent tokens through its remapped slots (about $1.8\times10^{8}$ token-slots per run).
\item \textbf{Derived numbers.} Every reported number is computed from saved per-item outputs, not transcribed by hand, and every interval in the paper was re-checked against those outputs before release.
\end{itemize}

\subsection{Compute}
\label{app:G4}

All runs used the four A6000s. Representative costs:

\begin{itemize}
\item one rank-32 fit pass over 1,000 sequences: \textasciitilde{}1.7 GB of float64 statistics per layer, sharded by layer across GPUs;
\item one GPTQ quantization: 5–17 min per model;
\item one MoRA bias run: 27 min (1.6 s per step, 34–38 GB);
\end{itemize}

\subsection{Availability}
\label{app:G5}

Code, masks and fitted operators are not yet public. This appendix and Appendices B--H give the settings needed to reproduce each result.

\section{Mask grid, further operators and statistics}
\label{app:stats}

This appendix gives every Holm-corrected comparison, the crossed mask $\times$ operator grid, the parameter-matched contrasts, slot remapping and COMPEL's compensation in detail, the merging baselines, and second-draw masks from MoNE's own code. Intervals use $B=10{,}000$ paired, task-stratified item-bootstrap replicates.

\subsection{Holm-corrected comparisons}
\label{app:holm}

Two-sided bootstrap p-values ($B=10{,}000$, floor $2\times10^{-4}$), Holm-adjusted within each family and over all comparisons. Verdict: \emph{both} = survives both adjustments; \emph{table} = survives within its family only; \emph{lost} = significant per comparison but not after Holm within its family; n.s. = not significant per comparison. Families: \cref{fig:operators}; \cref{tab:renorm}; \cref{tab:gsm8k}; the crossed grid (\cref{tab:grid}); slot remapping (\cref{fig:gatescale}); extra experts (\cref{tab:budget}); the rank sweep and equal-budget copies (Rank); MBPP (\cref{tab:mbpp}); COMPEL; Granite's renorm $-$ keep rows (\cref{tab:granite}). Intervals here are the same as in the main text. These families were fixed on 24 October 2026, after most of the results they contain were known, so Holm is a conservative correction over the contrasts the conclusions rest on rather than a confirmatory test; they cover those comparisons, not every comparison in the paper. Outside them, and exploratory: Granite's contrasts other than renorm $-$ keep, the per-mask grid contrasts (\cref{tab:H3}), the in-domain GSM8K comparisons, the other calibration-domain results, and the follow-up runs of \cref{sec:granite,sec:domain} other than their seven rows in the families.

{\scriptsize
\setlength{\tabcolsep}{3pt}
\begin{longtable}{@{}l>{\raggedright\arraybackslash}p{5.6cm}ccccl@{}}
\caption{Every paired comparison with Holm-adjusted p-values (adjusted values capped at 1). Keep: rerouting at the trained gate scale (renormalized on Moonlight).}\label{tab:H2}\\
\toprule
Family & Comparison & $\Delta$ [95\% CI] & $p$ & $p_{\text{Holm}}$ table & $p_{\text{Holm}}$ all & Verdict \\
\midrule
\endfirsthead
\toprule
Family & Comparison & $\Delta$ [95\% CI] & $p$ & $p_{\text{Holm}}$ table & $p_{\text{Holm}}$ all & Verdict \\
\midrule
\endhead
Fig.~\ref{fig:operators} & OLMoE 25\% tilt $-$ constant & +1.83 [+1.21, +2.47]* & 2.0e$-$4 & 0.005 & 0.021 & both \\
Fig.~\ref{fig:operators} & OLMoE 25\% keep $-$ constant & +1.46 [+0.85, +2.11]* & 2.0e$-$4 & 0.005 & 0.021 & both \\
Fig.~\ref{fig:operators} & OLMoE 25\% tilt $-$ keep & +0.37 [$-$0.28, +1.00] & 0.262 & 1 & 1 & n.s. \\
Fig.~\ref{fig:operators} & OLMoE 50\% tilt $-$ constant & +3.45 [+2.62, +4.27]* & 2.0e$-$4 & 0.005 & 0.021 & both \\
Fig.~\ref{fig:operators} & OLMoE 50\% keep $-$ constant & +2.38 [+1.42, +3.33]* & 2.0e$-$4 & 0.005 & 0.021 & both \\
Fig.~\ref{fig:operators} & OLMoE 50\% tilt $-$ keep & +1.07 [+0.07, +2.08]* & 0.035 & 0.381 & 1 & lost \\
Fig.~\ref{fig:operators} & OLMoE 75\% tilt $-$ constant & +2.80 [+1.81, +3.81]* & 2.0e$-$4 & 0.005 & 0.021 & both \\
Fig.~\ref{fig:operators} & OLMoE 75\% keep $-$ constant & +1.90 [+0.95, +2.86]* & 2.0e$-$4 & 0.005 & 0.021 & both \\
Fig.~\ref{fig:operators} & OLMoE 75\% tilt $-$ keep & +0.90 [$-$0.20, +2.01] & 0.104 & 0.932 & 1 & n.s. \\
Fig.~\ref{fig:operators} & DeepSeek 25\% tilt $-$ constant & +0.16 [$-$0.28, +0.56] & 0.463 & 1 & 1 & n.s. \\
Fig.~\ref{fig:operators} & DeepSeek 25\% keep $-$ constant & +0.26 [$-$0.35, +0.87] & 0.409 & 1 & 1 & n.s. \\
Fig.~\ref{fig:operators} & DeepSeek 25\% tilt $-$ keep & $-$0.10 [$-$0.79, +0.57] & 0.757 & 1 & 1 & n.s. \\
Fig.~\ref{fig:operators} & DeepSeek 50\% tilt $-$ constant & +0.48 [$-$0.03, +0.99] & 0.065 & 0.650 & 1 & n.s. \\
Fig.~\ref{fig:operators} & DeepSeek 50\% keep $-$ constant & +1.35 [+0.65, +2.08]* & 4.0e$-$4 & 0.007 & 0.026 & both \\
Fig.~\ref{fig:operators} & DeepSeek 50\% tilt $-$ keep & $-$0.87 [$-$1.65, $-$0.10]* & 0.026 & 0.317 & 1 & lost \\
Fig.~\ref{fig:operators} & DeepSeek 75\% tilt $-$ constant & +1.23 [+0.32, +2.15]* & 0.009 & 0.112 & 0.430 & lost \\
Fig.~\ref{fig:operators} & DeepSeek 75\% keep $-$ constant & +4.21 [+3.21, +5.24]* & 2.0e$-$4 & 0.005 & 0.021 & both \\
Fig.~\ref{fig:operators} & DeepSeek 75\% tilt $-$ keep & $-$2.98 [$-$4.08, $-$1.91]* & 2.0e$-$4 & 0.005 & 0.021 & both \\
Fig.~\ref{fig:operators} & Moonlight 25\% tilt $-$ constant & +0.13 [$-$0.16, +0.42] & 0.384 & 1 & 1 & n.s. \\
Fig.~\ref{fig:operators} & Moonlight 25\% keep $-$ constant & +0.05 [$-$0.29, +0.39] & 0.764 & 1 & 1 & n.s. \\
Fig.~\ref{fig:operators} & Moonlight 25\% tilt $-$ keep & +0.08 [$-$0.28, +0.43] & 0.662 & 1 & 1 & n.s. \\
Fig.~\ref{fig:operators} & Moonlight 50\% tilt $-$ constant & +1.37 [+0.73, +2.01]* & 2.0e$-$4 & 0.005 & 0.021 & both \\
Fig.~\ref{fig:operators} & Moonlight 50\% keep $-$ constant & $-$0.18 [$-$1.04, +0.68] & 0.671 & 1 & 1 & n.s. \\
Fig.~\ref{fig:operators} & Moonlight 50\% tilt $-$ keep & +1.55 [+0.66, +2.43]* & 0.002 & 0.024 & 0.091 & table \\
Fig.~\ref{fig:operators} & Moonlight 75\% tilt $-$ constant & +3.27 [+2.43, +4.15]* & 2.0e$-$4 & 0.005 & 0.021 & both \\
Fig.~\ref{fig:operators} & Moonlight 75\% keep $-$ constant & +1.59 [+0.62, +2.59]* & 1.0e$-$3 & 0.016 & 0.060 & table \\
Fig.~\ref{fig:operators} & Moonlight 75\% tilt $-$ keep & +1.68 [+0.58, +2.77]* & 0.003 & 0.042 & 0.162 & table \\
\midrule
Tab.~\ref{tab:renorm} & OLMoE 25\% renorm $-$ keep & $-$4.50 [$-$5.48, $-$3.57]* & 2.0e$-$4 & 0.001 & 0.021 & both \\
Tab.~\ref{tab:renorm} & OLMoE 50\% renorm $-$ keep & $-$3.39 [$-$4.66, $-$2.12]* & 2.0e$-$4 & 0.001 & 0.021 & both \\
Tab.~\ref{tab:renorm} & OLMoE 75\% renorm $-$ keep & $-$3.09 [$-$4.52, $-$1.68]* & 2.0e$-$4 & 0.001 & 0.021 & both \\
Tab.~\ref{tab:renorm} & DeepSeek 25\% renorm $-$ keep & $-$0.79 [$-$1.51, $-$0.07]* & 0.033 & 0.033 & 1 & table \\
Tab.~\ref{tab:renorm} & DeepSeek 50\% renorm $-$ keep & $-$4.67 [$-$5.71, $-$3.59]* & 2.0e$-$4 & 0.001 & 0.021 & both \\
Tab.~\ref{tab:renorm} & DeepSeek 75\% renorm $-$ keep & $-$8.47 [$-$9.98, $-$7.01]* & 2.0e$-$4 & 0.001 & 0.021 & both \\
\midrule
Tab.~\ref{tab:gsm8k} & OLMoE 25\% GSM8K tilt32 $-$ constant & $-$0.61 [$-$1.59, +0.38] & 0.233 & 1 & 1 & n.s. \\
Tab.~\ref{tab:gsm8k} & OLMoE 25\% GSM8K keep $-$ constant & +0.30 [$-$0.76, +1.29] & 0.608 & 1 & 1 & n.s. \\
Tab.~\ref{tab:gsm8k} & OLMoE 50\% GSM8K tilt32 $-$ constant & +0.00 [+0.00, +0.00] & 1 & 1 & 1 & n.s. \\
Tab.~\ref{tab:gsm8k} & OLMoE 50\% GSM8K keep $-$ constant & +0.68 [+0.30, +1.14]* & 4.0e$-$4 & 0.004 & 0.026 & both \\
Tab.~\ref{tab:gsm8k} & DeepSeek 25\% GSM8K tilt32 $-$ constant & $-$1.44 [$-$3.11, +0.30] & 0.110 & 0.769 & 1 & n.s. \\
Tab.~\ref{tab:gsm8k} & DeepSeek 25\% GSM8K keep $-$ constant & $-$0.76 [$-$2.58, +1.06] & 0.454 & 1 & 1 & n.s. \\
Tab.~\ref{tab:gsm8k} & DeepSeek 50\% GSM8K tilt32 $-$ constant & +0.83 [$-$0.53, +2.20] & 0.232 & 1 & 1 & n.s. \\
Tab.~\ref{tab:gsm8k} & DeepSeek 50\% GSM8K keep $-$ constant & +2.58 [+1.21, +4.02]* & 2.0e$-$4 & 0.002 & 0.021 & both \\
Tab.~\ref{tab:gsm8k} & Moonlight 25\% GSM8K tilt32 $-$ constant & +0.15 [$-$1.97, +2.35] & 0.923 & 1 & 1 & n.s. \\
Tab.~\ref{tab:gsm8k} & Moonlight 25\% GSM8K keep $-$ constant & +10.16 [+7.66, +12.66]* & 2.0e$-$4 & 0.002 & 0.021 & both \\
Tab.~\ref{tab:gsm8k} & Moonlight 50\% GSM8K tilt32 $-$ constant & +0.99 [+0.30, +1.67]* & 0.004 & 0.034 & 0.223 & table \\
Tab.~\ref{tab:gsm8k} & Moonlight 50\% GSM8K keep $-$ constant & +1.14 [+0.38, +1.90]* & 0.003 & 0.023 & 0.143 & table \\
\midrule
Tab.~\ref{tab:grid} & OLMoE 50\% tilt $-$ constant, mean of 4 masks & +3.93 [+3.36, +4.50]* & 2.0e$-$4 & 0.002 & 0.021 & both \\
Tab.~\ref{tab:grid} & OLMoE 50\% keep $-$ constant, mean of 4 masks & +1.54 [+0.98, +2.09]* & 2.0e$-$4 & 0.002 & 0.021 & both \\
Tab.~\ref{tab:grid} & OLMoE 50\% tilt $-$ keep, mean of 4 masks & +2.39 [+1.75, +3.04]* & 2.0e$-$4 & 0.002 & 0.021 & both \\
Tab.~\ref{tab:grid} & OLMoE 50\% renorm $-$ keep, mean of MoNE + REAP masks & $-$2.64 [$-$3.75, $-$1.57]* & 2.0e$-$4 & 0.002 & 0.021 & both \\
Tab.~\ref{tab:grid} & DeepSeek 50\% tilt $-$ constant, mean of 4 masks & +0.11 [$-$0.24, +0.45] & 0.526 & 0.955 & 1 & n.s. \\
Tab.~\ref{tab:grid} & DeepSeek 50\% keep $-$ constant, mean of 4 masks & +1.33 [+0.92, +1.74]* & 2.0e$-$4 & 0.002 & 0.021 & both \\
Tab.~\ref{tab:grid} & DeepSeek 50\% tilt $-$ keep, mean of 4 masks & $-$1.22 [$-$1.73, $-$0.73]* & 2.0e$-$4 & 0.002 & 0.021 & both \\
Tab.~\ref{tab:grid} & DeepSeek 50\% renorm $-$ keep, mean of MoNE + REAP masks & $-$4.28 [$-$5.16, $-$3.39]* & 2.0e$-$4 & 0.002 & 0.021 & both \\
Tab.~\ref{tab:grid} & OLMoE 50\% SGE $-$ MoNE mask under constant, mean of draws A-C & +0.77 [$-$0.15, +1.71] & 0.102 & 0.307 & 1 & n.s. \\
Tab.~\ref{tab:grid} & OLMoE 50\% SGE $-$ MoNE mask under tilt r32, mean of draws A-C & +0.94 [+0.00, +1.88]* & 0.049 & 0.198 & 1 & lost \\
Tab.~\ref{tab:grid} & OLMoE 50\% SGE $-$ MoNE mask under keep reroute, mean of draws A-C & +0.42 [$-$0.78, +1.65] & 0.478 & 0.955 & 1 & n.s. \\
\midrule
Remap & OLMoE 50\% remap nearest $-$ keep & $-$6.05 [$-$7.32, $-$4.81]* & 2.0e$-$4 & 0.002 & 0.021 & both \\
Remap & OLMoE 50\% remap functional $-$ nearest & +2.04 [+0.78, +3.27]* & 0.002 & 0.006 & 0.091 & table \\
Remap & OLMoE 50\% remap functional $-$ keep & $-$4.01 [$-$5.24, $-$2.81]* & 2.0e$-$4 & 0.002 & 0.021 & both \\
Remap & OLMoE 50\% tilt $-$ remap functional & +5.08 [+3.92, +6.29]* & 2.0e$-$4 & 0.002 & 0.021 & both \\
Remap & DeepSeek 50\% remap nearest $-$ keep & $-$3.04 [$-$4.15, $-$1.94]* & 2.0e$-$4 & 0.002 & 0.021 & both \\
Remap & DeepSeek 50\% remap functional $-$ nearest & $-$0.80 [$-$1.84, +0.23] & 0.135 & 0.268 & 1 & n.s. \\
Remap & DeepSeek 50\% remap functional $-$ keep & $-$3.84 [$-$5.05, $-$2.64]* & 2.0e$-$4 & 0.002 & 0.021 & both \\
Remap & DeepSeek 50\% tilt $-$ remap functional & +2.96 [+1.68, +4.26]* & 2.0e$-$4 & 0.002 & 0.021 & both \\
Remap & Moonlight 50\% remap nearest $-$ keep & $-$1.15 [$-$2.17, $-$0.16]* & 0.023 & 0.068 & 1 & lost \\
Remap & Moonlight 50\% remap functional $-$ nearest & $-$0.70 [$-$1.62, +0.23] & 0.134 & 0.268 & 1 & n.s. \\
Remap & Moonlight 50\% remap functional $-$ keep & $-$1.86 [$-$2.90, $-$0.84]* & 2.0e$-$4 & 0.002 & 0.021 & both \\
Remap & Moonlight 50\% tilt $-$ remap functional & +3.40 [+2.37, +4.44]* & 2.0e$-$4 & 0.002 & 0.021 & both \\
\midrule
Tab.~\ref{tab:budget} & OLMoE 50\% tilt r32 $-$ MoNE keeping 33 & +2.27 [+1.39, +3.18]* & 2.0e$-$4 & 0.003 & 0.021 & both \\
Tab.~\ref{tab:budget} & OLMoE 50\% tilt r32 $-$ MoNE keeping 34 & +1.49 [+0.49, +2.49]* & 0.005 & 0.043 & 0.250 & table \\
Tab.~\ref{tab:budget} & OLMoE 50\% keep reroute $-$ MoNE keeping 33 & +1.21 [+0.29, +2.14]* & 0.010 & 0.074 & 0.470 & lost \\
Tab.~\ref{tab:budget} & OLMoE 50\% MoNE keeping 33 $-$ constant & +1.18 [+0.56, +1.79]* & 4.0e$-$4 & 0.005 & 0.026 & both \\
Tab.~\ref{tab:budget} & DeepSeek 50\% tilt r32 $-$ MoNE keeping 33 & $-$0.37 [$-$1.01, +0.26] & 0.242 & 0.727 & 1 & n.s. \\
Tab.~\ref{tab:budget} & DeepSeek 50\% tilt r32 $-$ MoNE keeping 34 & $-$0.34 [$-$0.98, +0.31] & 0.298 & 0.727 & 1 & n.s. \\
Tab.~\ref{tab:budget} & DeepSeek 50\% keep reroute $-$ MoNE keeping 33 & +0.50 [$-$0.12, +1.13] & 0.110 & 0.548 & 1 & n.s. \\
Tab.~\ref{tab:budget} & DeepSeek 50\% MoNE keeping 33 $-$ constant & +0.86 [+0.37, +1.36]* & 0.001 & 0.014 & 0.083 & table \\
Tab.~\ref{tab:budget} & Moonlight 50\% tilt r32 $-$ MoNE keeping 33 & +0.84 [+0.21, +1.48]* & 0.009 & 0.074 & 0.451 & lost \\
Tab.~\ref{tab:budget} & Moonlight 50\% tilt r32 $-$ MoNE keeping 34 & +0.16 [$-$0.53, +0.85] & 0.651 & 0.727 & 1 & n.s. \\
Tab.~\ref{tab:budget} & Moonlight 50\% keep reroute $-$ MoNE keeping 33 & $-$0.71 [$-$1.65, +0.24] & 0.140 & 0.561 & 1 & n.s. \\
Tab.~\ref{tab:budget} & Moonlight 50\% MoNE keeping 33 $-$ constant & +0.53 [$-$0.06, +1.12] & 0.081 & 0.484 & 1 & n.s. \\
Tab.~\ref{tab:budget} & OLMoE 50\% tilt r8 $-$ MoNE keeping 33 & +1.65 [+0.79, +2.53]* & 4.0e$-$4 & 0.005 & 0.026 & both \\
\midrule
Rank & OLMoE 50\% tilt r64 $-$ r32 & +0.97 [+0.41, +1.56]* & 6.0e$-$4 & 0.002 & 0.037 & both \\
Rank & OLMoE 50\% tilt r128 $-$ r32 & +1.51 [+0.90, +2.17]* & 2.0e$-$4 & 0.001 & 0.021 & both \\
Rank & OLMoE 50\% tilt r32 $-$ top-21-neuron copy (1x) & $-$0.09 [$-$0.95, +0.78] & 0.852 & 1 & 1 & n.s. \\
Rank & OLMoE 50\% tilt r32 $-$ whitened SVD k14 (1x) & +3.78 [+2.96, +4.65]* & 2.0e$-$4 & 0.001 & 0.021 & both \\
Rank & OLMoE 50\% tilt r128 $-$ top-85-neuron copy (4x) & +0.01 [$-$1.02, +1.05] & 0.998 & 1 & 1 & n.s. \\
Rank & OLMoE 50\% tilt r128 $-$ whitened SVD k56 (4x) & +4.30 [+3.43, +5.20]* & 2.0e$-$4 & 0.001 & 0.021 & both \\
\midrule
Tab.~\ref{tab:mbpp} & OLMoE 25\% MBPP tilt $-$ constant & +0.00 [+0.00, +0.00] & 1 & 1 & 1 & n.s. \\
Tab.~\ref{tab:mbpp} & OLMoE 25\% MBPP keep $-$ constant & +0.00 [+0.00, +0.00] & 1 & 1 & 1 & n.s. \\
Tab.~\ref{tab:mbpp} & DeepSeek 25\% MBPP tilt $-$ constant & $-$1.20 [$-$3.20, +0.80] & 0.305 & 1 & 1 & n.s. \\
Tab.~\ref{tab:mbpp} & DeepSeek 25\% MBPP keep $-$ constant & +1.60 [$-$0.20, +3.40] & 0.103 & 0.517 & 1 & n.s. \\
Tab.~\ref{tab:mbpp} & Moonlight 25\% MBPP tilt $-$ constant & +0.00 [+0.00, +0.00] & 1 & 1 & 1 & n.s. \\
Tab.~\ref{tab:mbpp} & Moonlight 25\% MBPP keep $-$ constant & +1.20 [+0.40, +2.20]* & 0.005 & 0.031 & 0.265 & table \\
\midrule
Tab.~\ref{tab:granite} & Granite 25\% renorm $-$ keep & +0.28 [$-$0.54, +1.09] & 0.487 & 0.974 & 1 & n.s. \\
Tab.~\ref{tab:granite} & Granite 50\% renorm $-$ keep & +2.08 [+1.08, +3.11]* & 4.0e$-$4 & 0.001 & 0.026 & both \\
Tab.~\ref{tab:granite} & Granite 75\% renorm $-$ keep & +0.27 [$-$0.81, +1.36] & 0.623 & 0.974 & 1 & n.s. \\
\midrule
COMPEL & OLMoE 50\% COMPEL + keep $-$ keep & $-$3.26 [$-$4.43, $-$2.06]* & 2.0e$-$4 & 0.002 & 0.021 & both \\
COMPEL & OLMoE 50\% COMPEL + keep $-$ constant & $-$0.87 [$-$1.90, +0.15] & 0.094 & 0.350 & 1 & n.s. \\
COMPEL & OLMoE 50\% COMPEL + renorm $-$ renorm & $-$1.86 [$-$2.97, $-$0.75]* & 0.001 & 0.008 & 0.083 & table \\
COMPEL & DeepSeek 50\% COMPEL + keep $-$ keep & +0.39 [$-$0.17, +0.96] & 0.181 & 0.362 & 1 & n.s. \\
COMPEL & DeepSeek 50\% COMPEL + keep $-$ constant & +1.74 [+0.99, +2.53]* & 2.0e$-$4 & 0.002 & 0.021 & both \\
COMPEL & DeepSeek 50\% COMPEL + renorm $-$ renorm & +0.07 [$-$0.72, +0.81] & 0.869 & 0.869 & 1 & n.s. \\
COMPEL & Moonlight 50\% COMPEL + keep $-$ keep & $-$0.61 [$-$1.30, +0.08] & 0.087 & 0.350 & 1 & n.s. \\
COMPEL & Moonlight 50\% COMPEL + keep $-$ constant & $-$0.79 [$-$1.65, +0.05] & 0.066 & 0.331 & 1 & n.s. \\
\bottomrule
\end{longtable}
}

The smallest attainable raw $p$-value is $2\times10^{-4}$, so the smallest attainable adjusted value is about 0.021 over 104 comparisons and 0.022 over 111; entries at that value are at the resolution limit of $B$, and the strongest contrasts are not ranked against each other. Counts over the 104 comparisons: 62 significant per comparison, 55 after Holm within their family, 44 after Holm over all. The seven rows of the follow-up runs (Moonlight renorm $-$ keep at 25, 50 and 75\%, \cref{tab:moonkeep}; the four fit effects of \cref{tab:domainx}) form two further families; with them the overall family has 111 comparisons, 47 significant after Holm over all, and no verdict in this table changes. Seven comparisons are significant per comparison but not after Holm within their family: OLMoE and DeepSeek 50\% tilt $-$ keep on MoNE's mask (both survive when averaged over masks), DeepSeek 75\% tilt $-$ constant, OLMoE SGE $-$ MoNE under the tilt averaged over draws, Moonlight remap nearest $-$ keep, OLMoE keep $-$ keeping 33, and Moonlight tilt $-$ keeping 33. One verdict is borderline: Moonlight 75\% keep $-$ constant survives Holm within its table but narrowly misses over all comparisons (adjusted $p=0.060$). MBPP's only significant contrast, Moonlight's keep $-$ constant (+1.20), survives Holm within its family only. Granite's 50\% renorm $-$ keep survives both adjustments (0.001 and 0.026); its 25\% and 75\% rows are not significant. Adding the three Granite rows changed no verdict among the other 101.

\subsection{Equivalence bounds for the comparisons the text calls close}
\label{app:tost}
{For every pair of operators the text calls close, the table gives the paired difference with its 95\% interval and the 90\% interval. A margin $\delta$ passes two one-sided tests at $\alpha=0.05$ if the 90\% interval lies inside $(-\delta,\delta)$; the last column is the smallest such $\delta$. No margin was set before the data were seen, so these are bounds on what the data rule out, not tests of a prespecified equivalence hypothesis. Nine-task, paired item bootstrap, $B=10{,}000$.

\begin{table}[H]
\centering
\caption{Equivalence bounds (points, nine-task).}
\label{tab:tost}
\footnotesize
\begin{adjustbox}{max width=\linewidth}
\begin{tabular}{@{}lccc@{}}
\toprule
Comparison ($b-a$) & Difference [95\%] & 90\% interval & Differences ruled out beyond \\
\midrule
OLMoE 50\%, tilt r32 $-$ top-neuron copy 1$\times$ & $-$0.09 [$-$0.95, +0.78] & [$-$0.81, +0.64] & 0.81 \\
DeepSeek 50\%, tilt r32 $-$ top-neuron copy 1$\times$ & +0.14 [$-$0.56, +0.83] & [$-$0.46, +0.73] & 0.73 \\
Moonlight 50\%, tilt r32 $-$ top-neuron copy 1$\times$ & +0.66 [+0.09, +1.27]* & [+0.17, +1.17] & not close: tilt ahead \\
OLMoE 50\%, tilt r128 $-$ top-neuron copy 4$\times$ & +0.01 [$-$1.02, +1.05] & [$-$0.85, +0.89] & 0.89 \\
DeepSeek 50\%, tilt r32 $-$ MoNE keeping 33 & $-$0.37 [$-$1.01, +0.26] & [$-$0.90, +0.16] & 0.90 \\
Moonlight 50\%, tilt r32 $-$ MoNE keeping 33 & +0.84 [+0.21, +1.48]* & [+0.30, +1.37] & not close: tilt ahead per comparison \\
DeepSeek 50\%, keep reroute $-$ MoNE keeping 33 & +0.50 [$-$0.12, +1.13] & [$-$0.01, +1.03] & 1.03 \\
DeepSeek 50\%, MoNE keeping 34 $-$ keeping 33 & $-$0.04 [$-$0.43, +0.36] & [$-$0.37, +0.29] & 0.37 \\
Moonlight 25\%, tilt r32 $-$ trained-scale reroute & +0.08 [$-$0.28, +0.43] & [$-$0.22, +0.37] & 0.37 \\
DeepSeek 50\%, COMPEL + keep $-$ keep & +0.39 [$-$0.17, +0.96] & [$-$0.09, +0.86] & 0.86 \\
Moonlight 50\%, COMPEL + keep $-$ keep & $-$0.61 [$-$1.30, +0.08] & [$-$1.19, $-$0.02] & 1.19 \\
Granite 25\%, renorm $-$ keep & +0.28 [$-$0.54, +1.09] & [$-$0.41, +0.96] & 0.96 \\
Granite 75\%, renorm $-$ keep & +0.27 [$-$0.81, +1.36] & [$-$0.64, +1.19] & 1.19 \\
\bottomrule
\end{tabular}
\end{adjustbox}
\end{table}}

\subsection{The crossed grid in full}
\label{app:gridfull}

\begin{table}[H]
\centering
\caption{Operator effects on each 50\% mask, nine-task, draw A, 95\% intervals. Holm within these 28 per-mask contrasts (an exploratory family): 16 survive; tilt $-$ keep on OLMoE's MoNE and SGE masks, keep $-$ constant and renorm $-$ keep on OLMoE's REAP mask, tilt $-$ keep on DeepSeek's MoNE mask and keep $-$ constant and tilt $-$ keep on its SGE mask are significant per comparison only.}
\label{tab:H3}
\footnotesize
\begin{adjustbox}{max width=\linewidth}
\begin{tabular}{@{}llcccc@{}}
\toprule
Model & Mask & Tilt $-$ constant & Keep $-$ constant & Tilt $-$ keep & Renorm $-$ keep \\
\midrule
OLMoE & MoNE & +3.45 [+2.62, +4.27]* & +2.38 [+1.42, +3.33]* & +1.07 [+0.07, +2.08]* & $-$3.39 [$-$4.66, $-$2.12]* \\
 & SGE & +4.52 [+3.64, +5.43]* & +3.27 [+2.15, +4.36]* & +1.26 [+0.14, +2.36]* & --- \\
 & REAP & +3.89 [+2.82, +4.97]* & +1.11 [+0.10, +2.12]* & +2.78 [+1.58, +3.98]* & $-$1.88 [$-$3.44, $-$0.37]* \\
 & Frequency & +3.86 [+2.71, +5.04]* & $-$0.61 [$-$1.40, +0.19] & +4.47 [+3.24, +5.71]* & --- \\
\midrule
DeepSeek & MoNE & +0.48 [$-$0.03, +0.99] & +1.35 [+0.65, +2.08]* & $-$0.87 [$-$1.65, $-$0.10]* & $-$4.67 [$-$5.71, $-$3.59]* \\
 & SGE & $-$0.06 [$-$0.78, +0.63] & +1.07 [+0.21, +1.91]* & $-$1.12 [$-$2.09, $-$0.19]* & --- \\
 & REAP & +0.03 [$-$0.69, +0.74] & +1.76 [+1.06, +2.44]* & $-$1.72 [$-$2.58, $-$0.87]* & $-$3.88 [$-$5.03, $-$2.71]* \\
 & Frequency (SE kept) & $-$0.00 [$-$0.72, +0.66] & +1.15 [+0.48, +1.86]* & $-$1.15 [$-$1.97, $-$0.38]* & --- \\
\bottomrule
\end{tabular}
\end{adjustbox}
\end{table}

\begin{table}[H]
\centering
\caption{Selection effects: each criterion's 50\% mask minus MoNE's, under each operator, nine-task, draw A.}
\label{tab:H4}
\footnotesize
\begin{adjustbox}{max width=\linewidth}
\begin{tabular}{@{}llcccc@{}}
\toprule
Model & Mask $-$ MoNE & Constant & Tilt r32 & Keep & Renorm \\
\midrule
OLMoE & SGE & +0.72 [$-$0.39, +1.80] & +1.79 [+0.73, +2.86]* & +1.60 [+0.33, +2.86]* & --- \\
 & REAP & $-$1.21 [$-$2.41, $-$0.01]* & $-$0.77 [$-$2.07, +0.51] & $-$2.48 [$-$3.89, $-$1.10]* & $-$0.97 [$-$2.35, +0.36] \\
 & Frequency & $-$4.60 [$-$5.80, $-$3.46]* & $-$4.19 [$-$5.39, $-$3.02]* & $-$7.60 [$-$8.92, $-$6.32]* & --- \\
\midrule
DeepSeek & SGE & +0.84 [+0.05, +1.65]* & +0.30 [$-$0.50, +1.11] & +0.55 [$-$0.33, +1.44] & --- \\
 & REAP & $-$1.14 [$-$2.02, $-$0.24]* & $-$1.59 [$-$2.28, $-$0.87]* & $-$0.74 [$-$1.76, +0.23] & +0.06 [$-$1.05, +1.16] \\
 & Frequency (SE kept) & $-$0.35 [$-$1.23, +0.53] & $-$0.83 [$-$1.72, +0.05] & $-$0.55 [$-$1.42, +0.36] & --- \\
\bottomrule
\end{tabular}
\end{adjustbox}
\end{table}

\begin{table}[H]
\centering
\caption{OLMoE 50\%, MoNE against SGE masks on three calibration draws (mask and fits from that draw; MoNE's mask is always from slice A, so its keep-scale run is the same on every draw).}
\label{tab:H5}
\footnotesize
\begin{adjustbox}{max width=\linewidth}
\begin{tabular}{@{}lcccccc@{}}
\toprule
Draw & MoNE: const / tilt / keep & SGE: const / tilt / keep & SGE: tilt $-$ const & SGE: keep $-$ const & SGE $-$ MoNE, const & SGE $-$ MoNE, keep \\
\midrule
A & 49.09 / 52.54 / 51.48 & 49.81 / 54.34 / 53.08 & +4.52* & +3.27* & +0.72 & +1.60* \\
B & 49.18 / 52.82 / 51.48 & 49.99 / 53.33 / 51.13 & +3.34* & +1.14 & +0.82 & $-$0.34 \\
C & 49.26 / 52.70 / 51.48 & 50.04 / 53.21 / 51.49 & +3.17* & +1.45* & +0.78 & +0.01 \\
Mean of draws & & & & & +0.77 [$-$0.15, +1.71] & +0.42 [$-$0.78, +1.65] \\
\bottomrule
\end{tabular}
\end{adjustbox}
\end{table}

\textbf{Operator order on each mask.} OLMoE: tilt $>$ keep $>$ constant ($>$ renorm) on the MoNE, SGE and REAP masks; tilt $>$ constant $>$ keep on the frequency mask. DeepSeek: keep $>$ tilt $>$ constant ($>$ renorm) on the MoNE and REAP masks; keep $>$ constant $\geq$ tilt on the SGE and frequency masks. The highest-scoring operator never changes; the order below it does.

\textbf{The operator effect does not depend on MoNE's mask; SGE's selection advantage does depend on the draw.} The tilt's gain over the constant on the SGE mask stays between +3.17 and +4.52 on all three draws. SGE's advantage over MoNE's mask is significant only on draw A (\cref{tab:E3}); averaged over draws it is +0.77, +0.94 [+0.00, +1.88] and +0.42 under the constant, the tilt and keep, and none survives Holm correction.

\subsection{Parameter-matched contrasts}
\label{app:budget}

\begin{table}[H]

\caption{Nine-task accuracy at 50\% pruning against the parameters each replacement adds to MoNE's constant model (MoNE masks, draw A). Extra experts: MoNE's mask keeping 33 or 34 of 64 experts per layer, with constants for the rest; one extra expert per layer is 100.7\,M on OLMoE and 224.9\,M on the other two. Copies store a compressed version of each pruned expert at the tilt's rank-32 budget (1×) or four times it (4×, OLMoE only); their bias replaces the constant's stored mean, so they add $d$ values per pruned expert less than they store (\cref{tab:C2} lists what they store). Every arm is evaluated in the same process with the same constants; MoNE's saved keep-33/34 checkpoints, evaluated separately, give 49.95 and 51.22 on OLMoE (\cref{app:evalcheck}). \holmt/\holma: survives Holm within the table / over all comparisons.}
\label{tab:budget}
\centering\footnotesize
\setlength{\tabcolsep}{4pt}
\begin{tabular}{@{}lcccccc@{}}
\toprule
 & \multicolumn{2}{c}{OLMoE} & \multicolumn{2}{c}{DeepSeek} & \multicolumn{2}{c}{Moonlight} \\
\cmidrule(lr){2-3}\cmidrule(lr){4-5}\cmidrule(l){6-7}
Arm & Added & Acc. & Added & Acc. & Added & Acc. \\
\midrule
Constant (MoNE) & 0 & 49.09 & 0 & 59.84 & 0 & 60.24 \\
Rerouting, trained scale & 0 & 51.48 & 0 & 61.19 & 0 & 60.06 \\
Tilt r8 & 16.8\,M & 51.92 & 27.3\,M & 59.42 & 27.3\,M & 60.87 \\
Tilt r32 & 67.1\,M & 52.54 & 109.1\,M & 60.32 & 109.1\,M & 61.61 \\
Tilt r64 & 134.3\,M & 53.52 & --- & --- & --- & --- \\
Tilt r128 & 268.6\,M & 54.05 & --- & --- & --- & --- \\
MoNE keeping 33 & 100.7\,M & 50.27 & 224.9\,M & 60.69 & 224.9\,M & 60.77 \\
MoNE keeping 34 & 201.3\,M & 51.06 & 449.8\,M & 60.66 & 449.8\,M & 61.45 \\
Top-neuron copy, 1× & 66.1\,M & 52.63 & 107.4\,M & 60.18 & 107.4\,M & 60.94 \\
Whitened-SVD copy, 1× & 66.1\,M & 48.76 & 103.5\,M & 59.81 & 103.5\,M & 60.41 \\
Top-neuron copy, 4× & 267.4\,M & 54.04 & --- & --- & --- & --- \\
Whitened-SVD copy, 4× & 264.3\,M & 49.75 & --- & --- & --- & --- \\
\midrule
Tilt r32 $-$ keeping 33 & \multicolumn{2}{c}{+2.27 [+1.39, +3.18]*\holma} & \multicolumn{2}{c}{$-$0.37 [$-$1.01, +0.26]} & \multicolumn{2}{c}{+0.84 [+0.21, +1.48]*} \\
Tilt r32 $-$ keeping 34 & \multicolumn{2}{c}{+1.49 [+0.49, +2.49]*\holmt} & \multicolumn{2}{c}{$-$0.34 [$-$0.98, +0.31]} & \multicolumn{2}{c}{+0.16 [$-$0.53, +0.85]} \\
Tilt r8 $-$ keeping 33 & \multicolumn{2}{c}{+1.65 [+0.79, +2.53]*\holma} & \multicolumn{2}{c}{---} & \multicolumn{2}{c}{---} \\
Tilt r128 $-$ neuron copy 4× & \multicolumn{2}{c}{+0.01 [$-$1.02, +1.05]} & \multicolumn{2}{c}{---} & \multicolumn{2}{c}{---} \\
\bottomrule
\end{tabular}
\end{table}

\subsection{Evaluation check for the extra-expert arms}
\label{app:evalcheck}

\Cref{tab:budget} evaluates MoNE keeping 33 and 34 experts in the same process as every other arm, with constants taken from the slice-A fit files. MoNE's saved checkpoints, evaluated separately, give 49.95 and 51.22 on OLMoE. The constants are the same vectors: across the 512, 496 and 480 pruned experts of the 50\%, keep-33 and keep-34 checkpoints, the relative L2 distance between a checkpoint's constant and the fit's mean is at most $3.4\times10^{-3}$ (bf16 rounding), and the cosine is 1.0000. The two evaluation paths differ by +0.32 (keep 33) and $-$0.16 (keep 34) nine-task points, in opposite directions, and per task by at most 1.08 points (three RTE items):

\begin{table}[H]
\centering
\caption{In-process minus saved-checkpoint evaluation, OLMoE, same masks and same constants.}
\label{tab:H6}
\footnotesize
\begin{adjustbox}{max width=\linewidth}
\begin{tabular}{@{}lcccccccccc@{}}
\toprule
Arm & ARC-c & ARC-e & BoolQ & COPA & MMLU & OBQA & PIQA & RTE & Wino & Nine \\
\midrule
Keeping 33 & +0.51 & +0.13 & +0.37 & 0.00 & +0.09 & +0.20 & +0.22 & +1.08 & +0.24 & +0.32 \\
Keeping 34 & $-$0.34 & +0.13 & +0.03 & 0.00 & +0.16 & $-$1.00 & +0.22 & 0.00 & $-$0.71 & $-$0.16 \\
\bottomrule
\end{tabular}
\end{adjustbox}
\end{table}

\subsection{Slot remapping and COMPEL: implementation and details}
\label{app:others}

\textbf{Slot remapping (ConMoE).} Each pruned expert's module is replaced by a reference to its target retained expert in the same layer, so the router, the top-$k$ choice and the gate value are unchanged and the pruned expert's weights are dropped. When the target is also selected for the token, both slots run the target, with weight $g_i+g_{\pi(i)}$, as in ConMoE. \emph{Nearest} uses ConMoE's parameter distance (its Eq.~13) over the gate, up and down projections. \emph{Functional} follows ConMoE's Section 3.2 and Appendix A.1: the four retained experts nearest by that distance are candidates; up to 64 routed states per pruned expert are kept in each of two folds (even and odd calibration blocks, 32 blocks of 4,096 tokens from slice A, a uniform random subset with seed 0); each candidate's cost is $\mathrm{mean}_h\, g_i(h)^2\|E_i(h)-E_p(h)\|^2$; and the functional choice replaces the nearest only if both folds choose it and it beats the nearest on both. A routing replay reproduced each model's expert choice exactly in three layers before collection. Our deviations from ConMoE: routed states come from the unpruned model rather than the nearest-assigned compressed one, calibration is general text rather than one model per task, and the pool is MoNE's mask rather than ConMoE's Anchor Coverage pool. ConMoE has released no code (as of its arXiv v2, 12 October 2026).

\begin{table}[H]
\centering
\caption{Slot-remapping assignments at 50\% (MoNE masks). ``Changed'': slots where the functional choice differs from the nearest. Outcomes of the functional rule: functional / nearest already best / folds disagree (empty folds: none).}
\label{tab:H7}
\footnotesize
\begin{adjustbox}{max width=\linewidth}
\begin{tabular}{@{}lccccccc@{}}
\toprule
Model & Slots & Distinct targets (nearest / functional) & Most slots on one expert & Changed & Rule outcomes & Routed states per fold (median / min) & Token-slots served (nearest / functional) \\
\midrule
OLMoE & 512 & 200 / 170 & 16 / 19 & 284 & 284 / 85 / 143 & 64 / 64 & $1.82\times10^8$ / $1.74\times10^8$ \\
DeepSeek & 832 & 408 / 330 & 13 / 15 & 379 & 379 / 161 / 292 & 64 / 64 & $1.83\times10^8$ / $1.82\times10^8$ \\
Moonlight & 832 & 479 / 444 & 10 / 11 & 410 & 410 / 105 / 317 & 64 / 8 & $1.79\times10^8$ / $1.81\times10^8$ \\
\bottomrule
\end{tabular}
\end{adjustbox}
\end{table}

The functional rule changes about half of the slots and concentrates them on fewer retained experts (on OLMoE, 170 distinct targets instead of 200, with up to 19 pruned slots on one expert). The two folds disagree on 28--38\% of the pruned experts, so 64 states per fold estimate the functional cost noisily.

\textbf{COMPEL's compensation.} We implement Eq.~11 of \citet{yoon2026compel}: for each layer, every surviving expert $j$ is updated as $\theta_j \leftarrow \theta_j + \gamma\,F_j/(\sum_{k\in S}F_k+\epsilon)\sum_{i\in R}\theta_i$ for each of its three projection matrices, with $\gamma=0.1$ and $\epsilon=10^{-12}$, in float32 and stored back in bf16. $F_e$ (Eq.~5) is the squared norm of the gradient of each sequence's mean next-token loss with respect to expert $e$'s parameters, averaged over 128 sequences of 1,024 tokens (the first 1,024 tokens of the first 128 slice-A blocks); only within-layer ratios enter Eq.~11. The pruned experts are then removed by keep-scale or renormalized rerouting. The survivors' median relative weight change is 1.5\% (OLMoE), 1.6\% (DeepSeek) and 1.6\% (Moonlight). Before each Fisher pass, a self-test compares our differentiable MoE forward, which routes and mixes the experts itself, with the model's own MoE block on its first, middle and last MoE layer (512 random tokens): the selected experts are identical in all nine checks, and the relative output error is at most $4.0\times10^{-3}$ on OLMoE, whose block adds expert outputs in a different order in bf16, and exactly zero on \dsv{} and Moonlight, whose float32 mixing we replicate. Every routed expert received a gradient (0 of 1,024, 1,664 and 1,664 experts have zero Fisher), so Eq.~11 spreads the pruned weights over all survivors. The mean calibration loss during the pass was 2.30, 2.30 and 2.47 nats per token. Deviations from COMPEL: our masks are MoNE's uniform per-layer masks, not COMPEL's layer-adaptive ones; calibration is Zyda-2 slice A, not C4; COMPEL's OLMoE is the 0924 checkpoint, ours 0125. Its code was not public on 27 October 2026.

\begin{table}[H]
\centering
\caption{Slot remapping and COMPEL against the other operators, 50\%, MoNE masks, draw A, nine-task, 95\% intervals. Keep: rerouting at the trained gate scale (renormalized on Moonlight).}
\label{tab:H8}
\footnotesize
\begin{adjustbox}{max width=\linewidth}
\begin{tabular}{@{}lccc@{}}
\toprule
Contrast & OLMoE & DeepSeek & Moonlight \\
\midrule
Remap nearest $-$ keep & $-$6.05 [$-$7.32, $-$4.81]* & $-$3.04 [$-$4.15, $-$1.94]* & $-$1.15 [$-$2.17, $-$0.16]* \\
Remap nearest $-$ constant & $-$3.66 [$-$4.85, $-$2.51]* & $-$1.69 [$-$2.90, $-$0.48]* & $-$1.33 [$-$2.34, $-$0.34]* \\
Remap functional $-$ nearest & +2.04 [+0.78, +3.27]* & $-$0.80 [$-$1.84, +0.23] & $-$0.70 [$-$1.62, +0.23] \\
Remap functional $-$ keep & $-$4.01 [$-$5.24, $-$2.81]* & $-$3.84 [$-$5.05, $-$2.64]* & $-$1.86 [$-$2.90, $-$0.84]* \\
Remap functional $-$ constant & $-$1.63 [$-$2.85, $-$0.44]* & $-$2.48 [$-$3.77, $-$1.19]* & $-$2.04 [$-$2.99, $-$1.07]* \\
Tilt $-$ remap functional & +5.08 [+3.92, +6.29]* & +2.96 [+1.68, +4.26]* & +3.40 [+2.37, +4.44]* \\
\midrule
COMPEL + keep $-$ keep & $-$3.26 [$-$4.43, $-$2.06]* & +0.39 [$-$0.17, +0.96] & $-$0.61 [$-$1.30, +0.08] \\
COMPEL + keep $-$ constant & $-$0.87 [$-$1.90, +0.15] & +1.74 [+0.99, +2.53]* & $-$0.79 [$-$1.65, +0.05] \\
COMPEL + keep $-$ tilt & $-$4.32 [$-$5.38, $-$3.31]* & +1.26 [+0.48, +2.05]* & $-$2.15 [$-$3.01, $-$1.31]* \\
COMPEL + renorm $-$ renorm & $-$1.86 [$-$2.97, $-$0.75]* & +0.07 [$-$0.72, +0.81] & --- \\
\bottomrule
\end{tabular}
\end{adjustbox}
\end{table}

\begin{table}[H]
\centering
\caption{Second calibration draw for the fitted survivor operators on OLMoE: MoNE's 50\% mask (from slice A) with COMPEL's Fisher weights and the functional assignment refitted on slice B. Nine-task, 95\% intervals; mean(A,B) is one paired test of the average over both draws; B $-$ A compares the two draws.}
\label{tab:drawsB}
\footnotesize
\begin{adjustbox}{max width=\linewidth}
\begin{tabular}{@{}lcccc@{}}
\toprule
Contrast & Draw A & Draw B & Mean (A, B) & B $-$ A \\
\midrule
COMPEL + keep $-$ keep & $-$3.26 [$-$4.43, $-$2.06]* & $-$3.13 [$-$4.31, $-$1.94]* & $-$3.19 [$-$4.36, $-$2.01]* & +0.12 [$-$0.10, +0.34] \\
COMPEL + renorm $-$ renorm & $-$1.86 [$-$2.97, $-$0.75]* & $-$1.82 [$-$2.92, $-$0.75]* & $-$1.84 [$-$2.90, $-$0.78]* & +0.04 [$-$0.51, +0.55] \\
Remap functional $-$ keep & $-$4.01 [$-$5.24, $-$2.81]* & $-$4.24 [$-$5.34, $-$3.16]* & $-$4.13 [$-$5.12, $-$3.15]* & $-$0.23 [$-$1.45, +0.99] \\
Remap functional $-$ nearest & +2.04 [+0.78, +3.27]* & +1.80 [+0.67, +2.97]* & +1.92 [+0.88, +2.95]* & $-$0.23 [$-$1.45, +0.99] \\
\bottomrule
\end{tabular}
\end{adjustbox}
\end{table}

Keep and nearest remapping fit nothing, so every change between the draws comes from the refitted operator; the two remap rows therefore share their B $-$ A. Every contrast keeps its sign and stays significant on slice B.

\subsection{Relation to numbers reported by ConMoE, COMPEL and HOPE}
\label{app:disputes}\label{app:reported}
We describe how our results relate to numbers these papers report, including gaps we could not reconcile, without assigning a cause.

\textbf{ConMoE's reported OLMoE results.} Our finding that slot remapping loses to MoNE's constant (\cref{sec:others}) agrees with ConMoE's own results on the same OLMoE checkpoint \citep{yao2026conmoe}. There MoNE scores above ConMoE at 50\% and 25\% expert reduction (58.90 against 58.03 and 66.03 against 65.81, six tasks), although ConMoE compresses one model per task. With frequency and REAP pools, nearest assignment scores below deleting the same experts (42.61 against 49.95 and 48.72 against 51.32); with ConMoE's own pool and functional assignment it is above REAP. With Zyda-2 calibration and its own pool, ConMoE reports the functional rule 2.75 points above the nearest at 50\% on OLMoE, close to our +2.04. Our pool (MoNE's mask), calibration (general text) and routed states (the unpruned model) differ from ConMoE's.

\textbf{COMPEL's reported OLMoE perplexity.} Our setting differs from COMPEL's: MoNE's uniform masks instead of its layer-adaptive ones, Zyda-2 instead of C4 calibration, and OLMoE-1B-7B-0125 instead of 0924. COMPEL reports near-lossless 50\% pruning on OLMoE (WikiText-2 perplexity 13.76 against 13.52 dense); on the same model family at 50\% we measure WikiText perplexity rising from 8.31 to 51.65 under the constant (\cref{tab:ppl}), with a different perplexity definition and context length. We could not resolve the difference because COMPEL's code was not public.

\textbf{HOPE's off-diagonal shares.} These shares are not comparable to HOPE's. HOPE reports off-diagonal terms averaging 33\% of the diagonal on GLM-4.5-Air \citep{tseng2026hope}, but its terms are products of gate-weighted output norms, averaged over the tokens where both experts are routed, for deletion with renormalized gates. Norm products bound our signed inner products from above. Measured our way with deletion instead of the constant, the off-diagonal share on OLMoE is 16\%.

\subsection{Granite: the gate-scale test in detail}
\label{app:granite}

\textbf{Model and router.} Granite-3.0-3B-A800M-base \citep{granite2024} has 32 MoE layers of 40 routed experts (hidden size 1,536, expert width 512), top-8 routing and no shared experts: 3.3\,B parameters, 800\,M of them active. Its router takes the top-$k$ of the router logits and applies a softmax over those $k$ only, so the selected gates always sum to one. This is the same function as OLMoE's MoE block with its top-$k$ gates renormalized. We rebuilt every Granite MoE block as such a block, copying the weights, so that MoNE's scoring and every operator run on it unchanged.

\textbf{Conversion check.} On 2,048 tokens of slice A the converted and original blocks select the same experts for every token in all 32 layers, with a relative output error of at most $4.5\times10^{-3}$ in bf16. In float32 the next-token loss is 2.082205 nats for both models and top-1 agreement is 1.000. In bf16 the losses are 2.0829 and 2.0823 nats and top-1 agreement is 0.972, against 0.963 between the original model's own bf16 and float32 runs. With no expert pruned and with 20 pruned, in layers 0, 16 and 31, both rerouting semantics match references written from Granite's router definition exactly, and they differ from each other by 21--43\% of the output norm. No token reached a pruned expert in any Granite run.

\textbf{Masks.} MoNE's fusion score with per-layer ranking, on 1,000 sequences of 4,096 tokens from Zyda-2 slice A tokenized for Granite; MoNE's routing score uses the gates the model applies, which on Granite are renormalized. The masks prune 10, 20 and 30 experts per layer (320, 640 and 960 in total) and are nested: every expert pruned at 25\% is pruned at 50\%, and every expert pruned at 50\% at 75\%.

\textbf{Pruned experts without calibration tokens.} Nine pruned experts are never selected on the calibration text, all in the first two layers (layer 0: experts 20 and 29; layer 1: experts 1, 8, 11, 13, 19, 21 and 26). Because the masks are nested, the same nine are pruned at every ratio. With no routed tokens there is nothing to average, so their constant is zero, the value MoNE's own code assigns, and they get no tilt; the tilt models use the zero constant for them.

\textbf{Operators.} The constant and the rank-32 tilt (fitted on slice A), keep-scale and renormalized rerouting at every ratio, and nearest slot remapping (ConMoE's parameter distance) at 50\%. Granite enters only the gate-scale test.

\begin{table}[H]
\centering
\caption{Granite: every contrast, MoNE masks, draw A, 95\% intervals for the nine-task mean, big-5 and MMLU; $p$: two-sided bootstrap p-value for the nine-task mean. Only the three renorm $-$ keep rows are in the Holm families (\cref{tab:H2}). Dense 61.82; the constant leaves 3.87, 16.55 and 20.52 points of damage at 25, 50 and 75\%.}
\label{tab:granitefull}
\footnotesize
\begin{adjustbox}{max width=\linewidth}
\begin{tabular}{@{}lcccc@{}}
\toprule
Contrast & Nine-task & $p$ & Big-5 & MMLU \\
\midrule
25\% renorm $-$ keep & +0.28 [$-$0.54, +1.09] & 0.487 & +0.61 [$-$0.07, +1.29] & +1.08 [+0.47, +1.69]* \\
25\% tilt $-$ constant & +1.29 [+0.41, +2.18]* & 0.004 & +1.76 [+1.07, +2.45]* & +1.57 [+0.97, +2.19]* \\
25\% keep $-$ constant & $-$0.83 [$-$1.54, $-$0.15]* & 0.020 & $-$0.74 [$-$1.38, $-$0.08]* & $-$0.41 [$-$0.98, +0.16] \\
25\% renorm $-$ constant & $-$0.55 [$-$1.41, +0.26] & 0.190 & $-$0.13 [$-$0.78, +0.52] & +0.67 [+0.03, +1.30]* \\
25\% tilt $-$ renorm & +1.85 [+1.00, +2.71]* & 2.0e$-$4 & +1.89 [+1.14, +2.64]* & +0.90 [+0.21, +1.60]* \\
\midrule
50\% renorm $-$ keep & +2.08 [+1.08, +3.11]* & 4.0e$-$4 & +2.46 [+1.57, +3.38]* & +1.30 [+0.59, +1.98]* \\
50\% tilt $-$ constant & +3.36 [+2.19, +4.56]* & 2.0e$-$4 & +3.56 [+2.62, +4.50]* & +3.39 [+2.57, +4.19]* \\
50\% keep $-$ constant & $-$0.62 [$-$1.60, +0.33] & 0.192 & $-$0.72 [$-$1.61, +0.15] & +0.06 [$-$0.32, +0.46] \\
50\% renorm $-$ constant & +1.47 [+0.44, +2.45]* & 0.005 & +1.74 [+0.83, +2.66]* & +1.37 [+0.66, +2.06]* \\
50\% tilt $-$ renorm & +1.90 [+0.77, +3.05]* & 0.001 & +1.82 [+0.87, +2.77]* & +2.02 [+1.22, +2.82]* \\
\midrule
75\% renorm $-$ keep & +0.27 [$-$0.81, +1.36] & 0.623 & $-$0.89 [$-$1.99, +0.21] & +0.32 [$-$0.25, +0.88] \\
75\% tilt $-$ constant & +1.58 [+0.57, +2.59]* & 0.003 & +3.01 [+1.94, +4.08]* & $-$0.02 [$-$0.31, +0.28] \\
75\% keep $-$ constant & $-$0.43 [$-$1.46, +0.59] & 0.402 & $-$0.22 [$-$1.22, +0.79] & $-$0.06 [$-$0.53, +0.42] \\
75\% renorm $-$ constant & $-$0.16 [$-$1.28, +0.96] & 0.774 & $-$1.11 [$-$2.21, $-$0.04]* & +0.26 [$-$0.16, +0.70] \\
75\% tilt $-$ renorm & +1.75 [+0.53, +2.96]* & 0.004 & +4.12 [+3.03, +5.25]* & $-$0.28 [$-$0.72, +0.14] \\
\midrule
50\% remap nearest $-$ renorm & $-$3.51 [$-$4.55, $-$2.46]* & 2.0e$-$4 & $-$4.43 [$-$5.48, $-$3.38]* & $-$1.32 [$-$2.01, $-$0.64]* \\
50\% remap nearest $-$ keep & $-$1.43 [$-$2.34, $-$0.51]* & 0.002 & $-$1.96 [$-$2.96, $-$0.99]* & $-$0.01 [$-$0.52, +0.49] \\
50\% remap nearest $-$ constant & $-$2.04 [$-$3.01, $-$1.12]* & 2.0e$-$4 & $-$2.68 [$-$3.68, $-$1.68]* & +0.05 [$-$0.51, +0.62] \\
\bottomrule
\end{tabular}
\end{adjustbox}
\end{table}

\textbf{Where the 50\% reversal comes from.} Renormalized rerouting is ahead of keep-scale rerouting on all nine tasks: ARC-e +5.38, WinoGrande +2.85, RTE +2.53, PIQA +2.18, OpenBookQA +1.60, BoolQ +1.41, MMLU +1.30, COPA +1.00 and ARC-c +0.51. MMLU is not evidence, because both models score below its 25\% chance level (24.5 and 23.2); the big-5 gain, +2.46 [+1.57, +3.38], carries the result. At 25\% the only significant renorm $-$ keep difference is MMLU's (+1.08*); at 75\% the big-5 difference is negative ($-$0.89, n.s.) and the nine-task one positive, so neither ratio shows a consistent direction.

\textbf{Guessing level.} Uniform guessing scores 25\% on ARC-c, ARC-e, MMLU and OpenBookQA and 50\% on COPA, PIQA and WinoGrande; always giving the majority answer scores 62.2\% on BoolQ and 52.7\% on RTE. The nine-task mean of these is 40.5. At 75\% every Granite model scores 40.87--42.88, and both rerouted models score below BoolQ's majority rate (keep 54.13, renorm 45.87; \cref{tab:granitetasks}). BoolQ dominates the 75\% contrasts. The tilt's 1.58-point gain over the constant is significant and comes from the big-5 (+3.01*), not from MMLU ($-$0.02), but 0.99 of it is BoolQ (+8.93, to 62.05, the majority-answer rate), which may be a shift in answer bias rather than ability (\cref{sec:auc}); without BoolQ the gain is 0.59 points of the nine-task mean, with ARC-e (+3.45), PIQA (+3.05) and OpenBookQA (+2.20) ahead and COPA ($-$3.00) and ARC-c ($-$1.70) behind. Likewise renorm $-$ keep's negative big-5 difference at 75\% ($-$0.89) is BoolQ's $-$8.26; on the other four big-5 tasks renorm is ahead by 0.95 on average. Neither direction at 75\% is clear.

\begin{table}[H]
\centering
\caption{Granite: per-task accuracy (\%), MoNE masks, draw A; Nine: the nine-task mean. Guessing: 25\% on ARC-c, ARC-e, MMLU and OpenBookQA, 50\% on COPA, PIQA and WinoGrande; majority answer 62.2\% on BoolQ and 52.7\% on RTE.}
\label{tab:granitetasks}
\footnotesize
\begin{adjustbox}{max width=\linewidth}
\begin{tabular}{@{}lcccccccccc@{}}
\toprule
Operator & ARC-c & ARC-e & BoolQ & COPA & MMLU & OBQA & PIQA & RTE & Wino & Nine \\
\midrule
Dense & 46.50 & 75.46 & 64.89 & 82.00 & 47.32 & 41.60 & 78.18 & 55.23 & 65.19 & 61.82 \\
\midrule
\multicolumn{11}{@{}l}{\emph{25\% pruned}} \\
\quad Constant & 37.03 & 64.44 & 68.20 & 83.00 & 38.11 & 35.60 & 71.55 & 58.48 & 65.11 & 57.95 \\
\quad Tilt r32 & 39.76 & 67.55 & 69.82 & 85.00 & 39.68 & 35.20 & 73.12 & 58.12 & 64.88 & 59.24 \\
\quad Keep & 37.03 & 64.39 & 66.54 & 82.00 & 37.69 & 35.00 & 70.73 & 56.68 & 63.93 & 57.11 \\
\quad Renorm & 39.08 & 64.35 & 66.51 & 78.00 & 38.78 & 35.60 & 71.71 & 58.48 & 64.01 & 57.39 \\
\midrule
\multicolumn{11}{@{}l}{\emph{50\% pruned}} \\
\quad Constant & 26.62 & 39.69 & 63.30 & 62.00 & 23.14 & 26.20 & 57.78 & 54.87 & 53.83 & 45.27 \\
\quad Tilt r32 & 27.82 & 47.31 & 64.07 & 69.00 & 26.53 & 29.00 & 60.88 & 54.15 & 58.96 & 48.64 \\
\quad Keep & 28.16 & 39.02 & 61.16 & 61.00 & 23.21 & 27.00 & 57.34 & 53.07 & 51.93 & 44.65 \\
\quad Renorm & 28.67 & 44.40 & 62.57 & 62.00 & 24.51 & 28.60 & 59.52 & 55.60 & 54.78 & 46.74 \\
\quad Remap nearest & 27.13 & 36.03 & 57.80 & 60.00 & 23.19 & 25.00 & 56.09 & 53.07 & 50.75 & 43.23 \\
\midrule
\multicolumn{11}{@{}l}{\emph{75\% pruned}} \\
\quad Constant & 24.74 & 28.37 & 53.12 & 64.00 & 22.98 & 23.00 & 52.45 & 52.71 & 50.36 & 41.30 \\
\quad Tilt r32 & 23.04 & 31.82 & 62.05 & 61.00 & 22.96 & 25.20 & 55.50 & 52.71 & 51.70 & 42.88 \\
\quad Keep & 23.46 & 28.58 & 54.13 & 60.00 & 22.92 & 24.60 & 53.32 & 52.35 & 48.46 & 40.87 \\
\quad Renorm & 23.21 & 30.68 & 45.87 & 64.00 & 23.24 & 26.80 & 54.95 & 52.71 & 48.78 & 41.14 \\
\bottomrule
\end{tabular}
\end{adjustbox}
\end{table}

\textbf{Against the constant.} Keep-scale rerouting departs from Granite's trained gate scale and loses to the constant at 25\% ($-$0.83*); renormalized rerouting beats the constant only at 50\% (+1.47*). The tilt is above the constant and both rerouting scales at every ratio, significantly against the constant and against renormalized rerouting (keep-scale rerouting was not tested against it directly). At 50\% the order of the operators is the same as on OLMoE once each router's trained scale is taken into account: the tilt, rerouting at the trained scale, the constant, rerouting at the other scale, and nearest remapping last (Granite 48.64, 46.74, 45.27, 44.65, 43.23; OLMoE 52.54, 51.48, 49.09, 48.08, 45.43).

\subsection{Merging baselines as reported by MoRA}
\label{app:merging}

MoRA's paper reports MC-SMoE and HC-SMoE on \dsv{} and Moonlight with C4 calibration, on its nine tasks (MoRA-9: our nine with HellaSwag instead of COPA) \citep{sun2026mora}. We transcribed those rows (their per-task values reproduce their averages to 0.01) and compare them with our operators on the same task set (\cref{tab:D6}). The merging numbers are MoRA's runs, not ours, and our runs of two other baselines score above MoRA's rows for them (\cref{app:D4}), so we treat them as indicative.

\begin{table}[H]
\centering
\caption{MoRA-9 accuracy: merging as reported by MoRA against our operators (ours: MoNE masks; C4 or Zyda-2 calibration; keep: rerouting at the trained gate scale).}
\label{tab:H9}
\footnotesize
\begin{adjustbox}{max width=\linewidth}
\begin{tabular}{@{}lccccccc@{}}
\toprule
Setting & MC-SMoE (reported) & HC-SMoE (reported) & MoNE-C4 + const & MoNE-C4 + keep & Zyda + const & Zyda + tilt r32 & Zyda + keep \\
\midrule
DeepSeek 25\% & 57.66 & 62.66 & 64.17 & 64.29 & 63.79 & 64.04 & 64.13 \\
DeepSeek 50\% & 45.83 & 53.84 & 53.37 & 55.41 & 57.52 & 58.00 & 59.16 \\
Moonlight 25\% & 63.46 & 56.87 & 66.74 & 66.90 & 67.32 & 67.47 & 67.40 \\
Moonlight 50\% & 44.55 & 44.54 & 55.78 & 55.89 & 58.19 & 59.54 & 58.40 \\
\bottomrule
\end{tabular}
\end{adjustbox}
\end{table}

\textbf{Merging is below the training-free operators in every setting except one.} The exception is \dsv{} 50\% with MoNE's constant under C4 calibration (53.37 against HC-SMoE's 53.84). With keep-scale rerouting on the same C4 mask MoNE is ahead (+1.57), and with Zyda-2 calibration every operator is ahead by 3.7--5.3 points; on Moonlight our operators lead by 3.3--15.0 points. Part of the 50\% margin is the calibration corpus: on \dsv{} 50\% Zyda-2 instead of C4 adds 4.15 points to MoNE's constant, more than any operator.

\subsection{Second-draw masks from MoNE's own code}
\label{app:masksB}

So that every overlap in \cref{tab:overlap} comes from one scorer, we ran MoNE's released code on slice B for all three models. The overlaps with slice A's masks are 93.6\% (OLMoE), 95.8\% (\dsv{}) and 95.8\% (Moonlight), the same for the first two as a re-implementation of MoNE's score gives.

\section{Candidate Explanations and Scope Notes}
\label{app:moved}

\subsection{Candidate explanations}
\label{app:hypotheses}\label{app:mv-granitenull}\label{app:mv-remapshared}\label{app:mv-concentration}
\textbf{Granite's nulls at 25\% and 75\%.} Two observations, made after the results, may explain the nulls; we did not test either. On Granite, keep's gates differ from the trained ones only for tokens whose top-$k$ contained a pruned expert, and at 25\% MoNE's score, which carries routing frequency, prunes experts that are selected comparatively rarely; on OLMoE and \dsv{}, by contrast, renormalization rescales every token's gates, because their softmax runs over all experts and every pruned expert holds some of its mass. At 75\%, every Granite model is within 2.4 points of guessing (40.5, from uniform guessing with the majority answer on BoolQ and RTE), and with 10 of 40 experts left and $k=8$ each token uses eight of the ten survivors under either scale. Some differences are still measurable there (the tilt is 2.2--3.5 points ahead of the constant on ARC-e, OpenBookQA and PIQA), so a null between the two rerouted models is weak evidence either way. BoolQ moves by 8--9 points between models at 75\%, which mixes answer bias with ability (\cref{sec:auc,app:granite}).

\textbf{Remapping cost and shared experts.} Nearest remapping costs least on Moonlight. One candidate is shared experts, which carry part of every token's MoE output: against the trained scale, nearest remapping costs most on the two models without them (OLMoE 6.05, Granite 3.51) and less on the two with two each (\dsv{} 3.04, Moonlight 1.15). With four models that also differ in size and training, this is a hypothesis.

\textbf{Routing concentration under GSM8K calibration.} GSM8K concentrates routing on fewer experts, and MoNE's score carries routing frequency, so it prunes the experts GSM8K leaves idle. Per million calibration tokens, the least-used pruned expert receives about 28 times fewer tokens under GSM8K than under general text on OLMoE, 10 times fewer on \dsv{} and 25 times fewer on Moonlight. On Moonlight, 5 pruned experts receive 39--128 GSM8K tokens, too few to fit a rank-32 map; they fall back to the constant.

\subsection{Which results are exposed to calibration noise}
\label{app:mv-draws}
\emph{Draws and masks.} Every mask comes from slice A, so a second calibration draw can change a result only through what an operator fits. Rerouting (keep or renorm) and nearest-rule remapping fit nothing, so the rerouted and remapped models, including their generative and 5-shot scores, are the same for every draw. Their contrasts with the constant inherit the constant's draw noise, which is small: its nine-task score moves by at most 0.39 points across draws (\cref{tab:A4}). The tilt's gain over the constant moves by at most 0.20 points at 50\% across three draws and 0.58 at 25\% across two. On OLMoE, functional remapping and COMPEL were refitted on a second draw; both results move by at most 0.23 points (\cref{tab:drawsB}). Results fitted on draw A only therefore remain exposed to calibration noise only for functional remapping and COMPEL on \dsv{} and Moonlight, the whitened-SVD and neuron copies, the tilt's generative, 5-shot and crossed-grid results (except OLMoE's SGE mask), and every Granite result. The calibration-domain results use one GSM8K draw, and the mask grid covers 50\% on two models. On OLMoE, choosing the mask for the tilt and Fisher-weighted selection did not improve on interaction-aware selection for the constant (\cref{app:E5,app:E6}).